\documentclass[pdflatex,sn-nature]{sn-jnl}

\usepackage{graphicx}%
\usepackage{multirow}%
\usepackage{amsmath,amssymb,amsfonts}%
\usepackage{amsthm}%
\usepackage{mathrsfs}%
\usepackage[title]{appendix}%
\usepackage{xcolor}%
\usepackage{textcomp}%
\usepackage{manyfoot}%
\usepackage{booktabs}%
\usepackage{algorithm}%
\usepackage{algorithmicx}%
\usepackage{algpseudocode}%
\usepackage{listings}%
\usepackage{geometry}
\usepackage{subcaption}
\usepackage[percent]{overpic}
\usepackage{tikz}
\usetikzlibrary{calc}
\usepackage[table]{xcolor}
\usepackage{threeparttable}
\usepackage{tabularx}
\usepackage{arydshln}
\usepackage[normalem]{ulem}
\usepackage{times}

\newcolumntype{Y}{>{\centering\arraybackslash}X}

\newcommand{\imgpanellabel}[3]{%
\begin{tikzpicture}
    \node[anchor=north west, inner sep=0] (img) at (0,0)
        {\includegraphics[width=#1]{#3}};
    \path[use as bounding box] (img.south west) rectangle (img.north east);
    \node[overlay, anchor=north west, font=\bfseries] 
        at ([xshift=-0cm,yshift=0.2cm]img.north west) {#2};
\end{tikzpicture}%
}

\newcommand{\scalingpanellabel}[3]{%
\begin{tikzpicture}
    \node[anchor=north west, inner sep=0] (img) at (0,0)
        {\includegraphics[#1]{#3}};
    \path[use as bounding box] (img.south west) rectangle (img.north east);
    \node[overlay, anchor=north west, font=\bfseries]
        at ([xshift=-0.1cm,yshift=0.2cm]img.north west) {#2};
\end{tikzpicture}%
}

\newcommand{\tablepanellabel}[2]{%
\begin{tikzpicture}
    \node[anchor=north west, inner sep=0] (tab) at (0,0) {%
        \begin{minipage}{\textwidth}
        \centering
        #2
        \end{minipage}
    };
    \path[use as bounding box] (tab.south west) rectangle (tab.north east);
    \node[overlay, anchor=north west, font=\bfseries]
        at ([xshift=-0.1cm,yshift=0.4cm]tab.north west) {#1};
\end{tikzpicture}%
}

\newcommand{\overviewpanellabel}[3]{%
\begin{tikzpicture}
    \node[anchor=north west, inner sep=0] (img) at (0,0)
        {\includegraphics[#1]{#3}};
    \path[use as bounding box] (img.south west) rectangle (img.north east);
    \node[overlay, anchor=north west, font=\bfseries]
        at ([xshift=-0.1cm,yshift=0.35cm]img.north west) {#2};
\end{tikzpicture}%
}

\newcommand{\analysispanellabel}[3]{%
\begin{tikzpicture}
    \node[anchor=north west, inner sep=0] (img) at (0,0)
        {\includegraphics[#1]{#3}};
    \path[use as bounding box] (img.south west) rectangle (img.north east);
    \node[overlay, anchor=north west, font=\bfseries]
        at ([xshift=-0.10cm,yshift=0.18cm]img.north west) {#2};
\end{tikzpicture}%
}

\theoremstyle{thmstyleone}%
\theoremstyle{thmstyletwo}%

\theoremstyle{thmstylethree}%

\begin{document}



\title[INCEPT]{Taming foundation model with invariance-oriented pre-training for broad-spectrum EEG analysis across signal-level, brain-state, and brain-health tasks}








\author[1]{\fnm{Yulong} \sur{Dou}}\email{douyl2023@shanghaitech.edu.cn}

\author[1]{\fnm{Han} \sur{Wu}}\email{wuhan2022@shanghaitech.edu.cn}

\author[1]{\fnm{Guo} \sur{Chen}}\email{chenguo2022@shanghaitech.edu.cn}

\author[1]{\fnm{Fangmao} \sur{Ju}}\email{jufm2025@shanghaitech.edu.cn}

\author*[1]{\break\fnm{Zhiming} \sur{Cui}}\email{cuizhm@shanghaitech.edu.cn}

\author*[1,2,3]{\fnm{Dinggang} \sur{Shen}}\email{Dinggang.Shen@gmail.com}

\affil[1]{\orgdiv{School of Biomedical Engineering \& State Key Laboratory of Advanced Medical Materials and Devices}, 
\orgname{ShanghaiTech University}, 
\orgaddress{\city{Shanghai}, \postcode{201210}, \country{China}}}

\affil[2]{\orgname{Shanghai United Imaging Intelligence Co., Ltd.}, 
\orgaddress{\city{Shanghai}, \postcode{200230}, \country{China}}}

\affil[3]{\orgname{Shanghai Clinical Research and Trial Center}, 
\orgaddress{\city{Shanghai}, \postcode{201210}, \country{China}}}

\abstract{
Electroencephalography (EEG) is a widely used window into human brain function, but most EEG models remain tied to a one-dataset-one-model supervised paradigm. Recent EEG foundation models offer a route toward reusable representations, but most remain reconstruction-centered, assuming that EEG content predictable from local context is necessarily transferable neural information. Here we present INCEPT, an invariance-oriented EEG foundation model trained on over 11,000 hours of unlabelled clinical EEG. Rather than prioritizing signal recovery alone, INCEPT learns representation-level stability across correlated EEG observations, separating stable neural structure and essential subject-sensitive information from the nuisance variability that dominates scalp recordings while preserving subject-, state- and condition-discriminative information. We evaluate INCEPT on a broad-spectrum benchmark of ten datasets spanning three levels of post-acquisition EEG analysis: signal-level assessment, brain-state decoding, and brain-health evaluation. INCEPT ranks first among recent EEG foundation models on 26 of 30 linear-probing metrics and 24 of 30 fine-tuning metrics, and also surpasses strong task-specific specialist encoders across diverse downstream settings. Objective ablations and representation analyses further show that invariance-oriented pre-training improves transfer and organizes subject-sensitive neural representations beyond reconstruction alone. These results establish invariance learning as a promising principle for building reusable EEG foundation models.}

\keywords{Electroencephalography, EEG foundation model, Transferable neural representation, Self-supervised learning}

\maketitle
\section{Introduction}
Electroencephalography (EEG) is a central tool for studying human brain activity in neuroscience, clinical neurophysiology and brain-computer interfaces \cite{niedermeyer2005electroencephalography}. After acquisition, EEG supports a broad spectrum of analysis tasks that can be viewed at three progressively higher levels: signal-level assessment, brain-state decoding, and brain-health evaluation. Signal-level assessment focuses on the quality, abnormality and reliability of the recording itself. Brain-state decoding aims to infer transient physiological or behavioural states from ongoing neural activity. Brain-health evaluation further asks whether EEG contains clinically relevant variation associated with psychiatric, cognitive or neurological conditions. These tasks differ in temporal scale, label granularity, electrode configuration, cohort composition and clinical or behavioural context, yet they all require models to extract meaningful neural information from noisy scalp recordings. An ideal EEG representation model should therefore support not only one benchmark or one label space, but broad-spectrum EEG analysis.

This broad-spectrum setting makes reusable EEG modelling both important and difficult. Although EEG is non-invasive, inexpensive, portable and temporally precise, scalp measurements have a low signal-to-noise ratio and reflect the interaction of neural activity, volume conduction, scalp anatomy, electrode placement, referencing, behavioural state and environmental artifacts \cite{nunez1997eeg,jiang2019removal}. Consequently, signals recorded from different subjects, devices, montages or task contexts can vary substantially even when they contain related neural information. A model intended for broad-spectrum EEG analysis must therefore learn representations that remain useful across signal-level assessment, brain-state decoding and brain-health evaluation, rather than representations that are only optimized for one dataset or one application level.

However, most deep learning approaches for EEG decoding still follow a task-specific supervised paradigm, as exemplified by EEGNet \cite{lawhern2018eegnet}, ST-Transformer \cite{song2021transformer}, EEGConformer \cite{song2022eeg} and SPaRCNet \cite{jing2023development}.
In this setting, a model is trained from scratch on a single dataset with one acquisition protocol and one label space.
Such models can perform well when training and test conditions are tightly matched, but they often learn features entangled with subject cohort, montage design, preprocessing choices, annotation protocol and device characteristics.
This one-dataset-one-model paradigm fragments EEG analysis and limits the reuse of neural representations across tasks, subjects and acquisition configurations, thereby constraining broad-spectrum EEG applications.

Self-supervised foundation pre-training offers a route toward reusable EEG representations.
Large clinical and experimental EEG archives contain thousands of hours of unlabelled recordings that capture rich variation in neural dynamics and physiological state.
An EEG foundation model should convert such data into a transferable starting point for broad-spectrum downstream analysis \cite{liu2026eeg,dai2026eeg}.
However, the pre-training objective is critical.
Unlike text or natural images \cite{devlin2019bert,brown2020language,he2022masked,oquab2024dinov2}, EEG is neither symbolic nor spatially stable.
It is also strongly subject-sensitive: short scalp recordings can contain stable individual physiological signatures, and state- or condition-related effects are often expressed as deviations from a subject-specific baseline rather than as absolute waveform patterns.
A transferable EEG encoder should therefore preserve subject-stable physiological structure while keeping state-, disease- and artifact-related variations accessible for downstream discrimination.
Yet, pure reconstruction-based objectives, as commonly adopted by recent EEG foundation models such as CBraMod \cite{wang2025cbramod}, CSBrain \cite{NEURIPS2025_7e199ad8} and CodeBrain \cite{ma2026codebrain}, do not explicitly enforce this form of discrimination and can therefore learn the signal details that are useful for reconstruction but not necessarily transferable.

\begin{figure*}[!t]
    \centering
    \includegraphics[width=\textwidth]{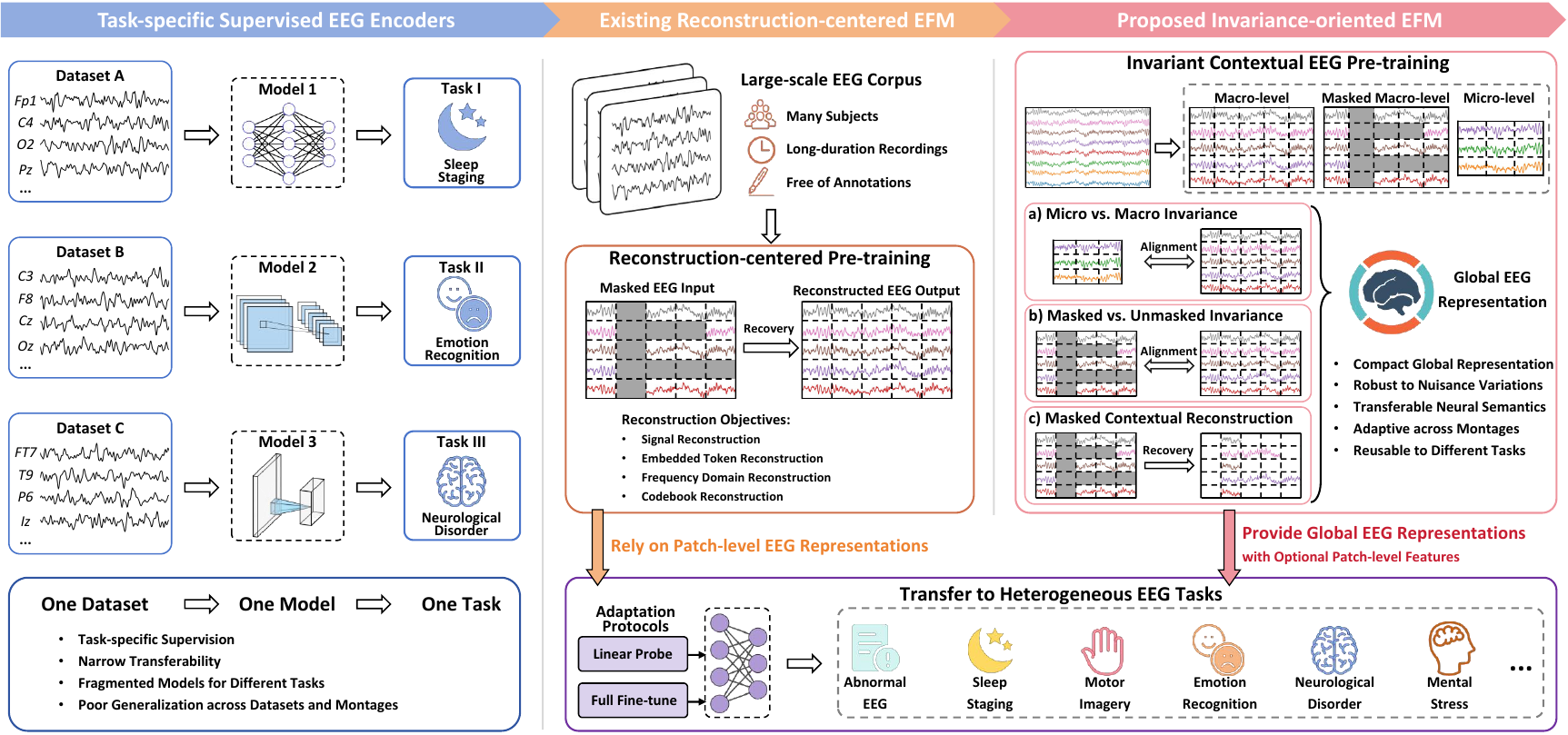}
    \caption{
    \textbf{Comparison of EEG modelling paradigms.}
    Schematic comparison of task-specific supervised EEG encoders, reconstruction-centered EEG foundation models and the proposed invariance-oriented INCEPT framework.
    INCEPT learns compact global EEG representations through invariance-oriented contextual pre-training while retaining patch-level features as optional outputs.
    }
    \label{fig:overview}
\end{figure*}

We introduce INCEPT, short for \underline{IN}variance-oriented \underline{C}ontextual \underline{E}EG \underline{P}re-\underline{T}raining, an invariance-oriented EEG foundation model designed to learn transferable whole-brain representations from unlabelled EEG.
INCEPT combines masked contextual learning with invariance-oriented pre-training: contextual learning captures local temporal, spectral and cross-channel dependencies, whereas invariance learning asks which essential subject-sensitive information remains stable across different observations of the same subject and recording context.
To reduce dependence on fixed channel identities or channel order, INCEPT further encodes continuous scalp positions using spherical-harmonics-based electrode embeddings \cite{seeley1966spherical}.
The resulting compact global representation is designed to be readable by a lightweight supervised head when the backbone is frozen, and to be reshaped by downstream labels when the backbone is fully fine-tuned.
Together, these design choices define the invariance-oriented modelling paradigm summarized in Fig.~\ref{fig:overview}, which contrasts task-specific supervised encoders trained separately for each dataset, reconstruction-centered EEG foundation models that emphasize local signal recovery, and INCEPT as a compact whole-brain representation learner combining contextual modelling with cross-view invariance learning.

We evaluate INCEPT across ten downstream datasets organized into three domains of post-acquisition EEG analysis: signal-level EEG assessment, brain-state decoding, and brain-health evaluation.
These datasets cover abnormal EEG detection, artifact-type recognition, emotion recognition, motor imagery, sleep staging, mental-stress recognition, depression-related assessment, neurodegenerative disease evaluation and seizure detection \cite{obeid2016temple,chen2023large,liu2021comparing,PhysioNet-eegmmidb-1.0.0,schalk2004bci2000,goldberger2000physiobank,khalighi2016isruc,zyma2019electroencephalograms,mumtaz2016mdd,mumtaz2018machine,ds004504:1.0.8,PhysioNet-siena-scalp-eeg-1.0.0,detti2020eeg}.
We test each foundation model under linear probing, which measures the readability of frozen representations, and full fine-tuning, which measures supervised adaptability.
Across this heterogeneous benchmark, INCEPT achieves the most consistent performance among recent EEG foundation models, while controlled ablations show that invariance-oriented pre-training improves transfer beyond reconstruction alone.
Together, these results support INCEPT as a transferable backbone for broad-spectrum EEG analysis.

\section{Results}
\label{sec:results}

The Results are organized around four questions:
\begin{itemize}
\item Section~\ref{sec:results-unified} defines the pre-training corpus, downstream benchmark and adaptation protocols.
\item Section~\ref{sec:results-quantitative} quantitatively evaluates transfer performance across heterogeneous EEG applications.
\item Section~\ref{sec:results-representation} examines neural representation structure across subject- and state-level EEG organization.
\item Section~\ref{sec:results-invariance} evaluates whether cross-view invariance learning improves transfer beyond masked reconstruction alone.
\end{itemize}

\begin{figure}[!t]
    \centering
    \begin{minipage}[t]{0.75\textwidth}
        \vspace{0pt}
        \overviewpanellabel{width=\textwidth}{a}{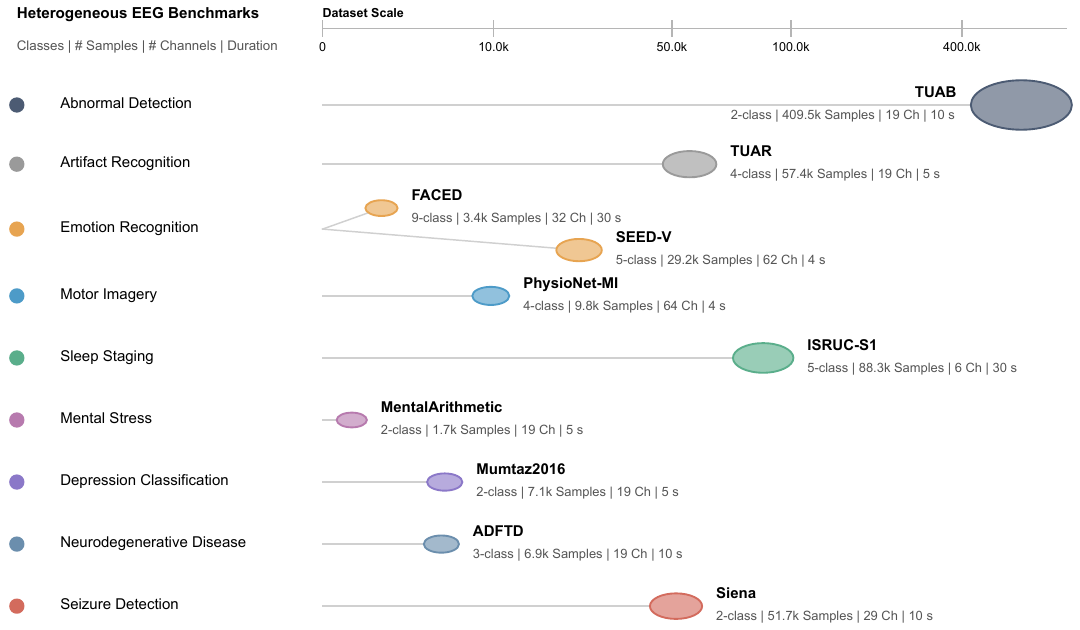}
    \end{minipage}
    \hfill
    \begin{minipage}[t]{0.24\textwidth}
        \vspace{2pt}
        \overviewpanellabel{width=\textwidth}{b}{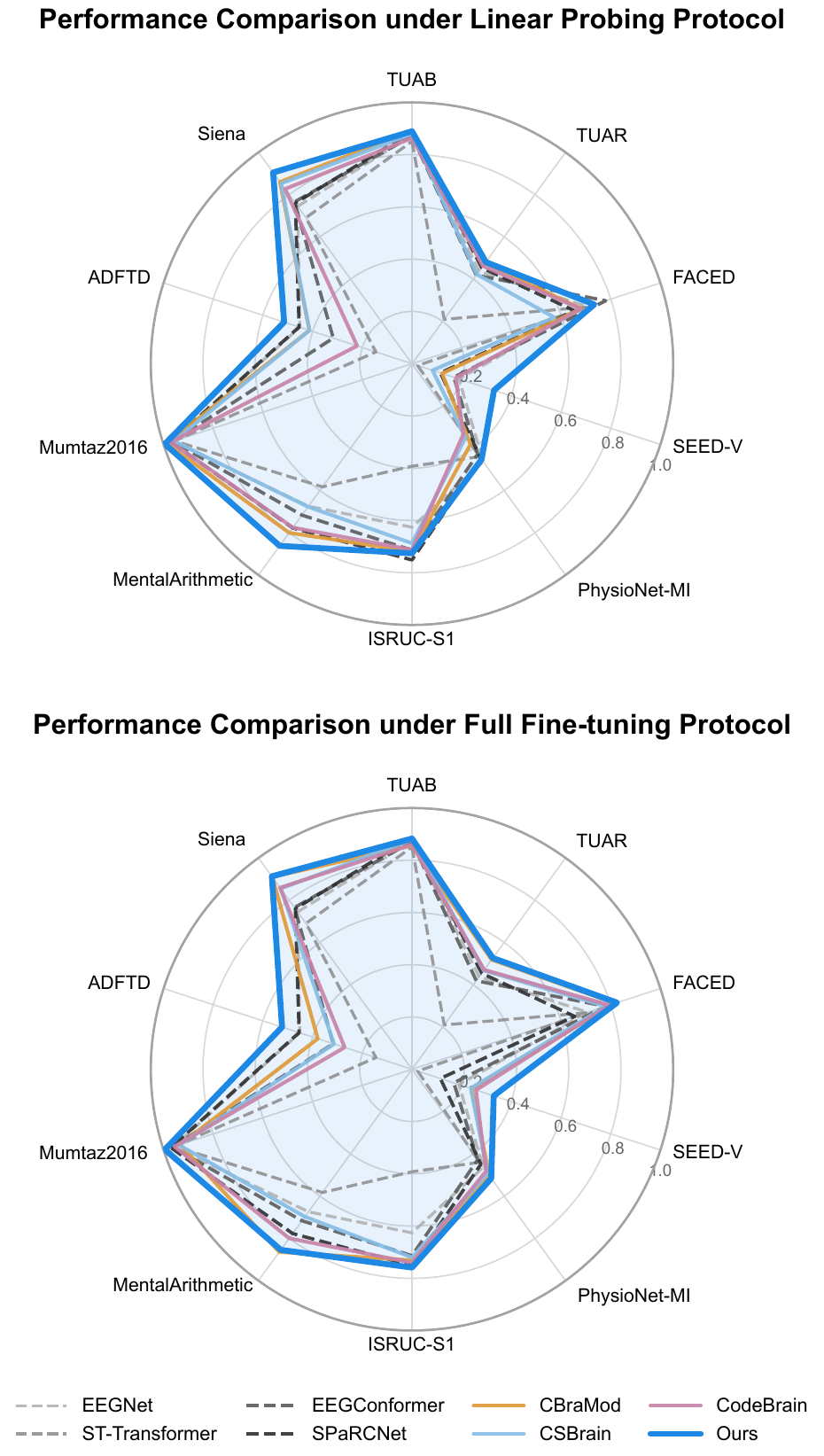}
    \end{minipage}
    \caption{\textbf{Heterogeneous downstream EEG benchmarks and transfer overview.}
    \textbf{a}, Downstream EEG benchmarks spanning signal-level EEG assessment, brain-state decoding, and brain-health evaluation, with varying sample scales, class numbers, channel counts and segment durations.
    \textbf{b}, Radar plots providing a compact overview of model transfer performance across the ten downstream EEG datasets under linear probing and full fine-tuning. 
    }
    \label{fig:dataset_overview}
\end{figure}

\begin{table*}[!h]
\centering
\caption{
\textbf{Dataset characteristics and cohort information for pre-training and downstream evaluation.}
\textbf{a}, Task and signal characteristics of the pre-training corpus and downstream EEG benchmarks, including original sampling rate, channel count, segment duration, sample scale, subject count and label space. Sampling rates refer to the original recordings before resampling in our preprocessing pipeline, and the TUEG sample scale denotes the amount of usable EEG retained for self-supervised pre-training after preprocessing and quality filtering.
\textbf{b}, Source country or region, acquisition scenario and available participant information for each dataset. Age is reported as mean $\pm$ s.d. when available, either by sex or by diagnostic group according to the original metadata, and sex distribution is reported as the percentage of male participants.
Dashed separators indicate the pre-training corpus, signal-level EEG assessment tasks, brain-state decoding tasks, and brain-health evaluation tasks.
}
\label{tab:dataset_summary}
\vspace{1.0em}
\begin{subtable}{\textwidth}
\phantomsubcaption\label{tab:supp_dataset_characteristics}
\centering
\tablepanellabel{a}{%
\begingroup
\setlength{\tabcolsep}{4pt}
\renewcommand{\arraystretch}{1.08}
\resizebox{\linewidth}{!}{
\begin{tabular}{llcccccc}
\toprule
Task & Dataset & Rate & \#Channels & Duration & \#Samples & \#Subjects & Label \\
\midrule
Self-supervised Pre-training & TUEG & Mostly 250 Hz & 19 retained & / & $\sim$11,000 h & 14,987 & Unlabeled \\
\hdashline
Abnormal EEG Detection & TUAB & Mostly 250 Hz & 19 & 10 s & 409,455 & 2,383 & 2-class \\
Artifact Recognition & TUAR & Mostly 250 Hz & 19 & 5 s & 57,392 & 213 & 4-class \\
\hdashline
\multirow{2}{*}{Emotion Recognition} & FACED & 250 Hz & 32 & 30 s & 3,444 & 123 & 9-class \\
 & SEED-V & 1,000 Hz & 62 & 4 s & 29,168 & 16 & 5-class \\
Motor Imagery Classification & PhysioNet-MI & 160 Hz & 64 & 4 s & 9,837 & 109 & 4-class \\
Sleep Staging & ISRUC-S1 & 200 Hz & 6 & 30 s & 88,340 & 100 & 5-class \\
\hdashline
Mental Stress Detection & MentalArithmetic & 500 Hz & 19 & 5 s & 1,707 & 36 & 2-class \\
Depression-related Classification & Mumtaz2016 & 256 Hz & 19 & 5 s & 7,143 & 64 & 2-class \\
Neurodegenerative Disease Classification & ADFTD & 500 Hz & 19 & 10 s & 6,948 & 88 & 3-class \\
Seizure Detection & Siena & 512 Hz & 29 & 10 s & 51,697 & 14 & 2-class \\
\bottomrule
\end{tabular}
}
\endgroup
}
\end{subtable}

\vspace{1.8em}

\begin{subtable}{\textwidth}
\phantomsubcaption\label{tab:supp_dataset_demographics}
\centering
\tablepanellabel{b}{%
\begingroup
\setlength{\tabcolsep}{4pt}
\renewcommand{\arraystretch}{1.08}
\resizebox{\linewidth}{!}{
\begin{tabular}{lllccc}
\toprule
Dataset & Country/region & Scenario & \#Subjects & Age (years) & Sex (male, \%) \\
\midrule
TUEG & United States & Clinical EEG Archive & 14,987 & Male: 49.3 $\pm$ 20.1; Female: 50.1 $\pm$ 20.5 & 48.8 \\
\hdashline
TUAB & United States & Clinical EEG Archive & 2,383 & / & / \\
TUAR & United States & Clinical EEG Archive & 213 & / & / \\
\hdashline
FACED & China & Video-elicited Emotion Recognition & 123 & Male: 23.5 $\pm$ 5.1; Female: 23.0 $\pm$ 4.4 & 39.0 \\
SEED-V & China & Video-elicited Emotion Recognition & 16 & Male: 21.2 $\pm$ 1.7; Female: 21.9 $\pm$ 1.6 & 37.5 \\
PhysioNet-MI & United States & Motor Execution/Imagery BCI Task & 109 & / & / \\
ISRUC-S1 & Portugal & Sleep Clinical Cohort & 100 & Male: 52.3 $\pm$ 17.4; Female: 49.6 $\pm$ 14.1 & 56.0 \\
\hdashline
MentalArithmetic & Ukraine & Resting-state and Mental Arithmetic Task & 36 & Male: 19.7 $\pm$ 3.5; Female: 17.8 $\pm$ 1.3 & 25.0 \\
Mumtaz2016 & Malaysia & Depression-related Resting-state EEG & 64 & MDD: 40.3 $\pm$ 12.9; HC: 38.3 $\pm$ 15.6 & 59.4 \\
ADFTD & Greece & Neurodegenerative Disease EEG & 88 & AD: 66.4 $\pm$ 7.9; FTD: 63.6 $\pm$ 8.2; HC: 67.9 $\pm$ 5.4 & 50.0 \\
Siena & Italy & Epilepsy Monitoring & 14 & Male: 47.3 $\pm$ 13.8; Female: 38.5 $\pm$ 14.4 & 57.1 \\
\bottomrule
\end{tabular}
}
\endgroup
}
\end{subtable}
\end{table*}

\subsection{A unified EEG foundation model for transferable neural decoding}
\label{sec:results-unified}

This section establishes the evaluation setting:
\begin{itemize}
\item Section~\ref{sec:results-pretraining-benchmark} introduces the clinical EEG pre-training corpus and downstream evaluation across signal-level assessment, brain-state decoding, and brain-health evaluation.
\item Section~\ref{sec:results-protocols} defines two adaptation protocols that measure frozen readability and supervised adaptability.
\end{itemize}

\subsubsection{Large-scale clinical EEG pre-training and heterogeneous downstream evaluation}
\label{sec:results-pretraining-benchmark}
We use INCEPT to test whether a single EEG foundation encoder can support a broad EEG analysis workflow rather than improve one isolated benchmark. INCEPT is self-supervised pre-trained on the Temple University Hospital EEG Corpus (TUEG) \cite{obeid2016temple}, a large archive of routine clinical EEG recordings. After unified preprocessing, artifact control and quality filtering, approximately 11,000 hours of usable unlabeled EEG are retained for pre-training. These recordings are based primarily on the international 10-20 system and contain substantial variation in subjects, recording sessions and clinical acquisition conditions, allowing the model to learn reusable EEG representations before any downstream task-specific labels are introduced.

We organize downstream evaluation as a heterogeneous EEG application landscape rather than a collection of isolated benchmarks. This landscape covers three complementary levels of EEG use. At the signal and recording level, TUAB \cite{obeid2016temple} is used for abnormal EEG screening from routine clinical recordings, and TUAR \cite{obeid2016temple} is used for signal-quality assessment by distinguishing common EEG artifacts, including eye-movement, muscle and electrode-related artifacts, from background EEG segments. At the brain-state level, FACED \cite{chen2023large}, SEED-V \cite{liu2021comparing}, PhysioNet-MI \cite{PhysioNet-eegmmidb-1.0.0,schalk2004bci2000,goldberger2000physiobank} and ISRUC-S1 \cite{khalighi2016isruc} are used to evaluate affective state, sensorimotor state and sleep state decoding. At the brain-health level, MentalArithmetic \cite{zyma2019electroencephalograms}, Mumtaz2016 \cite{mumtaz2016mdd,mumtaz2018machine}, ADFTD \cite{ds004504:1.0.8} and Siena \cite{PhysioNet-siena-scalp-eeg-1.0.0,detti2020eeg,goldberger2000physiobank} are used to assess mental-stress or cognitive-demand states, depression-related EEG differences, neurodegenerative disease and seizure-related activity. Together, these levels cover a broad EEG application space, from assessing the recording itself, to decoding transient neural states, to evaluating clinically relevant brain function. Because the datasets within and across these levels differ in label semantics, sample scale, channel density, montage and segment duration, the benchmark tests whether one pre-trained encoder can remain useful across diverse EEG applications rather than only within a single task category or acquisition setting.

Fig.~\ref{fig:dataset_overview} summarizes this downstream evaluation landscape by showing how the ten datasets differ in task type, sample scale, channel density, segment duration and transfer performance under the two adaptation protocols. Table~\ref{tab:dataset_summary} complements this overview with dataset-level signal characteristics and available demographic information. Together, they define the heterogeneous benchmark used to test whether a single pre-trained EEG encoder can support broad-spectrum EEG analysis across different application levels, acquisition settings and subject populations.

\subsubsection{Evaluation protocols for frozen readability and supervised adaptability}
\label{sec:results-protocols}

We evaluate EEG foundation models under two supervised downstream adaptation protocols. Linear probing freezes the pre-trained backbone and trains only a lightweight classifier with labelled downstream data. The labels supervise the readout layer, but they do not update the backbone or reshape the feature extractor. Linear probing therefore tests frozen readability, namely whether task-relevant information can be extracted from the learned representation space by a simple supervised classifier without backbone adaptation.

Full fine-tuning provides a complementary test. In this protocol, both the pre-trained backbone and the downstream classifier are updated with labelled downstream data. It tests supervised adaptability, namely whether the same pre-trained representation can be reshaped when the target task differs from the pre-training corpus in electrode layout, label space, brain-state context or cohort composition. We interpret these two protocols together throughout the results. A useful EEG foundation encoder should provide representations that are readable when frozen and adaptable when downstream supervision is available.

Binary tasks are evaluated using balanced accuracy, AUROC and AUC-PR, while multi-class tasks are evaluated using balanced accuracy, weighted F1 score and Cohen's kappa. Each experiment is repeated over five random seeds, and results are reported as mean $\pm$ standard deviation, with the standard deviation used to summarize seed-to-seed stability.

\subsection{Quantitative analysis across heterogeneous EEG applications}
\label{sec:results-quantitative}

This section quantitatively reports downstream performance in four steps:
\begin{itemize}
\item Section~\ref{sec:results-overall} summarizes the transfer profile overall across the full downstream benchmark.
\item Section~\ref{sec:results-signal} evaluates signal-level EEG quality assessment, including clinical abnormality screening and artifact recognition.
\item Section~\ref{sec:results-brain-state} evaluates brain-state decoding across affective, sensorimotor and sleep-state tasks.
\item Section~\ref{sec:results-brain-health} evaluates brain-health assessment across psychiatric, cognitive and neurological conditions.
\end{itemize}

\subsubsection{Overall performance across downstream EEG tasks}
\label{sec:results-overall}
The radar summary (Fig.~\ref{fig:dataset_overview}b) shows that no task-specific supervised encoder is uniformly strong across the full EEG workflow. These models remain effective when the downstream dataset aligns with their architectural bias. SPaRCNet \cite{jing2023development}, for example, is competitive on several clinical or diagnostic tasks, including TUAB abnormal EEG detection (Fig.~\ref{fig:result_bar_signal}a; Table~\ref{table:result_signal}), ISRUC-S1 sleep staging (Fig.~\ref{fig:result_bar_state}d; Table~\ref{table:result_state}) and ADFTD neurodegenerative disease classification (Fig.~\ref{fig:result_bar_health}c; Table~\ref{table:result_health}). EEGConformer \cite{song2022eeg} performs strongly on FACED emotion recognition (Fig.~\ref{fig:result_bar_state}a; Table~\ref{table:result_state}). However, these advantages are not stable across tasks: SPaRCNet declines on affective EEG datasets such as FACED and SEED-V (Fig.~\ref{fig:result_bar_state}a,b; Table~\ref{table:result_state}), whereas EEGConformer is less competitive on ADFTD (Fig.~\ref{fig:result_bar_health}c; Table~\ref{table:result_health}). Thus, supervised encoders can be strong closed-task baselines, but their rankings remain dataset- and montage-dependent.

EEG foundation models change the comparison by replacing one-dataset-one-model training with representation reuse. Across the evaluated EFMs, linear probing provides a strict test of whether labelled downstream classifiers can read useful information from a frozen backbone, whereas full fine-tuning tests whether the same backbone can adapt to new task and montage settings. Existing reconstruction-centered EFMs improve reuse in several datasets, but their performance remains uneven across the heterogeneous evaluation. INCEPT shows a more consistent profile across signal-level assessment, brain-state decoding, and brain-health evaluation. Under linear probing, it ranks first or remains highly competitive on TUAB, TUAR, ADFTD, Mumtaz2016, MentalArithmetic, ISRUC-S1 and FACED. Under full fine-tuning, it is strongest across all three metrics on TUAR, FACED, SEED-V, PhysioNet-MI, ISRUC-S1 and ADFTD, showing broad gains when downstream labels are allowed to update the backbone. Detailed numerical comparisons, including relative gains and seed-to-seed variability, are provided in Supplementary Section~\ref{supplement:detailed_downstream}. These comparisons show the empirical advantage of INCEPT, while the objective ablations below test whether this advantage arises from the proposed invariance-oriented pre-training paradigm rather than from foundation pre-training alone (Section~\ref{sec:objective_ablation}).

\subsubsection{EEG abnormality and artifact assessment for signal quality control}
\label{sec:results-signal}

This signal-level analysis contains two tasks:
\begin{itemize}
\item Clinical abnormality assessment is evaluated using routine EEG recordings from TUAB.
\item Artifact-type recognition for EEG signal quality control is evaluated using TUAR.
\end{itemize}

\begin{figure*}[!t]
    \centering
    \setlength{\tabcolsep}{2pt}
    \renewcommand{\arraystretch}{1}
    \begin{tabular}{@{}ccc@{}}
    \imgpanellabel{0.32\textwidth}{a}{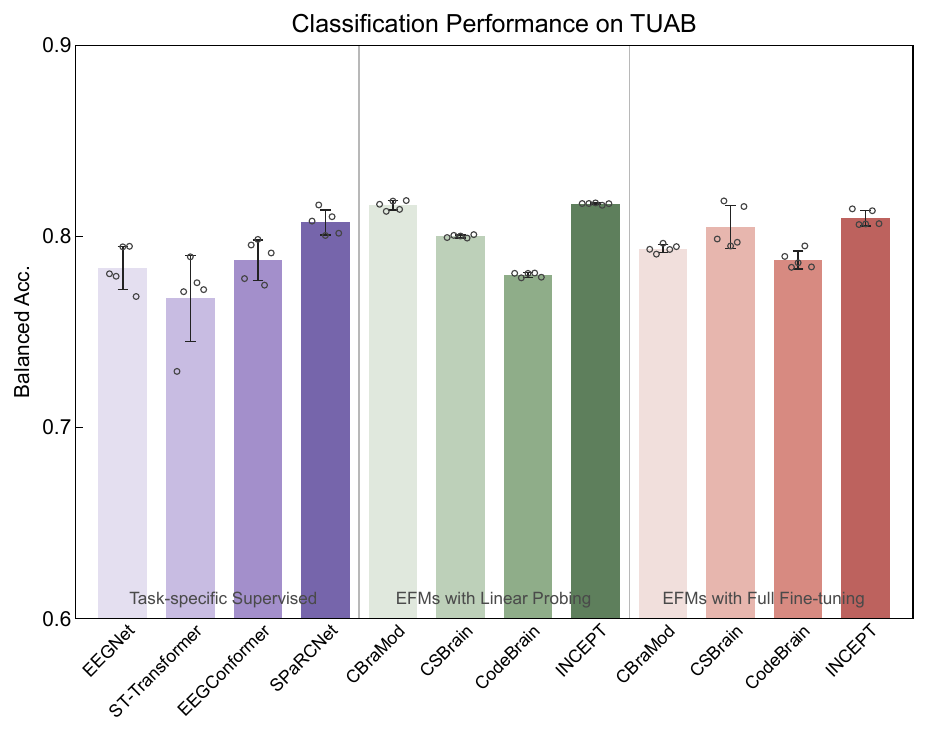}
    &
    \includegraphics[width=0.32\textwidth]{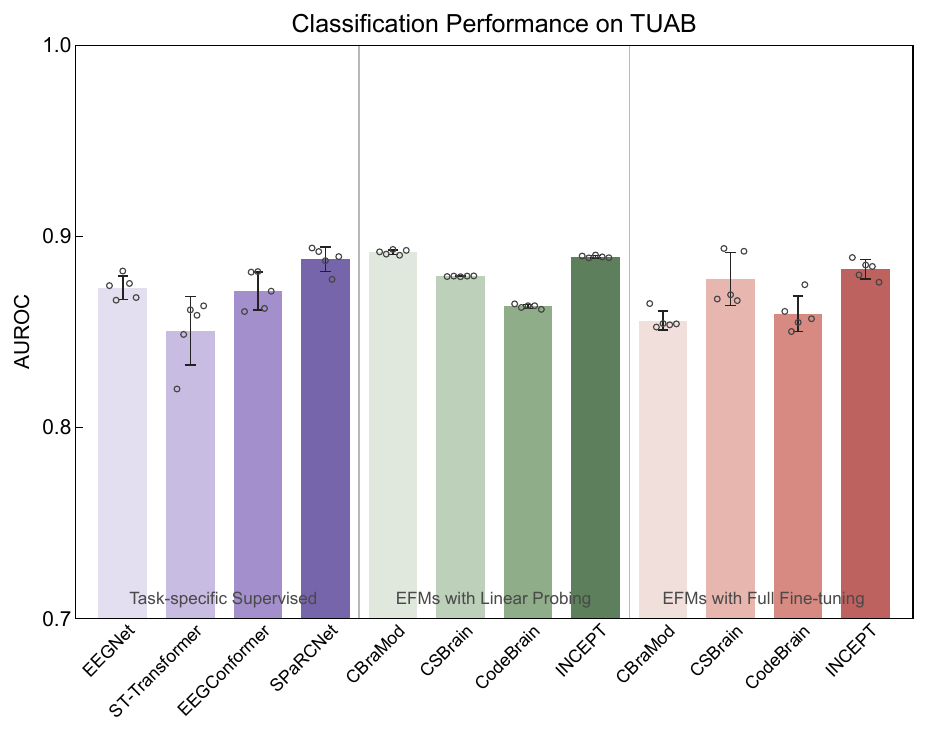}
    &
    \includegraphics[width=0.32\textwidth]{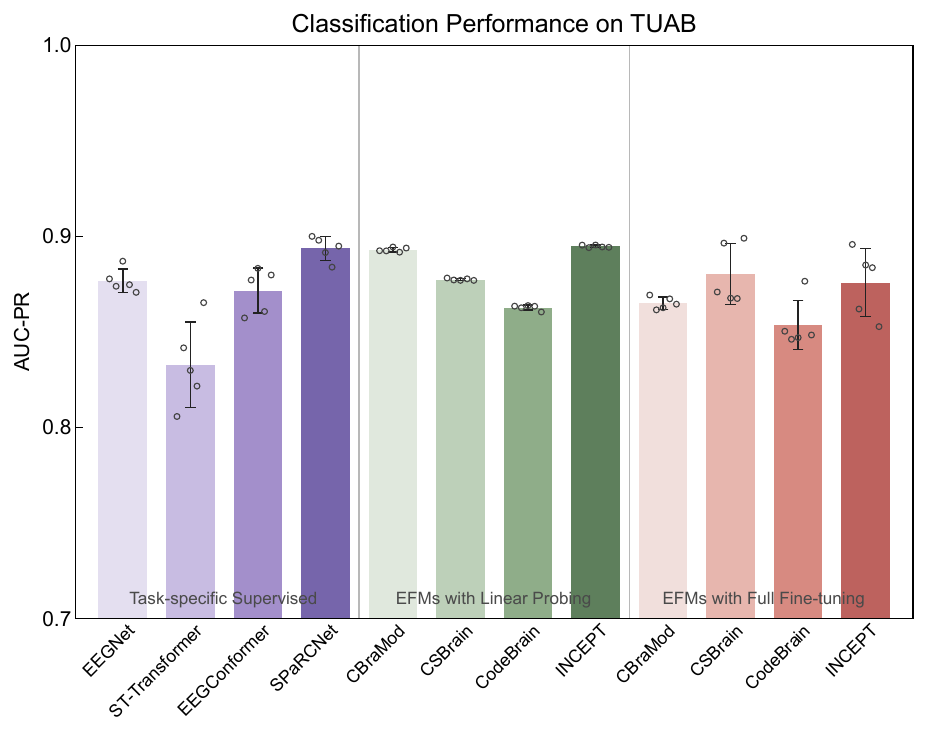}
    \\[-0.45em]
    \imgpanellabel{0.32\textwidth}{b}{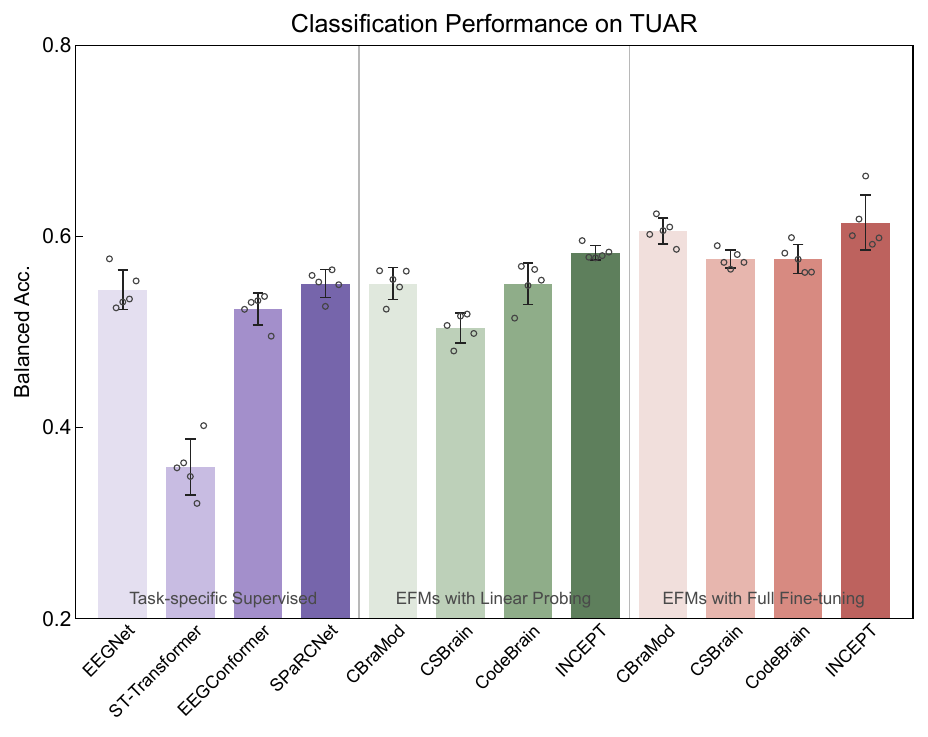}
    &
    \includegraphics[width=0.32\textwidth]{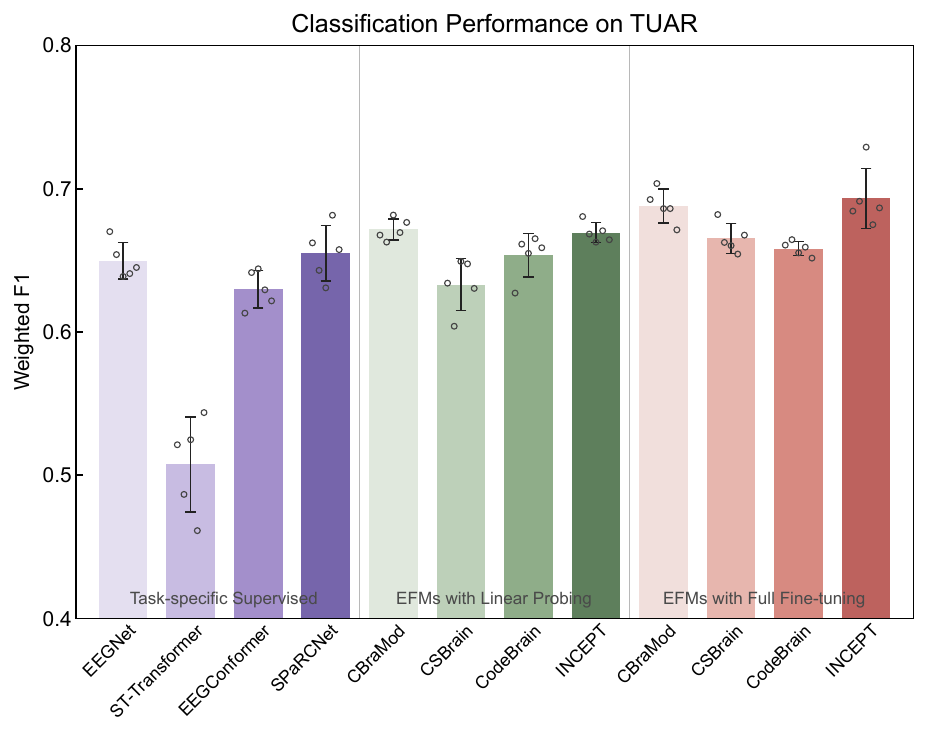}
    &
    \includegraphics[width=0.32\textwidth]{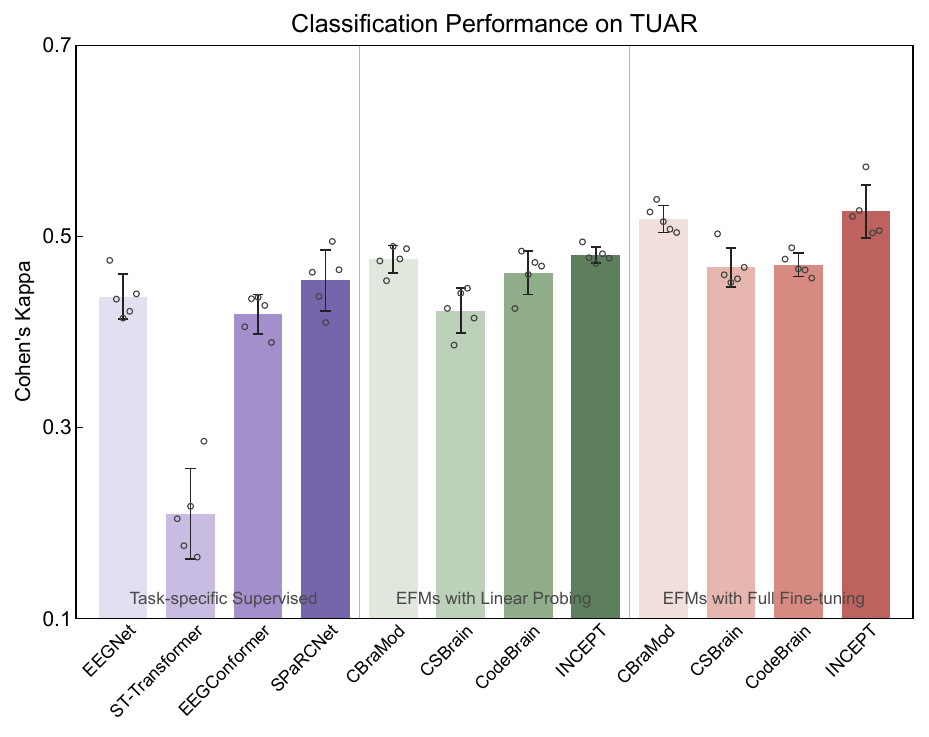}
    \end{tabular}
    \caption{
        \textbf{Signal-level EEG quality assessment performance across three evaluation metrics.}
        \textbf{a}, TUAB abnormal EEG detection using balanced accuracy, AUROC and AUC-PR.
        \textbf{b}, TUAR artifact-type recognition using balanced accuracy, weighted F1 score and Cohen's kappa.
        Bars indicate mean~$\pm$~s.d. over five random seeds, with hollow circles denoting individual seeds.
    }
    \label{fig:result_bar_signal}
\end{figure*}

\paragraph{(i) Clinical abnormality assessment from routine EEG}
In routine clinical EEG, signal-level assessment begins with abnormality screening: the model asks whether a recording deviates from normal background activity. TUAB \cite{obeid2016temple} therefore tests whether a reusable encoder can support clinical signal triage rather than a narrow event detector. Fig.~\ref{fig:result_bar_signal}a and Table~\ref{table:result_signal} show that task-specific supervised encoders are competitive in mean performance, with SPaRCNet reaching 80.74\% balanced accuracy. However, foundation models provide a more stable abnormality-assessment representation. Under linear probing, INCEPT reaches the highest balanced accuracy (81.71\%) and AUC-PR (89.48\%), and it is also the second-highest model on AUROC (88.94\%). Compared with SPaRCNet, this gives a 1.2\% balanced-accuracy gain while keeping AUROC and AUC-PR at the top tier. Stability improves more clearly: the average seed-to-seed standard deviation of balanced accuracy for linear-probing EFMs is reduced by 88.0\% relative to task-specific encoders.

TUAB also illustrates a broader pattern for montage-matched clinical EEG: full fine-tuning is not always beneficial when the frozen representation is already aligned with the downstream label space. INCEPT decreases from 81.71\% to 80.95\% balanced accuracy after full fine-tuning, and CBraMod shows a similar drop. Even after this decrease, INCEPT still remains the strongest fine-tuned model for balanced accuracy and AUROC. Thus, TUAB is mainly a frozen-readout result: high balanced accuracy, top-tier AUROC/AUC-PR and substantially improved stability, rather than a large fine-tuning gain.

\begin{table*}[!h]
    \centering
    \caption{\textbf{Signal-level EEG quality assessment performance on TUAB and TUAR.}}
    \label{table:result_signal}
    \small
    \setlength{\tabcolsep}{3.5pt}
    \renewcommand{\arraystretch}{1.08}
    \begin{tabularx}{\textwidth}{l *{6}{Y}}
        \toprule
        \multirow{2}{*}{Method} & \multicolumn{3}{c}{TUAB (2-class)} & \multicolumn{3}{c}{TUAR (4-class)} \\
        \cmidrule(lr){2-4} \cmidrule(lr){5-7}
        & Bal. Acc. & AUROC & AUC-PR & Bal. Acc. & Weighted F1 & Kappa \\
        \midrule
        \rowcolor{gray!15} \multicolumn{7}{c}{\textit{Task-specific Supervised Encoders}} \\
        EEGNet & 78.35$_{\pm 1.00\phantom{0}}$ & 87.33$_{\pm 0.55\phantom{0}}$ & 87.69$_{\pm 0.56\phantom{0}}$ & 54.42$_{\pm 1.87\phantom{0}}$ & 64.97$_{\pm 1.14\phantom{0}}$ & 43.71$_{\pm 2.10\phantom{0}}$ \\
        ST-Transformer & 76.76$_{\pm 2.02\phantom{0}}$ & 85.06$_{\pm 1.61\phantom{0}}$ & 83.29$_{\pm 2.00\phantom{0}}$ & 35.84$_{\pm 2.62\phantom{0}}$ & 50.76$_{\pm 2.96\phantom{0}}$ & 20.96$_{\pm 4.25\phantom{0}}$ \\
        EEGConformer & 78.76$_{\pm 0.96\phantom{0}}$ & 87.15$_{\pm 0.89\phantom{0}}$ & 87.17$_{\pm 1.06\phantom{0}}$ & 52.41$_{\pm 1.49\phantom{0}}$ & 63.00$_{\pm 1.17\phantom{0}}$ & 41.87$_{\pm 1.85\phantom{0}}$ \\
        SPaRCNet & 80.74$_{\pm 0.59\phantom{0}}$ & 88.80$_{\pm 0.57\phantom{0}}$ & 89.37$_{\pm 0.57\phantom{0}}$ & 55.06$_{\pm 1.31\phantom{0}}$ & 65.50$_{\pm 1.73\phantom{0}}$ & 45.39$_{\pm 2.86\phantom{0}}$ \\
        \midrule
        \rowcolor{gray!15} \multicolumn{7}{c}{\textit{EFMs with Linear Probing}} \\
        CBraMod & \underline{81.63}$_{\pm 0.23\phantom{0}}$ & \textbf{89.18}$_{\pm 0.12\phantom{0}}$ & \underline{89.30}$_{\pm 0.10\phantom{0}}$ & \underline{55.07}$_{\pm 1.48\phantom{0}}$ & \textbf{67.16}$_{\pm 0.67\phantom{0}}$ & \underline{47.62}$_{\pm 1.27\phantom{0}}$ \\
        CSBrain & 80.01$_{\pm 0.07\phantom{0}}$ & 87.92$_{\pm 0.02\phantom{0}}$ & 87.75$_{\pm 0.05\phantom{0}}$ & 50.41$_{\pm 1.40\phantom{0}}$ & 63.31$_{\pm 1.63\phantom{0}}$ & 42.24$_{\pm 2.12\phantom{0}}$ \\
        CodeBrain & 77.99$_{\pm 0.11\phantom{0}}$ & 86.34$_{\pm 0.10\phantom{0}}$ & 86.28$_{\pm 0.12\phantom{0}}$ & 55.03$_{\pm 1.93\phantom{0}}$ & 65.34$_{\pm 1.36\phantom{0}}$ & 46.22$_{\pm 2.04\phantom{0}}$ \\
        Ours & \textbf{81.71}$_{\pm 0.04\phantom{0}}$ & \underline{88.94}$_{\pm 0.06\phantom{0}}$ & \textbf{89.48}$_{\pm 0.06\phantom{0}}$ & \textbf{58.30}$_{\pm 0.67\phantom{0}}$ & \underline{66.93}$_{\pm 0.63\phantom{0}}$ & \textbf{48.06}$_{\pm 0.75\phantom{0}}$ \\
        \midrule
        \rowcolor{gray!15} \multicolumn{7}{c}{\textit{EFMs with Full Fine-tuning}} \\
        CBraMod & 79.37$_{\pm 0.19\phantom{0}}$ & 85.60$_{\pm 0.45\phantom{0}}$ & 86.51$_{\pm 0.29\phantom{0}}$ & \underline{60.57}$_{\pm 1.20\phantom{0}}$ & \underline{68.79}$_{\pm 1.05\phantom{0}}$ & \underline{51.84}$_{\pm 1.26\phantom{0}}$ \\
        CSBrain & \underline{80.50}$_{\pm 1.00\phantom{0}}$ & \underline{87.78}$_{\pm 1.24\phantom{0}}$ & \textbf{88.03}$_{\pm 1.43\phantom{0}}$ & 57.66$_{\pm 0.84\phantom{0}}$ & 66.53$_{\pm 0.93\phantom{0}}$ & 46.75$_{\pm 1.84\phantom{0}}$ \\
        CodeBrain & 78.77$_{\pm 0.42\phantom{0}}$ & 85.95$_{\pm 0.83\phantom{0}}$ & 85.37$_{\pm 1.15\phantom{0}}$ & 57.65$_{\pm 1.36\phantom{0}}$ & 65.82$_{\pm 0.44\phantom{0}}$ & 47.04$_{\pm 1.09\phantom{0}}$ \\
        Ours & \textbf{80.95}$_{\pm 0.37\phantom{0}}$ & \textbf{88.28}$_{\pm 0.45\phantom{0}}$ & \underline{87.59}$_{\pm 1.59\phantom{0}}$ & \textbf{61.45}$_{\pm 2.59\phantom{0}}$ & \textbf{69.32}$_{\pm 1.87\phantom{0}}$ & \textbf{52.62}$_{\pm 2.50\phantom{0}}$ \\
        \bottomrule
    \end{tabularx}
    \vspace{0.35em}
    \begin{minipage}{\textwidth}
    \footnotesize
    Values are reported as mean $\pm$ s.d. across five runs with different random seeds. Results are shown for task-specific supervised encoders trained from scratch and EEG foundation models evaluated under linear probing with frozen backbones or full fine-tuning. Best and second-best results within each evaluation protocol and dataset are shown in bold and underlined, respectively.
    \end{minipage}
\end{table*}

\paragraph{(ii) Artifact-type recognition for EEG quality control}
Explicit artifact recognition extends signal quality control from broad abnormality screening to direct identification of non-neural contamination. TUAR asks whether the representation can distinguish eye movement, muscle artifact, electrode-related artifact and background EEG segments. Fig.~\ref{fig:result_bar_signal}b and Table~\ref{table:result_signal} show a clearer role for supervised adaptation than on TUAB. With a frozen backbone, INCEPT obtains the strongest balanced accuracy and kappa among foundation models (58.30\% and 48.06\%), improving over the average task-specific encoder by 17.9\% and 26.5\%, respectively. Its weighted F1 is also close to the best frozen result, and the seed-to-seed standard deviations of balanced accuracy, weighted F1 and kappa are reduced by 63.5\%, 64.0\% and 72.7\% relative to the task-specific encoder average. Full fine-tuning raises INCEPT to 61.45\% balanced accuracy, 69.32\% weighted F1 and 52.62\% kappa, corresponding to gains of 5.4\%, 3.6\% and 9.5\% over its own linear-probing result. Unlike TUAB, TUAR therefore gains more from backbone adaptation, especially in balanced accuracy and kappa.

\vspace{0.5em}
\noindent\textbf{\textit{Together,}} TUAB and TUAR show that foundation pre-training supports signal-level EEG assessment at two complementary levels: stable clinical abnormality screening and more fine-grained artifact recognition. The former emphasizes frozen readability and stability, whereas the latter shows that task labels can further sharpen artifact-specific boundaries. More detailed quantitative comparisons and dataset-specific interpretations are reported in Supplementary Section~\ref{supplement:detailed_downstream}.

\subsubsection{Brain-state analysis across affective, sensorimotor and sleep states}
\label{sec:results-brain-state}

This brain-state analysis covers three representative decoding settings:
\begin{itemize}
\item Affective-state decoding is evaluated under different emotion EEG settings using FACED and SEED-V.
\item Sensorimotor-state decoding is evaluated from high-density motor-imagery EEG using PhysioNet-MI.
\item Sleep-state staging is evaluated from sparse clinical sleep EEG using ISRUC-S1.
\end{itemize}

\begin{figure*}[!b]
    \centering
    \setlength{\tabcolsep}{2pt}
    \renewcommand{\arraystretch}{1}
    \begin{tabular}{@{}ccc@{}}
    \imgpanellabel{0.32\textwidth}{a}{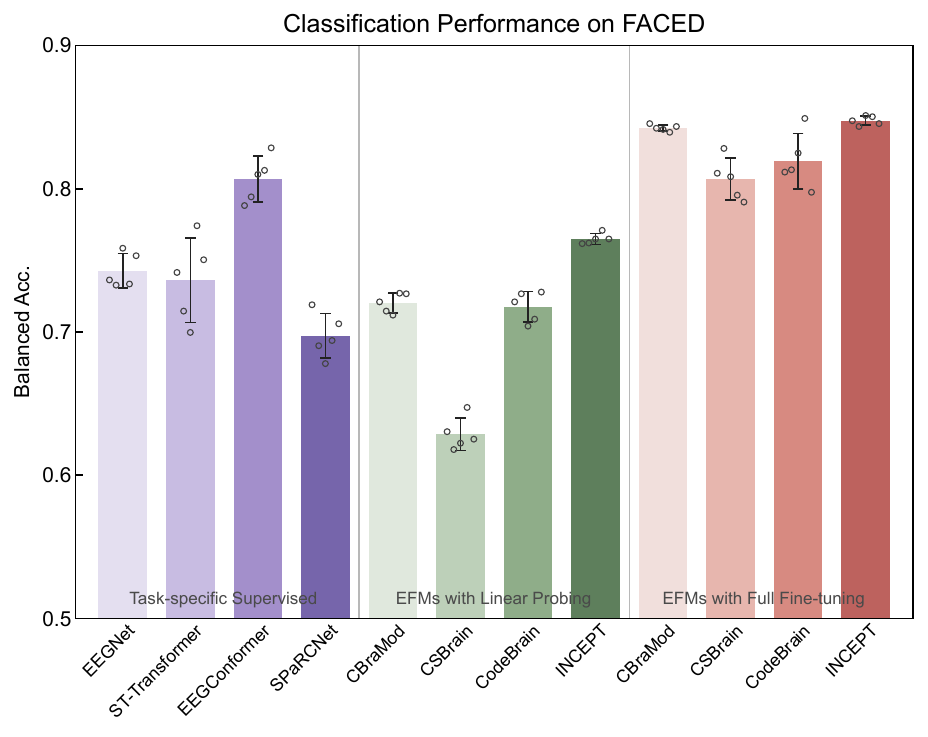} &
    \includegraphics[width=0.32\textwidth]{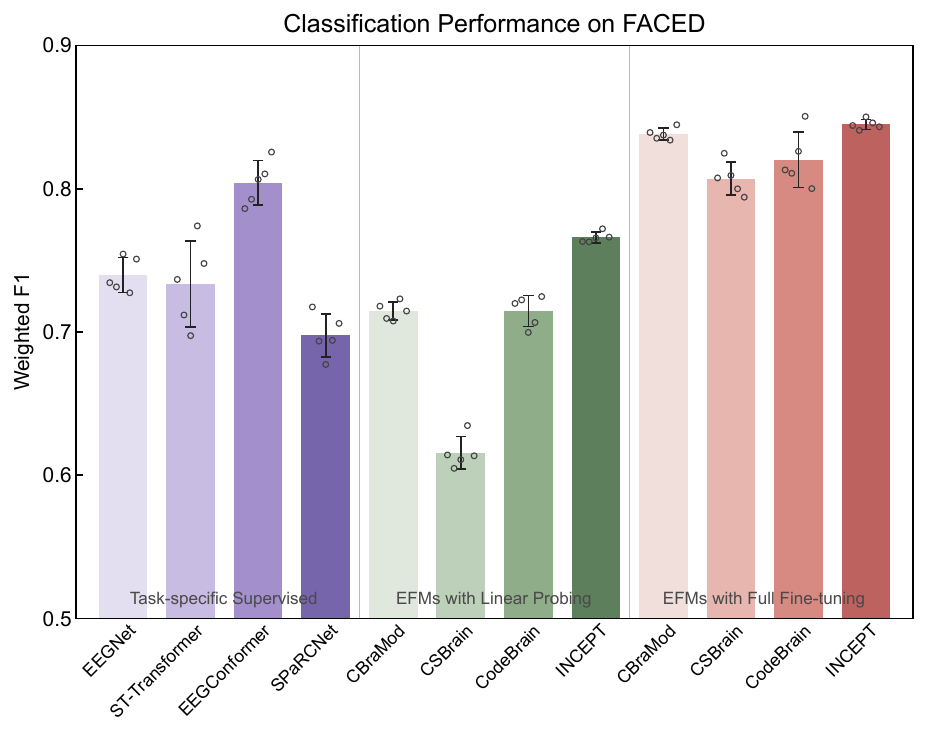} &
    \includegraphics[width=0.32\textwidth]{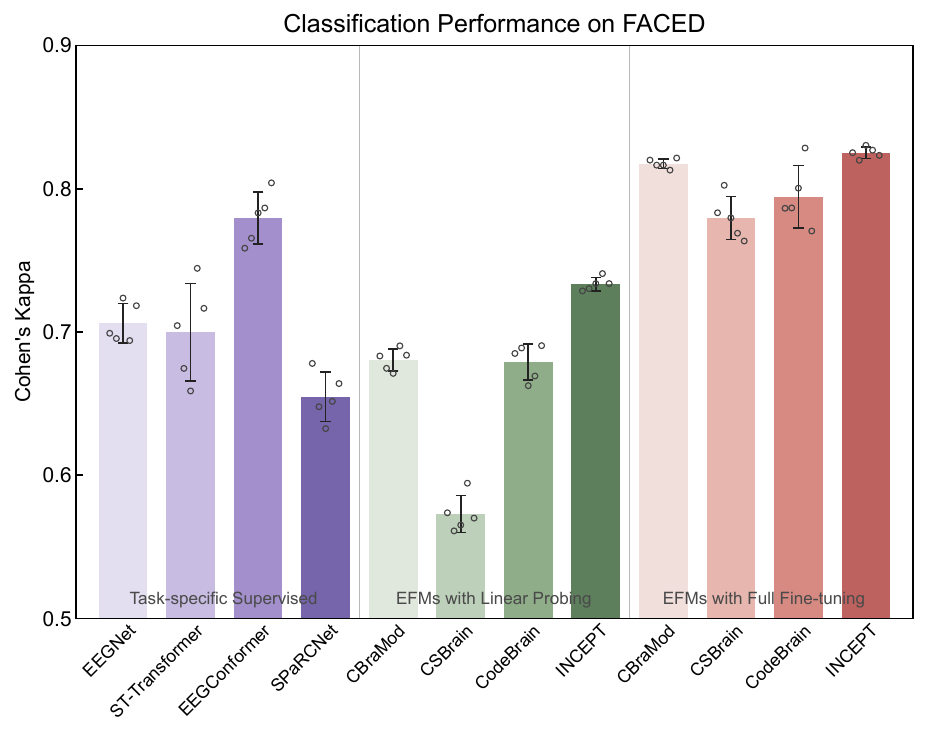} \\[-0.45em]
    \imgpanellabel{0.32\textwidth}{b}{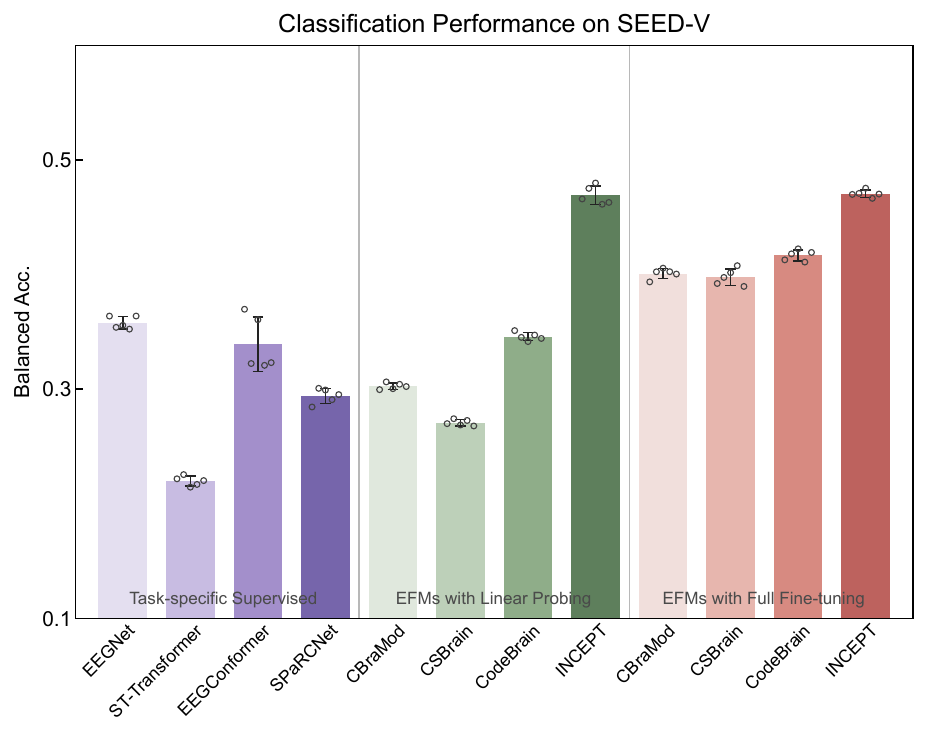} &
    \includegraphics[width=0.32\textwidth]{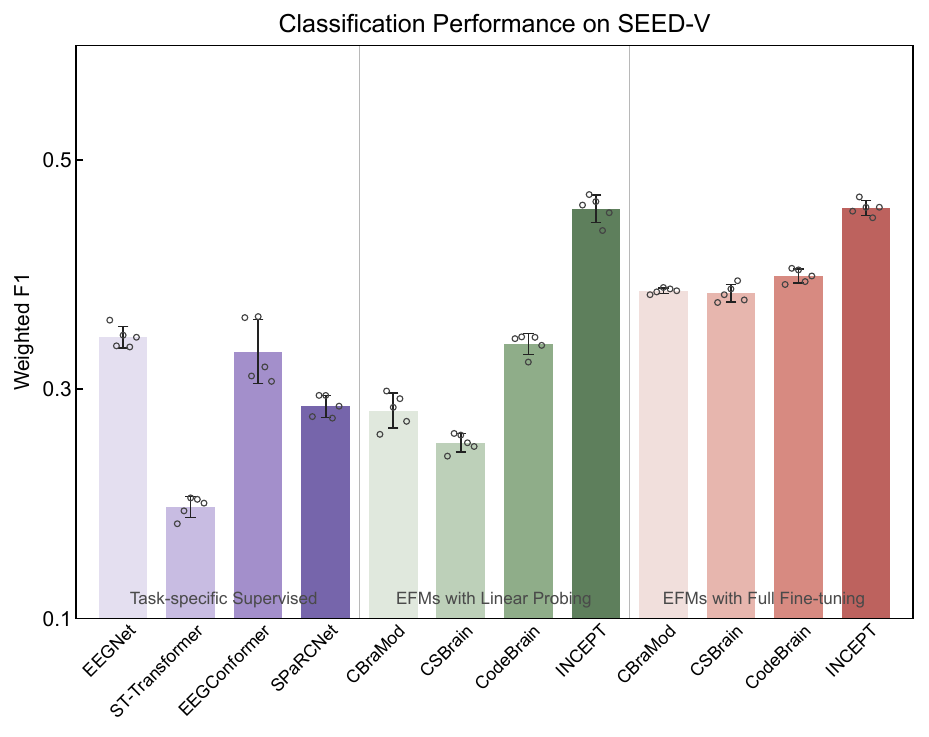} &
    \includegraphics[width=0.32\textwidth]{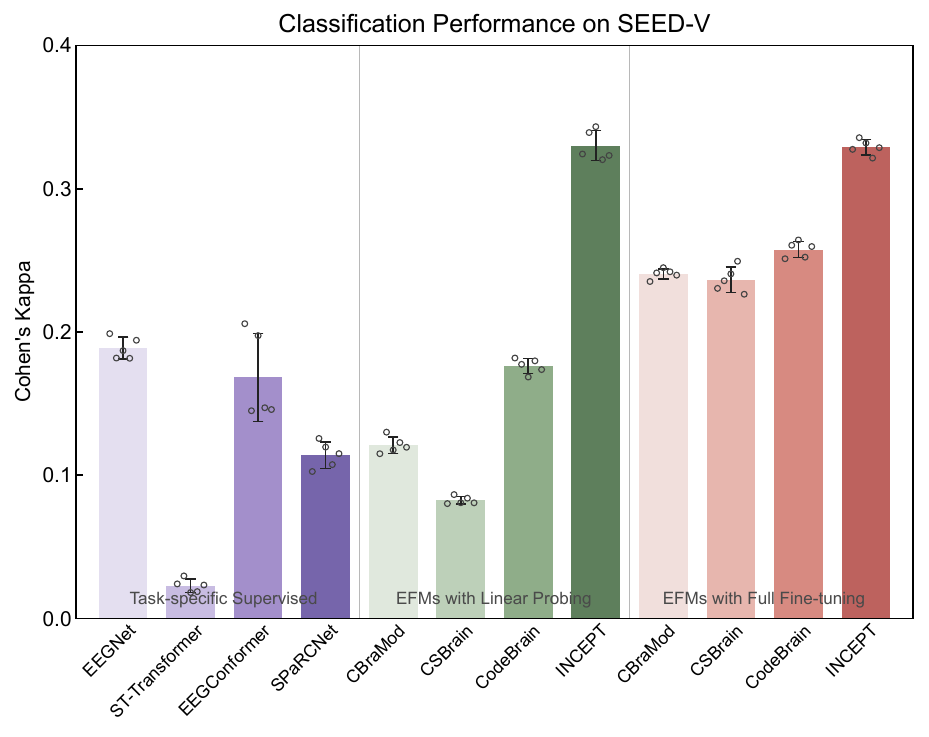} \\[-0.45em]
    \imgpanellabel{0.32\textwidth}{c}{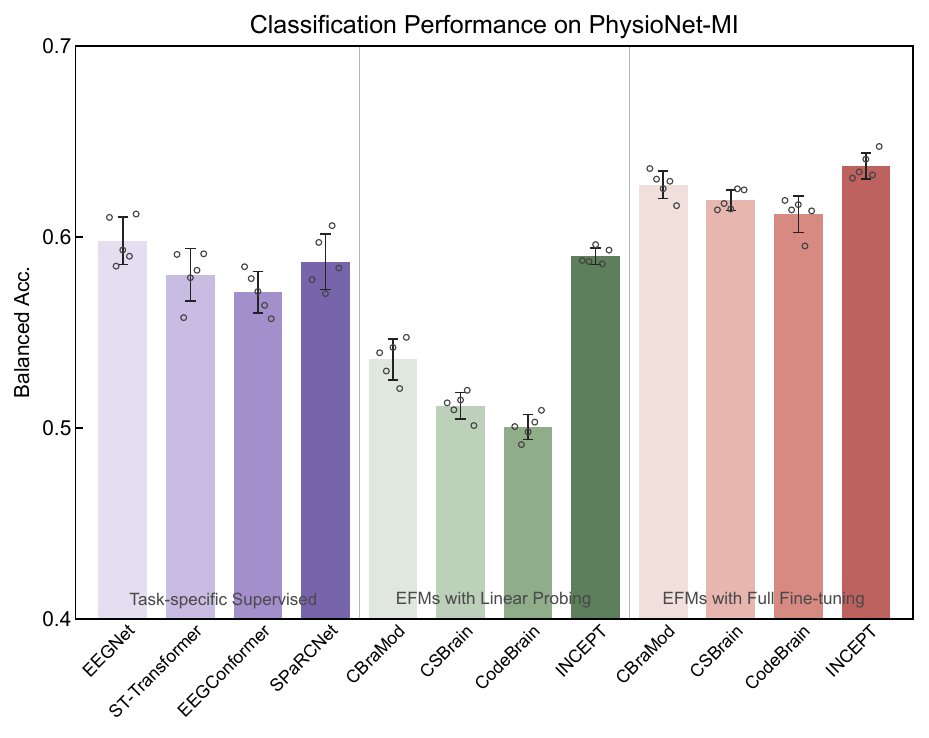} &
    \includegraphics[width=0.32\textwidth]{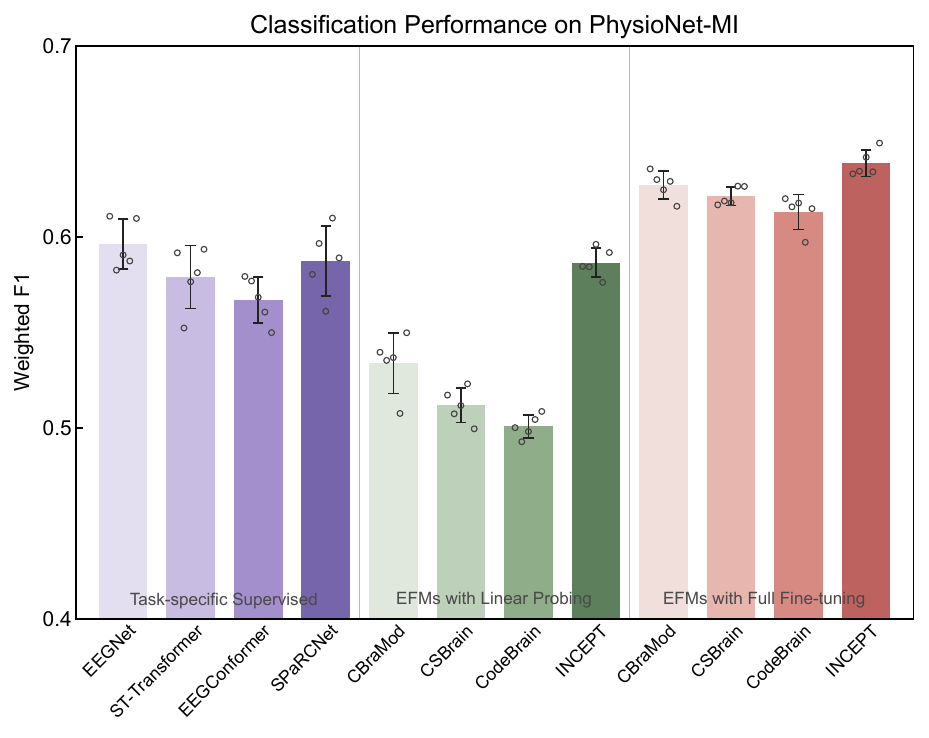} &
    \includegraphics[width=0.32\textwidth]{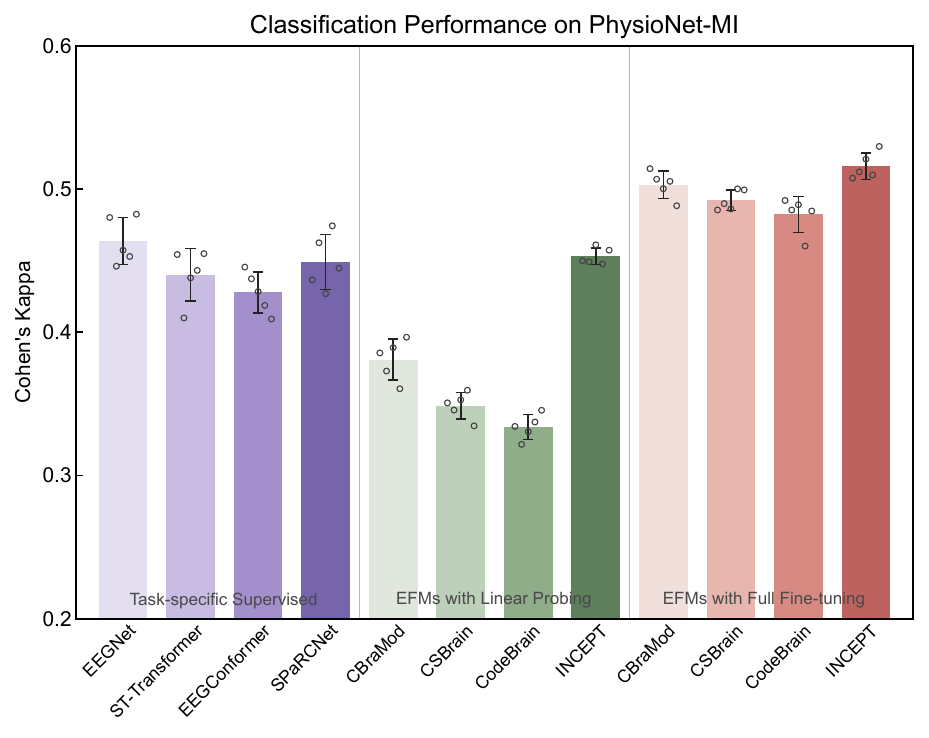} \\[-0.45em]
    \imgpanellabel{0.32\textwidth}{d}{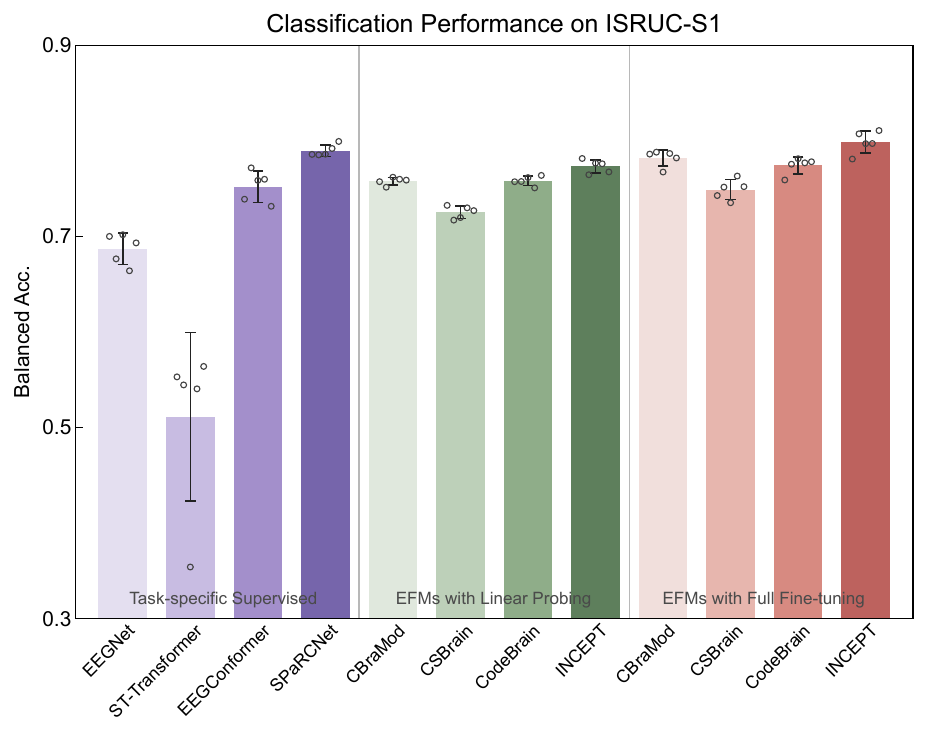} &
    \includegraphics[width=0.32\textwidth]{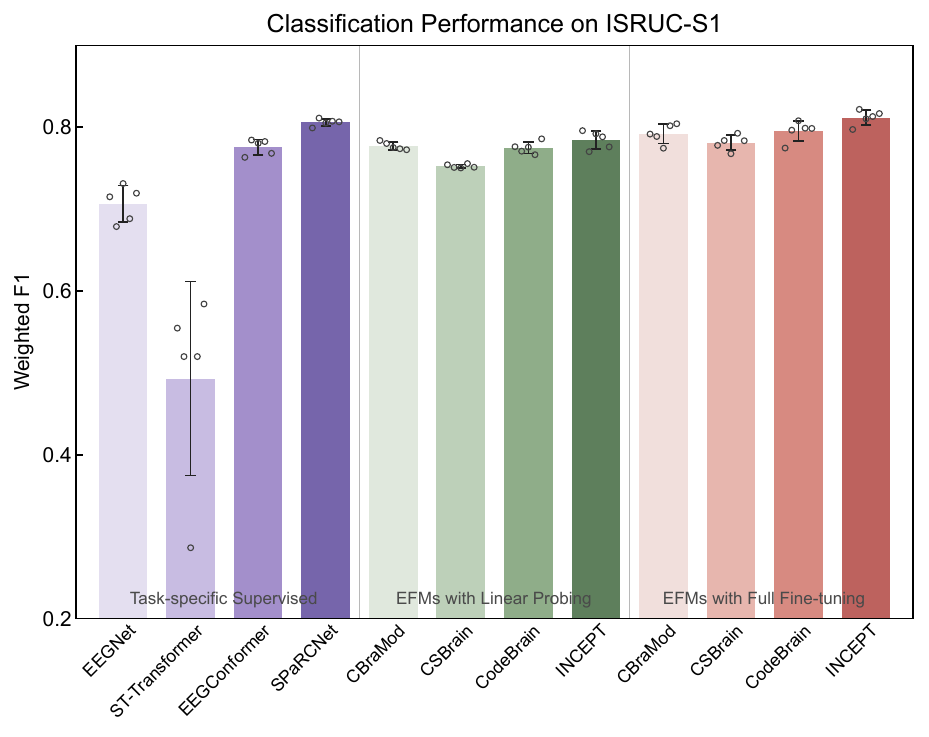} &
    \includegraphics[width=0.32\textwidth]{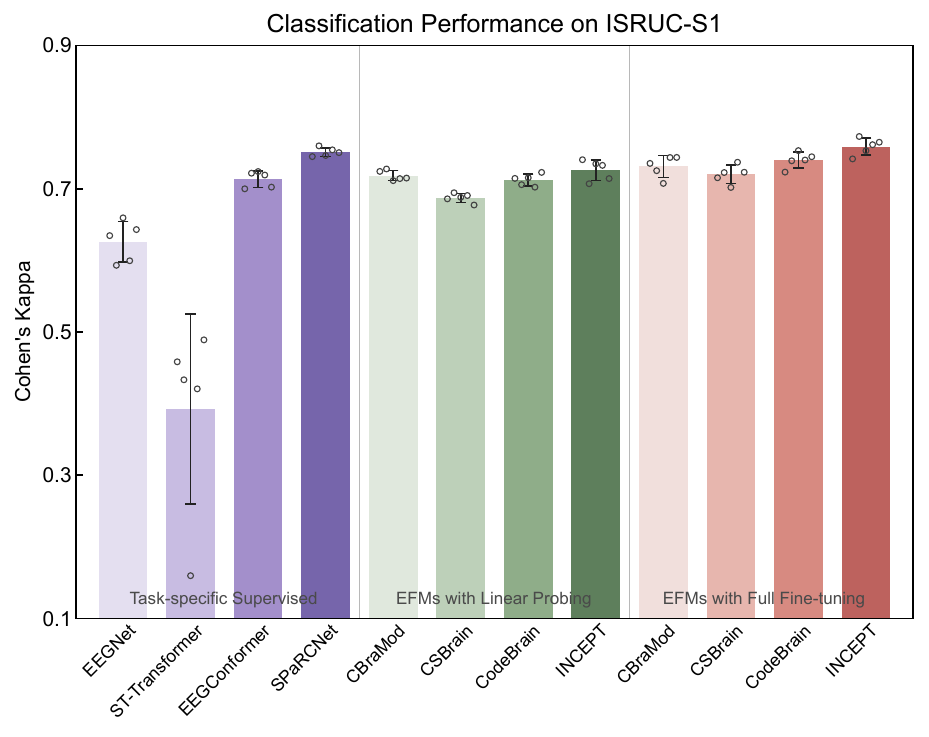}
    \end{tabular}
    \caption{
        \textbf{Brain-state decoding performance across three evaluation metrics.}
        \textbf{a}, FACED; \textbf{b}, SEED-V; \textbf{c}, PhysioNet-MI; \textbf{d}, ISRUC-S1.
        Columns show balanced accuracy, weighted F1 score and Cohen's kappa for each dataset.
    }
    \label{fig:result_bar_state}
\end{figure*}

\begin{table*}[!t]
    \centering
    \caption{\textbf{Brain-state decoding performance across montage-shifted EEG tasks.}}
    \label{table:result_state}
    \small
    \setlength{\tabcolsep}{3.5pt}
    \renewcommand{\arraystretch}{1.08}
    \begin{tabularx}{\textwidth}{l *{6}{Y}}
        \toprule
        \multirow{2}{*}{Method} & \multicolumn{3}{c}{FACED (9-class)} & \multicolumn{3}{c}{SEED-V (5-class)} \\
        \cmidrule(lr){2-4} \cmidrule(lr){5-7}
        & Bal. Acc. & Weighted F1 & Kappa & Bal. Acc. & Weighted F1 & Kappa \\
        \midrule
        \rowcolor{gray!15} \multicolumn{7}{c}{\textit{Task-specific Supervised Encoders}} \\
        EEGNet & 74.28$_{\pm 1.08\phantom{0}}$ & 73.97$_{\pm 1.09\phantom{0}}$ & 70.61$_{\pm 1.24\phantom{0}}$ & 35.80$_{\pm 0.49\phantom{0}}$ & 34.56$_{\pm 0.84\phantom{0}}$ & 18.87$_{\pm 0.68\phantom{0}}$ \\
        ST-Transformer & 73.61$_{\pm 2.64\phantom{0}}$ & 73.36$_{\pm 2.69\phantom{0}}$ & 69.98$_{\pm 3.04\phantom{0}}$ & 21.99$_{\pm 0.38\phantom{0}}$ & 19.73$_{\pm 0.83\phantom{0}}$ & \phantom{0}2.29$_{\pm 0.42\phantom{0}}$ \\
        EEGConformer & 80.68$_{\pm 1.43\phantom{0}}$ & 80.42$_{\pm 1.39\phantom{0}}$ & 77.95$_{\pm 1.62\phantom{0}}$ & 33.94$_{\pm 2.13\phantom{0}}$ & 33.28$_{\pm 2.50\phantom{0}}$ & 16.83$_{\pm 2.74\phantom{0}}$ \\
        SPaRCNet & 69.74$_{\pm 1.40\phantom{0}}$ & 69.77$_{\pm 1.35\phantom{0}}$ & 65.48$_{\pm 1.54\phantom{0}}$ & 29.42$_{\pm 0.59\phantom{0}}$ & 28.51$_{\pm 0.86\phantom{0}}$ & 11.41$_{\pm 0.83\phantom{0}}$ \\
        \midrule
        \rowcolor{gray!15} \multicolumn{7}{c}{\textit{EFMs with Linear Probing}} \\
        CBraMod & \underline{72.02}$_{\pm 0.62\phantom{0}}$ & \underline{71.46}$_{\pm 0.56\phantom{0}}$ & \underline{68.06}$_{\pm 0.69\phantom{0}}$ & 30.27$_{\pm 0.25\phantom{0}}$ & 28.15$_{\pm 1.36\phantom{0}}$ & 12.10$_{\pm 0.52\phantom{0}}$ \\
        CSBrain & 62.87$_{\pm 1.02\phantom{0}}$ & 61.56$_{\pm 1.01\phantom{0}}$ & 57.30$_{\pm 1.16\phantom{0}}$ & 27.08$_{\pm 0.24\phantom{0}}$ & 25.33$_{\pm 0.72\phantom{0}}$ & \phantom{0}8.25$_{\pm 0.24\phantom{0}}$ \\
        CodeBrain & 71.77$_{\pm 0.96\phantom{0}}$ & 71.47$_{\pm 0.98\phantom{0}}$ & 67.92$_{\pm 1.12\phantom{0}}$ & \underline{34.60}$_{\pm 0.32\phantom{0}}$ & \underline{33.95}$_{\pm 0.83\phantom{0}}$ & \underline{17.63}$_{\pm 0.47\phantom{0}}$ \\
        Ours & \textbf{76.49}$_{\pm 0.33\phantom{0}}$ & \textbf{79.60}$_{\pm 0.33\phantom{0}}$ & \textbf{73.35}$_{\pm 0.42\phantom{0}}$ & \textbf{46.91}$_{\pm 0.72\phantom{0}}$ & \textbf{45.74}$_{\pm 1.07\phantom{0}}$ & \textbf{33.00}$_{\pm 0.93\phantom{0}}$ \\
        \midrule
        \rowcolor{gray!15} \multicolumn{7}{c}{\textit{EFMs with Full Fine-tuning}} \\
        CBraMod & \underline{84.24}$_{\pm 0.20\phantom{0}}$ & \underline{83.81}$_{\pm 0.37\phantom{0}}$ & \underline{81.74}$_{\pm 0.30\phantom{0}}$ & 40.10$_{\pm 0.40\phantom{0}}$ & 38.60$_{\pm 0.28\phantom{0}}$ & 24.06$_{\pm 0.32\phantom{0}}$ \\
        CSBrain & 80.67$_{\pm 1.31\phantom{0}}$ & 80.71$_{\pm 1.04\phantom{0}}$ & 77.95$_{\pm 1.35\phantom{0}}$ & 39.78$_{\pm 0.65\phantom{0}}$ & 38.37$_{\pm 0.69\phantom{0}}$ & 23.65$_{\pm 0.80\phantom{0}}$ \\
        CodeBrain & 81.92$_{\pm 1.72\phantom{0}}$ & 82.01$_{\pm 1.73\phantom{0}}$ & 79.44$_{\pm 1.94\phantom{0}}$ & \underline{41.68}$_{\pm 0.43\phantom{0}}$ & \underline{39.88}$_{\pm 0.55\phantom{0}}$ & \underline{25.75}$_{\pm 0.51\phantom{0}}$ \\
        Ours & \textbf{84.75}$_{\pm 0.29\phantom{0}}$ & \textbf{84.48}$_{\pm 0.31\phantom{0}}$ & \textbf{82.51}$_{\pm 0.35\phantom{0}}$ & \textbf{47.06}$_{\pm 0.28\phantom{0}}$ & \textbf{45.81}$_{\pm 0.59\phantom{0}}$ & \textbf{32.90}$_{\pm 0.48\phantom{0}}$ \\
        \midrule[0.8pt]
        \multirow{2}{*}{Method} & \multicolumn{3}{c}{PhysioNet-MI (4-class)} & \multicolumn{3}{c}{ISRUC-S1 (5-class)} \\
        \cmidrule(lr){2-4} \cmidrule(lr){5-7}
        & Bal. Acc. & Weighted F1 & Kappa & Bal. Acc. & Weighted F1 & Kappa \\
        \midrule
        \rowcolor{gray!15} \multicolumn{7}{c}{\textit{Task-specific Supervised Encoders}} \\
        EEGNet & 59.79$_{\pm 1.11\phantom{0}}$ & 59.61$_{\pm 1.17\phantom{0}}$ & 46.37$_{\pm 1.48\phantom{0}}$ & 68.71$_{\pm 1.46\phantom{0}}$ & 70.66$_{\pm 1.98\phantom{0}}$ & 62.59$_{\pm 2.55\phantom{0}}$ \\
        ST-Transformer & 58.01$_{\pm 1.22\phantom{0}}$ & 57.90$_{\pm 1.48\phantom{0}}$ & 44.00$_{\pm 1.64\phantom{0}}$ & 51.12$_{\pm 7.90\phantom{0}}$ & 49.31$_{\pm 10.60}$ & 39.22$_{\pm 11.85}$ \\
        EEGConformer & 57.10$_{\pm 0.97\phantom{0}}$ & 56.69$_{\pm 1.08\phantom{0}}$ & 42.78$_{\pm 1.29\phantom{0}}$ & 75.22$_{\pm 1.47\phantom{0}}$ & 77.58$_{\pm 0.85\phantom{0}}$ & 71.33$_{\pm 1.02\phantom{0}}$ \\
        SPaRCNet & 58.68$_{\pm 1.29\phantom{0}}$ & 58.73$_{\pm 1.63\phantom{0}}$ & 44.90$_{\pm 1.72\phantom{0}}$ & 78.98$_{\pm 0.54\phantom{0}}$ & 80.59$_{\pm 0.39\phantom{0}}$ & 75.11$_{\pm 0.54\phantom{0}}$ \\
        \midrule
        \rowcolor{gray!15} \multicolumn{7}{c}{\textit{EFMs with Linear Probing}} \\
        CBraMod & \underline{53.58}$_{\pm 0.95\phantom{0}}$ & \underline{53.37}$_{\pm 1.41\phantom{0}}$ & \underline{38.09}$_{\pm 1.28\phantom{0}}$ & 75.80$_{\pm 0.35\phantom{0}}$ & \underline{77.72}$_{\pm 0.43\phantom{0}}$ & \underline{71.84}$_{\pm 0.63\phantom{0}}$ \\
        CSBrain & 51.15$_{\pm 0.61\phantom{0}}$ & 51.17$_{\pm 0.81\phantom{0}}$ & 34.87$_{\pm 0.83\phantom{0}}$ & 72.53$_{\pm 0.59\phantom{0}}$ & 75.25$_{\pm 0.21\phantom{0}}$ & 68.72$_{\pm 0.57\phantom{0}}$ \\
        CodeBrain & 50.03$_{\pm 0.59\phantom{0}}$ & 50.07$_{\pm 0.54\phantom{0}}$ & 33.39$_{\pm 0.78\phantom{0}}$ & \underline{75.82}$_{\pm 0.45\phantom{0}}$ & 77.50$_{\pm 0.65\phantom{0}}$ & 71.20$_{\pm 0.73\phantom{0}}$ \\
        Ours & \textbf{58.98}$_{\pm 0.39\phantom{0}}$ & \textbf{58.65}$_{\pm 0.39\phantom{0}}$ & \textbf{45.30}$_{\pm 0.52\phantom{0}}$ & \textbf{77.33}$_{\pm 0.63\phantom{0}}$ & \textbf{78.45}$_{\pm 0.98\phantom{0}}$ & \textbf{72.58}$_{\pm 1.29\phantom{0}}$ \\
        \midrule
        \rowcolor{gray!15} \multicolumn{7}{c}{\textit{EFMs with Full Fine-tuning}} \\
        CBraMod & \underline{62.73}$_{\pm 0.64\phantom{0}}$ & \underline{62.70}$_{\pm 0.65\phantom{0}}$ & \underline{50.29}$_{\pm 0.86\phantom{0}}$ & \underline{78.22}$_{\pm 0.77\phantom{0}}$ & 79.22$_{\pm 1.07\phantom{0}}$ & 73.10$_{\pm 1.36\phantom{0}}$ \\
        CSBrain & 61.91$_{\pm 0.48\phantom{0}}$ & 62.12$_{\pm 0.43\phantom{0}}$ & 49.21$_{\pm 0.64\phantom{0}}$ & 74.90$_{\pm 0.94\phantom{0}}$ & 78.12$_{\pm 0.82\phantom{0}}$ & 71.99$_{\pm 1.15\phantom{0}}$ \\
        CodeBrain & 61.17$_{\pm 0.85\phantom{0}}$ & 61.30$_{\pm 0.82\phantom{0}}$ & 48.22$_{\pm 1.13\phantom{0}}$ & 77.43$_{\pm 0.78\phantom{0}}$ & \underline{79.53}$_{\pm 1.11\phantom{0}}$ & \underline{74.00}$_{\pm 0.98\phantom{0}}$ \\
        Ours & \textbf{63.70}$_{\pm 0.62\phantom{0}}$ & \textbf{63.84}$_{\pm 0.62\phantom{0}}$ & \textbf{51.59}$_{\pm 0.82\phantom{0}}$ & \textbf{79.88}$_{\pm 1.04\phantom{0}}$ & \textbf{81.19}$_{\pm 0.82\phantom{0}}$ & \textbf{75.87}$_{\pm 1.07\phantom{0}}$ \\
        \bottomrule
    \end{tabularx}

    \vspace{0.35em}
    \begin{minipage}{\textwidth}
    \footnotesize
    Values are reported as mean $\pm$ s.d. across five runs with different random seeds. Results are shown for task-specific supervised encoders trained from scratch and EEG foundation models evaluated under linear probing with frozen backbones or full fine-tuning. Best and second-best results within each evaluation protocol and dataset are shown in bold and underlined, respectively.
    \end{minipage}
\end{table*}

\paragraph{(i) Affective-state decoding across low- and high-density emotion EEG}
Affective decoding is a subject-sensitive brain-state problem, and FACED \cite{chen2023large} and SEED-V \cite{liu2021comparing} expose different transfer conditions through their segment duration, channel density and label spaces. Fig.~\ref{fig:result_bar_state}a,b and Table~\ref{table:result_state} show that supervised encoders can still be strong in favourable affective settings: on FACED, EEGConformer reaches 80.68\% balanced accuracy. INCEPT is lower on frozen balanced accuracy but provides the strongest frozen weighted F1 (79.60\%) and kappa (73.35\%) among foundation models. Relative to the average of the other frozen EFMs, INCEPT improves weighted F1 by 16.8\% and reduces weighted-F1 standard deviation by 61.2\%. After full fine-tuning, it becomes the strongest model across all three FACED metrics, reaching 84.75\% balanced accuracy, 84.48\% weighted F1 and 82.51\% kappa. Thus, FACED combines a competitive supervised baseline with a clear fine-tuning gain for INCEPT.

SEED-V is more challenging because its five affective classes are decoded from shorter, high-density segments. Here the advantage of INCEPT is more direct: under linear probing, it reaches 46.91\% balanced accuracy, 45.74\% weighted F1 and 33.00\% kappa, outperforming the strongest task-specific encoder by 31.0\%, 32.4\% and 74.9\%, respectively. Full fine-tuning keeps INCEPT highest across all three metrics and improves weighted F1 by 17.6\% over the average of the other fine-tuned EFMs. The small difference between linear probing and full fine-tuning shows that most of the SEED-V gain is already present in the frozen readout.

\paragraph{(ii) Sensorimotor-state decoding from high-density motor-imagery EEG}
Motor-imagery decoding tests whether a reusable representation can support BCI-related sensorimotor states rather than only clinical or affective EEG. PhysioNet-MI uses a 64-channel montage and four imagery classes \cite{PhysioNet-eegmmidb-1.0.0,schalk2004bci2000,goldberger2000physiobank}, making it a high-density, task-specific brain-state setting. Fig.~\ref{fig:result_bar_state}c and Table~\ref{table:result_state} show that task-specific supervised encoders remain competitive, as expected for labelled BCI trials, whereas most frozen foundation models lag behind them. INCEPT is the exception: under linear probing it reaches 58.98\% balanced accuracy, 58.65\% weighted F1 and 45.30\% kappa, close to the strongest supervised results. Compared with the average of the other frozen EFMs, INCEPT improves balanced accuracy, weighted F1 and kappa by 14.4\%, 13.6\% and 27.6\%, respectively. Full fine-tuning further raises performance to 63.70\%, 63.84\% and 51.59\%, surpassing the supervised baselines across all three metrics. PhysioNet-MI therefore provides the clearest example in this group of a frozen advantage that becomes larger after full fine-tuning.

\paragraph{(iii) Sleep-state staging from sparse clinical sleep montages}
Sleep staging requires temporally organized state decoding from sparse clinical EEG. ISRUC-S1 \cite{khalighi2016isruc} uses only six EEG channels, making it a strong montage-shifted test of an encoder pre-trained mainly on 19-channel clinical EEG. Fig.~\ref{fig:result_bar_state}d and Table~\ref{table:result_state} show that SPaRCNet remains a competitive supervised baseline, reaching 80.59\% weighted F1. INCEPT is nevertheless the strongest foundation model under both protocols. Linear probing gives 77.33\% balanced accuracy, 78.45\% weighted F1 and 72.58\% kappa. Full fine-tuning increases these scores to 79.88\%, 81.19\% and 75.87\%, exceeding the supervised average by 16.4\%, 15.5\% and 20.5\%, respectively. The improvement after fine-tuning is modest but consistent for this sparse sleep montage.

\vspace{0.5em}
\noindent\textbf{\textit{Together,}} the brain-state results show that INCEPT supports affective, sensorimotor and sleep-state decoding under substantially different montage and label conditions. Linear probing is already strong for several affective and sleep readouts, whereas full fine-tuning is most useful when task labels must reshape the representation for high-density motor imagery or montage-shifted multi-class state decoding. More detailed quantitative comparisons for FACED, SEED-V, PhysioNet-MI and ISRUC-S1 are reported in Supplementary Section~\ref{supplement:detailed_downstream}.

\subsubsection{Brain-health evaluation across psychiatric, cognitive and neurological conditions}
\label{sec:results-brain-health}

This brain-health analysis covers three condition-related settings:
\begin{itemize}
\item Depression- and stress-related EEG assessment is evaluated using Mumtaz2016 and MentalArithmetic.
\item Neurodegenerative disease evaluation is measured from resting-state EEG using ADFTD.
\item Seizure-related neurological assessment is evaluated using clinical seizure-monitoring EEG data from Siena.
\end{itemize}

\begin{figure*}[!b]
    \centering
    \setlength{\tabcolsep}{2pt}
    \renewcommand{\arraystretch}{1}
    \begin{tabular}{@{}ccc@{}}
    \imgpanellabel{0.32\textwidth}{a}{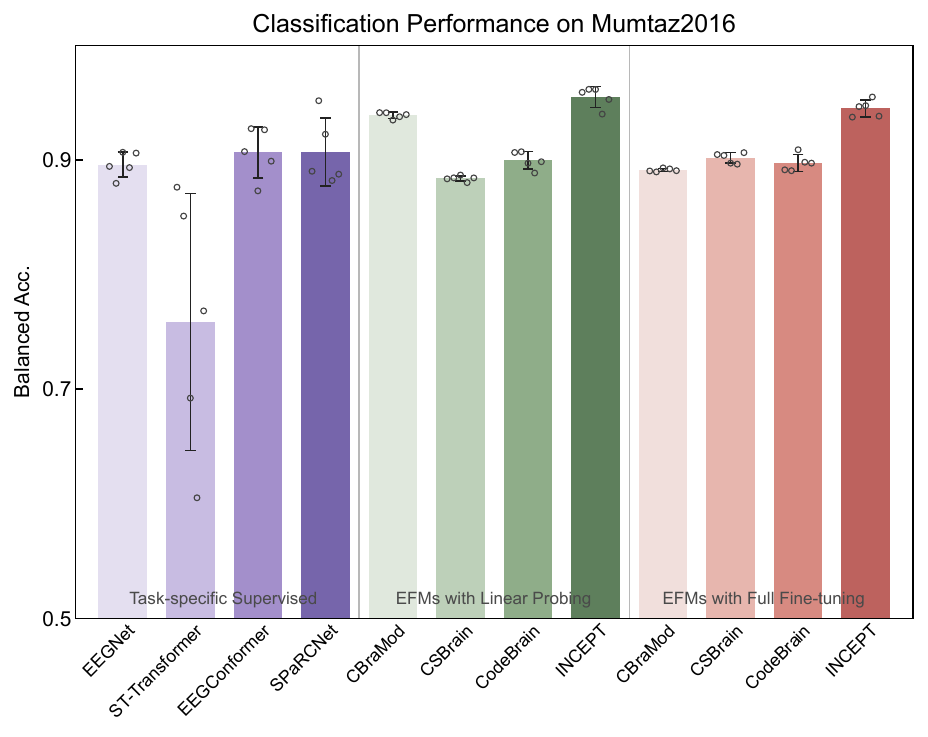} &
    \includegraphics[width=0.32\textwidth]{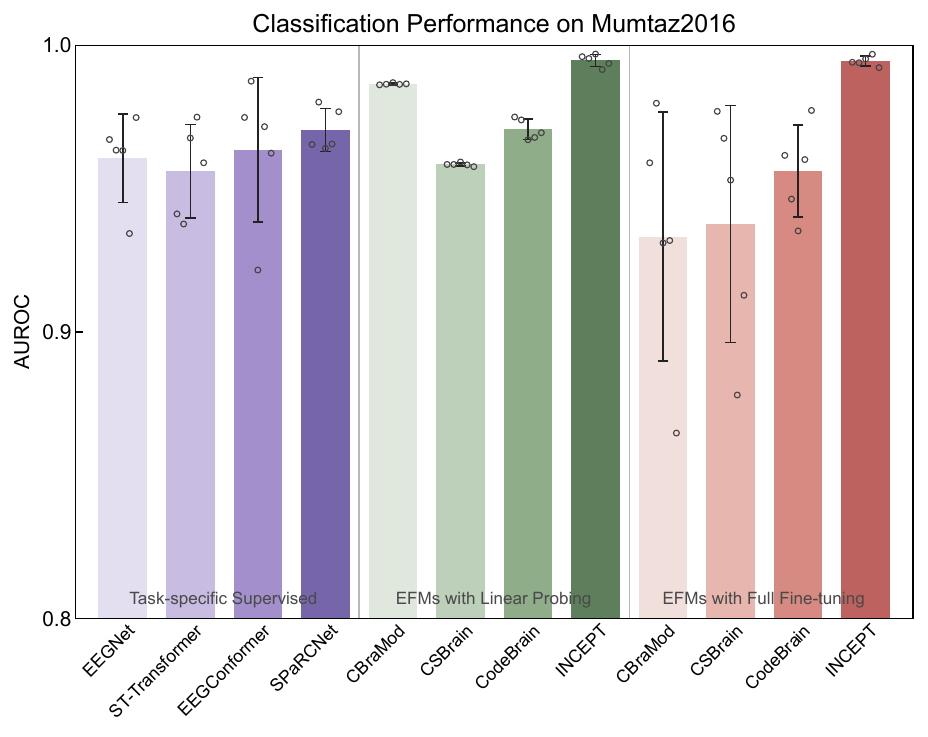} &
    \includegraphics[width=0.32\textwidth]{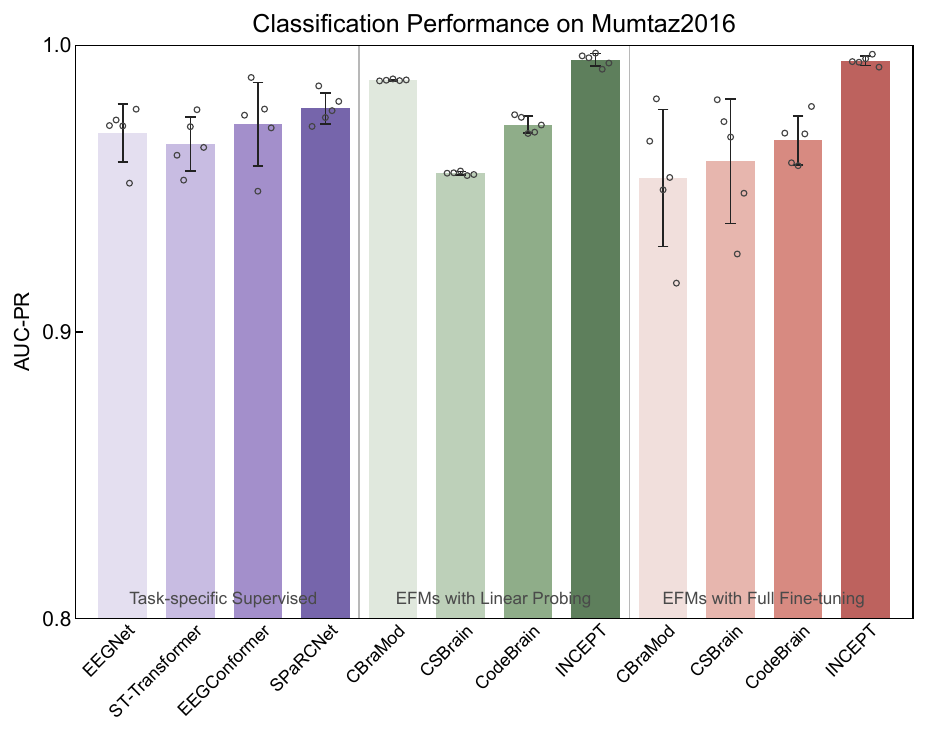} \\[-0.45em]
    \imgpanellabel{0.32\textwidth}{b}{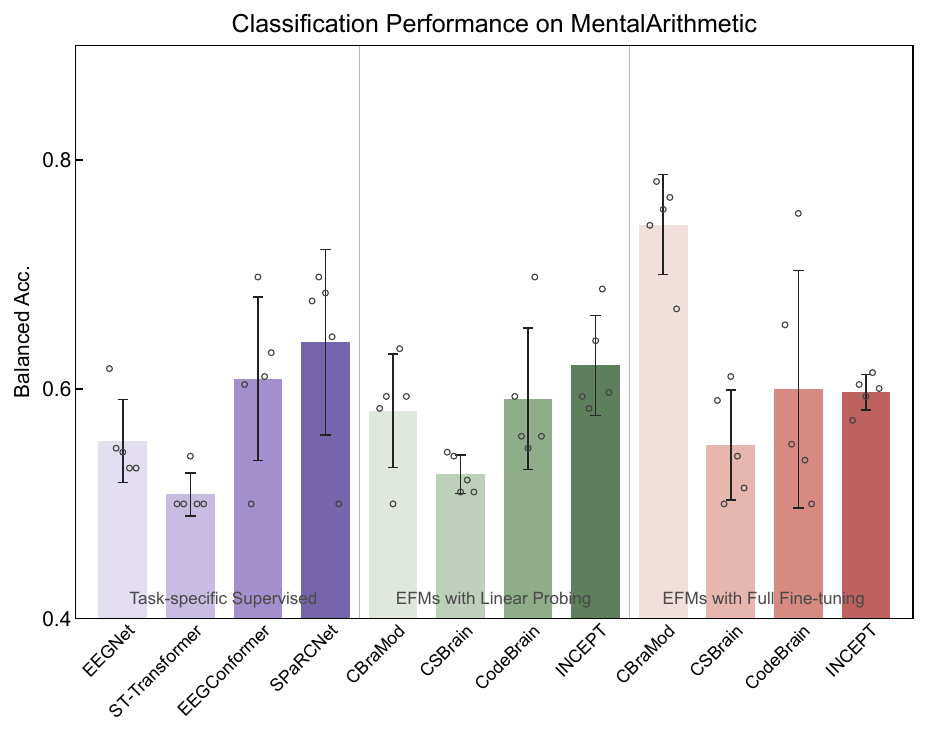} &
    \includegraphics[width=0.32\textwidth]{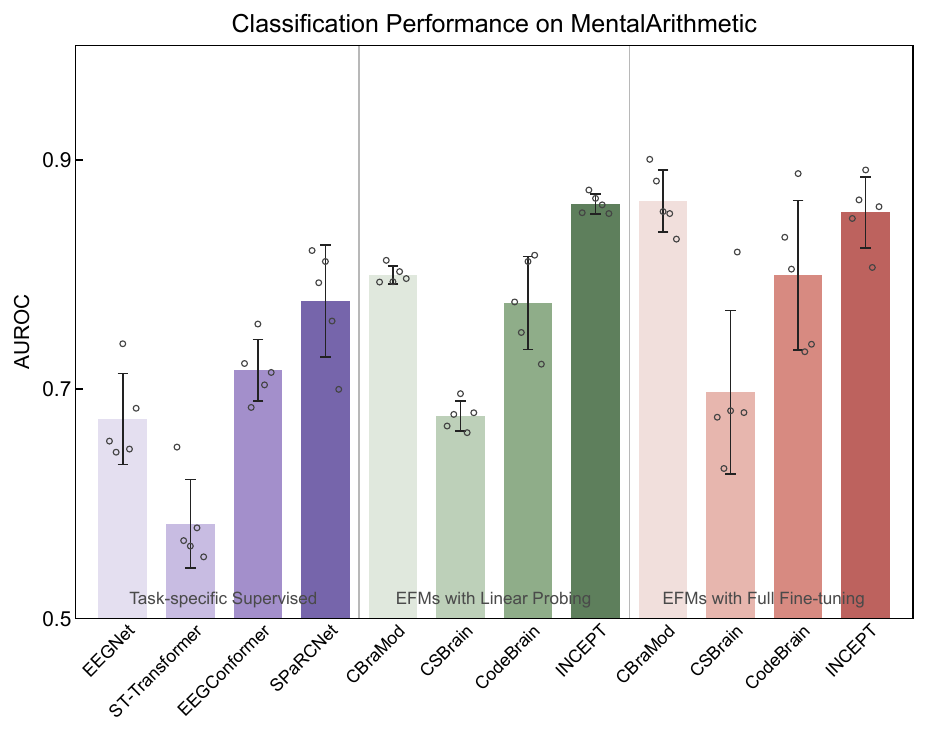} &
    \includegraphics[width=0.32\textwidth]{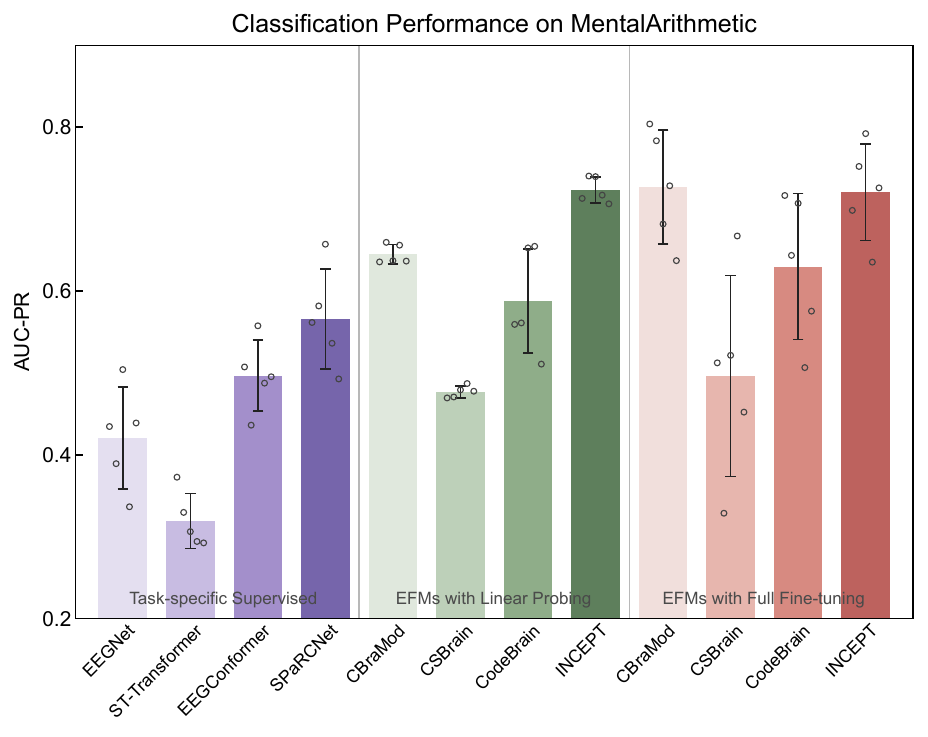} \\[-0.45em]
    \imgpanellabel{0.32\textwidth}{c}{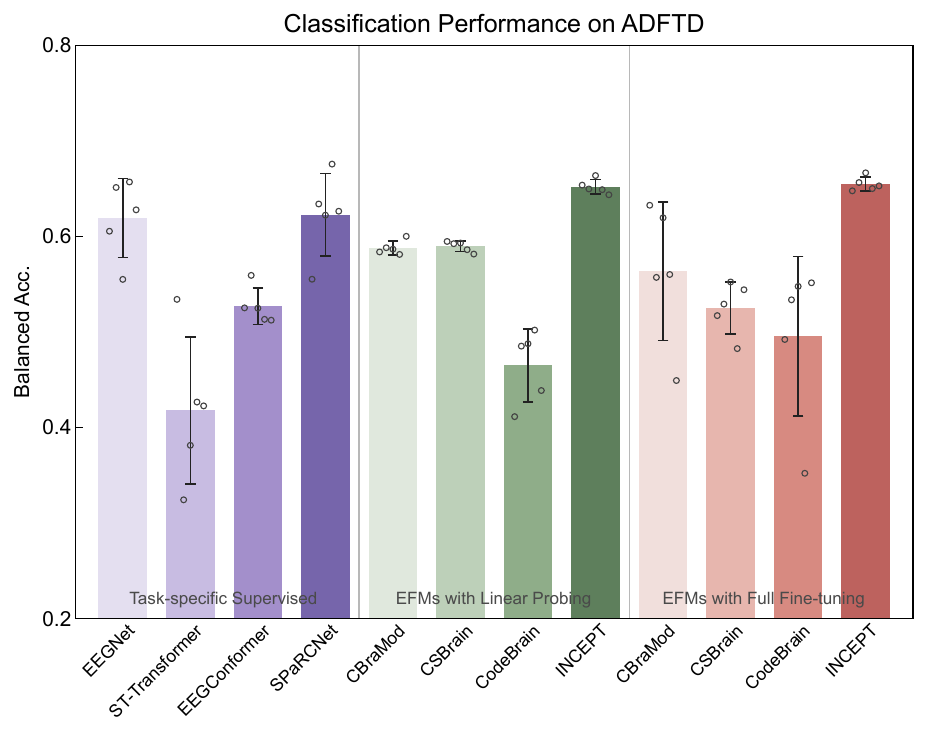} &
    \includegraphics[width=0.32\textwidth]{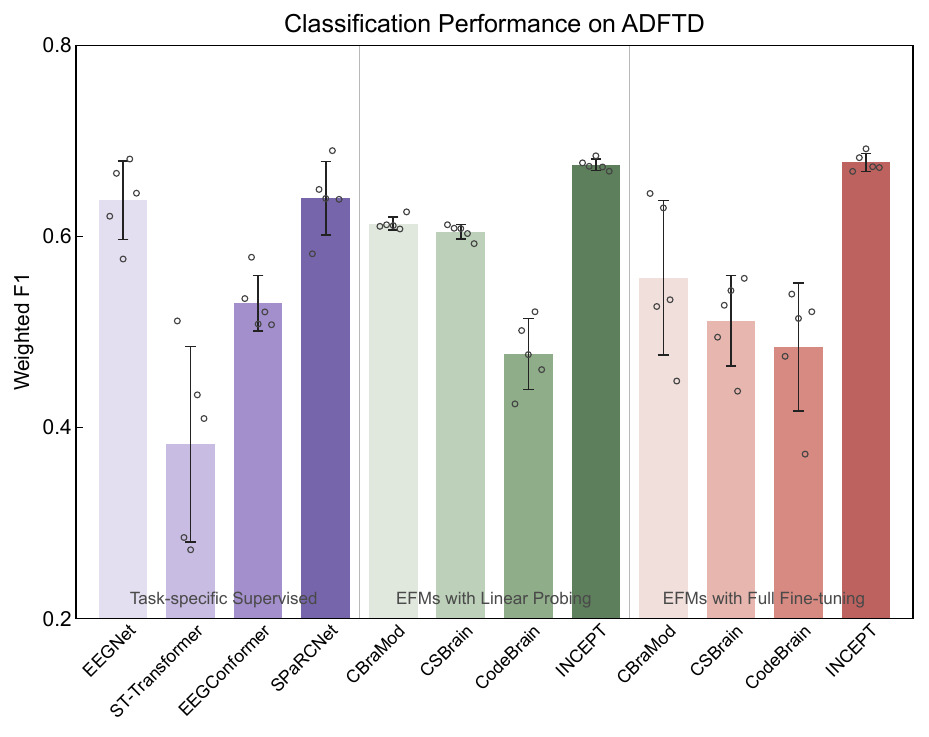} &
    \includegraphics[width=0.32\textwidth]{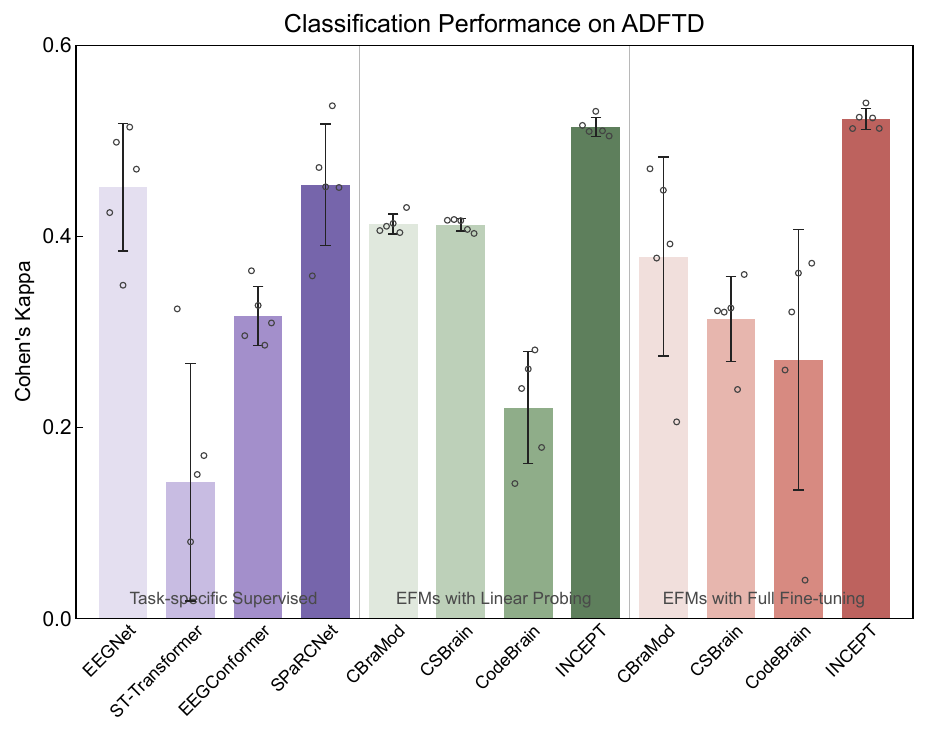} \\[-0.45em]
    \imgpanellabel{0.32\textwidth}{d}{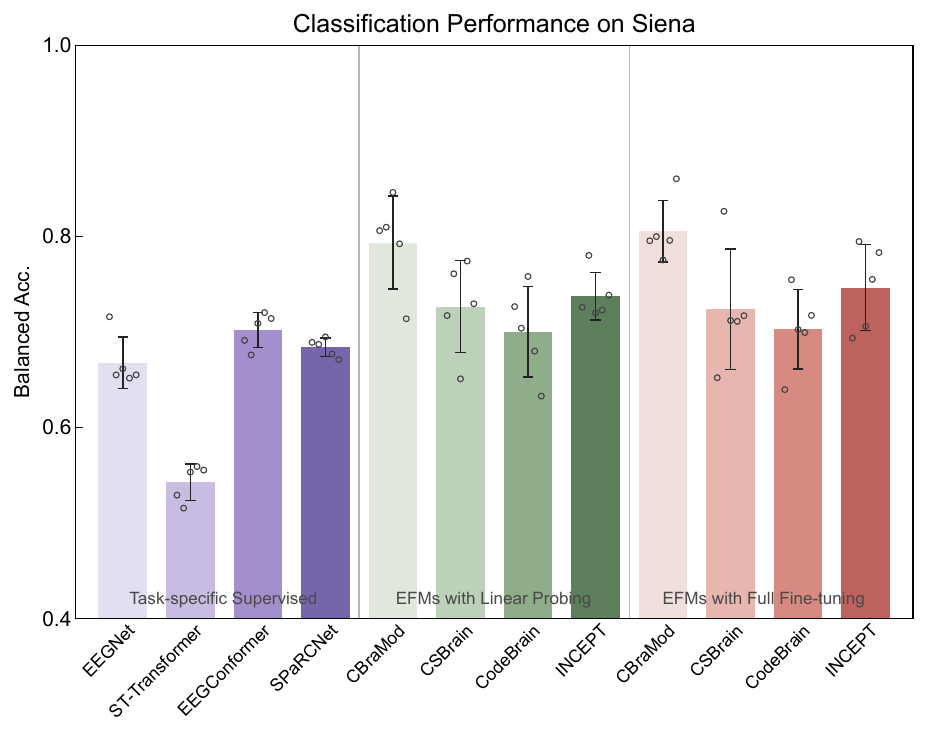} &
    \includegraphics[width=0.32\textwidth]{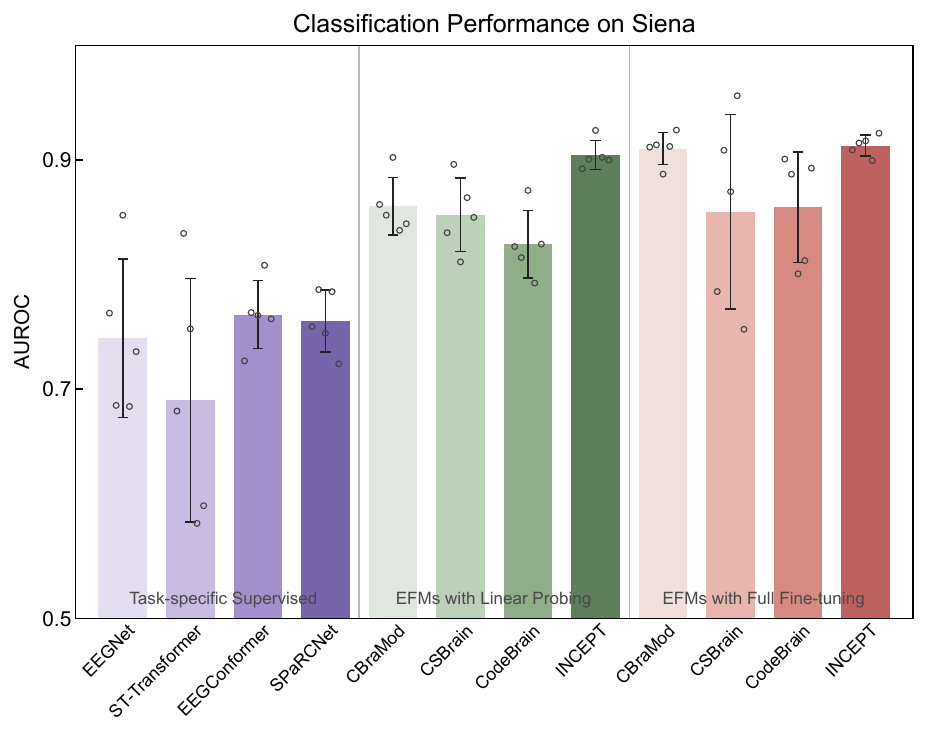} &
    \includegraphics[width=0.32\textwidth]{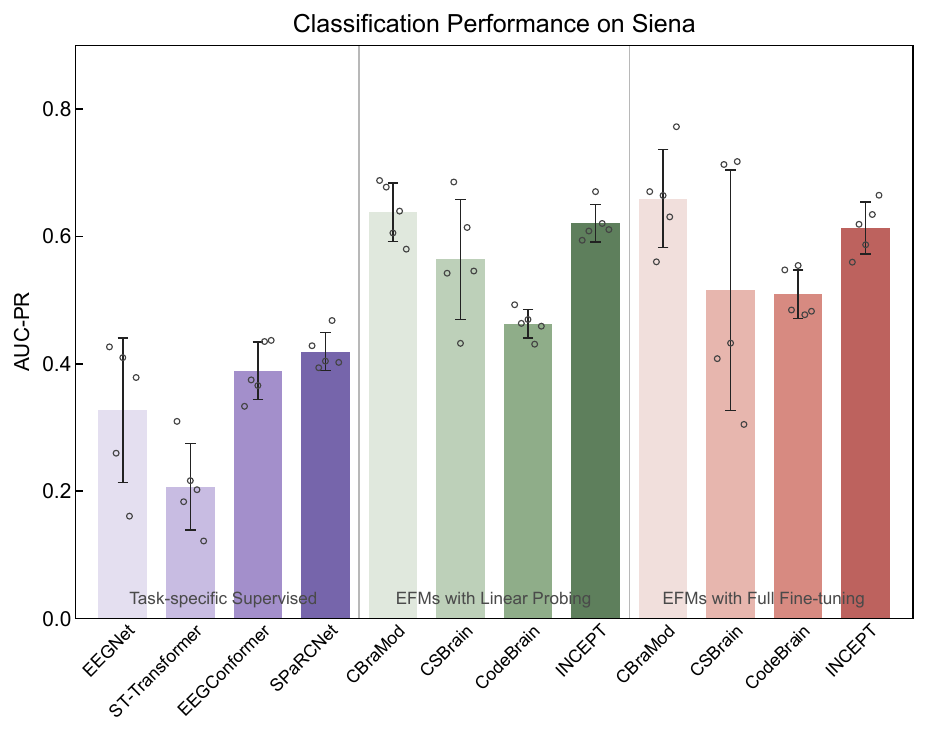}
    \end{tabular}
    \caption{
        \textbf{Brain-health evaluation performance across three evaluation metrics.}
        \textbf{a}, Mumtaz2016; \textbf{b}, MentalArithmetic; \textbf{c}, ADFTD; \textbf{d}, Siena.
        Binary datasets use balanced accuracy, AUROC and AUC-PR, whereas ADFTD uses balanced accuracy, weighted F1 score and Cohen's kappa.
    }
    \label{fig:result_bar_health}
\end{figure*}

\begin{table*}[!t]
    \centering
    \caption{\textbf{Brain-health evaluation performance across psychiatric, cognitive and neurological EEG tasks.}}
    \label{table:result_health}
    \small
    \setlength{\tabcolsep}{3.5pt}
    \renewcommand{\arraystretch}{1.08}
    \begin{tabularx}{\textwidth}{l *{6}{Y}}
        \toprule
        \multirow{2}{*}{Method} & \multicolumn{3}{c}{Mumtaz2016 (2-class)} & \multicolumn{3}{c}{MentalArithmetic (2-class)} \\
        \cmidrule(lr){2-4} \cmidrule(lr){5-7}
        & Bal. Acc. & AUROC & AUC-PR & Bal. Acc. & AUROC & AUC-PR \\
        \midrule
        \rowcolor{gray!15} \multicolumn{7}{c}{\textit{Task-specific Supervised Encoders}} \\
        EEGNet & 89.61$_{\pm 0.99\phantom{0}}$ & 96.06$_{\pm 1.38\phantom{0}}$ & 96.95$_{\pm 0.91\phantom{0}}$ & 55.49$_{\pm 3.24\phantom{0}}$ & 67.42$_{\pm 3.55\phantom{0}}$ & 42.06$_{\pm 5.58\phantom{0}}$ \\
        ST-Transformer & 75.87$_{\pm 10.04}$ & 95.61$_{\pm 1.46\phantom{0}}$ & 96.57$_{\pm 0.84\phantom{0}}$ & 50.83$_{\pm 1.67\phantom{0}}$ & 58.28$_{\pm 3.44\phantom{0}}$ & 31.90$_{\pm 2.99\phantom{0}}$ \\
        EEGConformer & 90.66$_{\pm 2.00\phantom{0}}$ & 96.36$_{\pm 2.25\phantom{0}}$ & 97.25$_{\pm 1.31\phantom{0}}$ & 60.90$_{\pm 6.38\phantom{0}}$ & 71.64$_{\pm 2.40\phantom{0}}$ & 49.68$_{\pm 3.89\phantom{0}}$ \\
        SPaRCNet & 90.69$_{\pm 2.65\phantom{0}}$ & 97.04$_{\pm 0.67\phantom{0}}$ & 97.80$_{\pm 0.49\phantom{0}}$ & 64.10$_{\pm 7.25\phantom{0}}$ & 77.70$_{\pm 4.39\phantom{0}}$ & 56.59$_{\pm 5.45\phantom{0}}$ \\
        \midrule
        \rowcolor{gray!15} \multicolumn{7}{c}{\textit{EFMs with Linear Probing}} \\
        CBraMod & \underline{93.90}$_{\pm 0.24\phantom{0}}$ & \underline{98.65}$_{\pm 0.03\phantom{0}}$ & \underline{98.79}$_{\pm 0.02\phantom{0}}$ & 58.13$_{\pm 4.44\phantom{0}}$ & \underline{79.98}$_{\pm 0.71\phantom{0}}$ & \underline{64.49}$_{\pm 1.05\phantom{0}}$ \\
        CSBrain & 88.39$_{\pm 0.21\phantom{0}}$ & 95.85$_{\pm 0.05\phantom{0}}$ & 95.53$_{\pm 0.05\phantom{0}}$ & 52.57$_{\pm 1.50\phantom{0}}$ & 67.68$_{\pm 1.16\phantom{0}}$ & 47.68$_{\pm 0.64\phantom{0}}$ \\
        CodeBrain & 89.97$_{\pm 0.68\phantom{0}}$ & 97.07$_{\pm 0.32\phantom{0}}$ & 97.24$_{\pm 0.26\phantom{0}}$ & \underline{59.17}$_{\pm 5.53\phantom{0}}$ & 77.52$_{\pm 3.62\phantom{0}}$ & 58.77$_{\pm 5.68\phantom{0}}$ \\
        Ours & \textbf{95.51}$_{\pm 0.81\phantom{0}}$ & \textbf{99.47}$_{\pm 0.19\phantom{0}}$ & \textbf{99.50}$_{\pm 0.20\phantom{0}}$ & \textbf{62.08}$_{\pm 3.90\phantom{0}}$ & \textbf{86.17}$_{\pm 0.78\phantom{0}}$ & \textbf{72.34}$_{\pm 1.41\phantom{0}}$ \\
        \midrule
        \rowcolor{gray!15} \multicolumn{7}{c}{\textit{EFMs with Full Fine-tuning}} \\
        CBraMod & 89.13$_{\pm 0.13\phantom{0}}$ & 93.33$_{\pm 3.88\phantom{0}}$ & 95.37$_{\pm 2.14\phantom{0}}$ & \textbf{74.38}$_{\pm 3.89\phantom{0}}$ & \textbf{86.43}$_{\pm 2.42\phantom{0}}$ & \textbf{72.70}$_{\pm 6.20\phantom{0}}$ \\
        CSBrain & \underline{90.18}$_{\pm 0.41\phantom{0}}$ & 93.77$_{\pm 3.70\phantom{0}}$ & 95.96$_{\pm 1.95\phantom{0}}$ & 55.14$_{\pm 4.29\phantom{0}}$ & 69.75$_{\pm 6.39\phantom{0}}$ & 49.64$_{\pm 10.97}$ \\
        CodeBrain & 89.73$_{\pm 0.66\phantom{0}}$ & \underline{95.61}$_{\pm 1.43\phantom{0}}$ & \underline{96.69}$_{\pm 0.76\phantom{0}}$ & \underline{60.00}$_{\pm 9.26\phantom{0}}$ & 79.95$_{\pm 5.84\phantom{0}}$ & 62.99$_{\pm 7.99\phantom{0}}$ \\
        Ours & \textbf{94.50}$_{\pm 0.65\phantom{0}}$ & \textbf{99.45}$_{\pm 0.16\phantom{0}}$ & \textbf{99.46}$_{\pm 0.15\phantom{0}}$ & 59.72$_{\pm 1.39\phantom{0}}$ & \underline{85.42}$_{\pm 2.78\phantom{0}}$ & \underline{72.09}$_{\pm 5.28\phantom{0}}$ \\
        \midrule[0.8pt]
        \multirow{2}{*}{Method} & \multicolumn{3}{c}{ADFTD (3-class)} & \multicolumn{3}{c}{Siena (2-class)} \\
        \cmidrule(lr){2-4} \cmidrule(lr){5-7}
        & Bal. Acc. & Weighted F1 & Kappa & Bal. Acc. & AUROC & AUC-PR \\
        \midrule
        \rowcolor{gray!15} \multicolumn{7}{c}{\textit{Task-specific Supervised Encoders}} \\
        EEGNet & 61.93$_{\pm 3.70\phantom{0}}$ & 63.80$_{\pm 3.68\phantom{0}}$ & 45.15$_{\pm 5.96\phantom{0}}$ & 66.78$_{\pm 2.42\phantom{0}}$ & 74.44$_{\pm 6.18\phantom{0}}$ & 32.71$_{\pm 10.16}$ \\
        ST-Transformer & 41.78$_{\pm 6.89\phantom{0}}$ & 38.24$_{\pm 9.14\phantom{0}}$ & 14.26$_{\pm 11.12}$ & 54.25$_{\pm 1.71\phantom{0}}$ & 69.03$_{\pm 9.50\phantom{0}}$ & 20.68$_{\pm 6.08\phantom{0}}$ \\
        EEGConformer & 52.70$_{\pm 1.70\phantom{0}}$ & 53.00$_{\pm 2.61\phantom{0}}$ & 31.67$_{\pm 2.74\phantom{0}}$ & 70.21$_{\pm 1.63\phantom{0}}$ & 76.52$_{\pm 2.65\phantom{0}}$ & 38.92$_{\pm 4.06\phantom{0}}$ \\
        SPaRCNet & 62.27$_{\pm 3.88\phantom{0}}$ & 63.99$_{\pm 3.45\phantom{0}}$ & 45.42$_{\pm 5.71\phantom{0}}$ & 68.38$_{\pm 0.86\phantom{0}}$ & 75.96$_{\pm 2.43\phantom{0}}$ & 41.93$_{\pm 2.69\phantom{0}}$ \\
        \midrule
        \rowcolor{gray!15} \multicolumn{7}{c}{\textit{EFMs with Linear Probing}} \\
        CBraMod & 58.80$_{\pm 0.65\phantom{0}}$ & \underline{61.36}$_{\pm 0.63\phantom{0}}$ & \underline{41.30}$_{\pm 0.93\phantom{0}}$ & \textbf{79.36}$_{\pm 4.37\phantom{0}}$ & \underline{85.97}$_{\pm 2.26\phantom{0}}$ & \textbf{63.81}$_{\pm 4.12\phantom{0}}$ \\
        CSBrain & \underline{58.96}$_{\pm 0.50\phantom{0}}$ & 60.49$_{\pm 0.69\phantom{0}}$ & 41.23$_{\pm 0.59\phantom{0}}$ & 72.66$_{\pm 4.31\phantom{0}}$ & 85.22$_{\pm 2.86\phantom{0}}$ & 56.40$_{\pm 8.42\phantom{0}}$ \\
        CodeBrain & 46.50$_{\pm 3.42\phantom{0}}$ & 47.69$_{\pm 3.33\phantom{0}}$ & 22.07$_{\pm 5.24\phantom{0}}$ & 70.03$_{\pm 4.24\phantom{0}}$ & 82.64$_{\pm 2.64\phantom{0}}$ & 46.32$_{\pm 1.99\phantom{0}}$ \\
        Ours & \textbf{65.20}$_{\pm 0.67\phantom{0}}$ & \textbf{67.52}$_{\pm 0.54\phantom{0}}$ & \textbf{51.46}$_{\pm 0.89\phantom{0}}$ & \underline{73.75}$_{\pm 2.23\phantom{0}}$ & \textbf{90.42}$_{\pm 1.13\phantom{0}}$ & \underline{62.08}$_{\pm 2.62\phantom{0}}$ \\
        \midrule
        \rowcolor{gray!15} \multicolumn{7}{c}{\textit{EFMs with Full Fine-tuning}} \\
        CBraMod & \underline{56.37}$_{\pm 6.49\phantom{0}}$ & \underline{55.68}$_{\pm 7.25\phantom{0}}$ & \underline{37.89}$_{\pm 9.32\phantom{0}}$ & \textbf{80.54}$_{\pm 2.88\phantom{0}}$ & \underline{91.00}$_{\pm 6.87\phantom{0}}$ & \textbf{65.96}$_{\pm 6.87\phantom{0}}$ \\
        CSBrain & 52.51$_{\pm 2.45\phantom{0}}$ & 51.20$_{\pm 4.23\phantom{0}}$ & 31.36$_{\pm 3.97\phantom{0}}$ & 72.37$_{\pm 5.66\phantom{0}}$ & 85.49$_{\pm 7.59\phantom{0}}$ & 51.52$_{\pm 1.69\phantom{0}}$ \\
        CodeBrain & 49.54$_{\pm 7.47\phantom{0}}$ & 48.43$_{\pm 6.00\phantom{0}}$ & 27.10$_{\pm 12.19}$ & 70.27$_{\pm 3.72\phantom{0}}$ & 85.89$_{\pm 4.31\phantom{0}}$ & 50.92$_{\pm 3.42\phantom{0}}$ \\
        Ours & \textbf{65.48}$_{\pm 0.66\phantom{0}}$ & \textbf{67.75}$_{\pm 0.86\phantom{0}}$ & \textbf{52.29}$_{\pm 0.98\phantom{0}}$ & \underline{74.65}$_{\pm 4.04\phantom{0}}$ & \textbf{91.26}$_{\pm 0.80\phantom{0}}$ & \underline{61.30}$_{\pm 3.66\phantom{0}}$ \\
        \bottomrule
    \end{tabularx}

    \vspace{0.35em}
    \begin{minipage}{\textwidth}
    \footnotesize
    Values are reported as mean $\pm$ s.d. across five runs with different random seeds. Results are shown for task-specific supervised encoders trained from scratch and EEG foundation models evaluated under linear probing with frozen backbones or full fine-tuning. Best and second-best results within each evaluation protocol and dataset are shown in bold and underlined, respectively.
    \end{minipage}
\end{table*}

\paragraph{(i) Depression- and stress-related EEG assessment}
Mumtaz2016 \cite{mumtaz2016mdd,mumtaz2018machine} and MentalArithmetic \cite{zyma2019electroencephalograms} test montage-matched brain-health assessment, where depression- or stress-related information is distributed across 19-channel EEG rather than tied to a single event. These two datasets therefore test whether a 19-channel clinical representation can support both psychiatric trait-related classification and cognitive-demand/stress assessment. On Mumtaz2016, the frozen-representation pattern is clear: INCEPT reaches 95.51\% balanced accuracy, 99.47\% AUROC and 99.50\% AUC-PR under linear probing, exceeding the strongest task-specific encoder across all three metrics. Because performance is near ceiling, the important observation is not only the absolute score but also the stability of the readout: the average linear-probing EFM reduces seed-to-seed standard deviation by 87.6\% for balanced accuracy relative to supervised encoders, and INCEPT remains the strongest frozen model. Full fine-tuning remains high but does not improve over the frozen readout for INCEPT, reinforcing the interpretation that montage-matched psychiatric EEG can already be well organized by large-scale self-supervised pre-training.

On MentalArithmetic, INCEPT gives the best frozen AUROC and AUC-PR (86.17\% and 72.34\%) and the highest frozen balanced accuracy among foundation models (62.08\%). Compared with the average of the other frozen EFMs, it improves AUROC by 14.8\% and AUC-PR by 27.0\%, with a smaller advantage in balanced accuracy. Full fine-tuning does not improve INCEPT on this task, with balanced accuracy decreasing to 59.72\%, but its AUROC and AUC-PR remain second-highest among fine-tuned models. The main MentalArithmetic result is therefore score-level separability under linear probing, rather than a consistent fine-tuning gain.

\paragraph{(ii) Neurodegenerative disease evaluation from resting-state EEG}
ADFTD evaluates three-class neurodegenerative disease classification from resting-state EEG \cite{ds004504:1.0.8}. The montage is close to pre-training, but the disease-related signal is distributed across subjects and segments, making it a stringent test of frozen readability. Fig.~\ref{fig:result_bar_health}c and Table~\ref{table:result_health} show that supervised encoders such as SPaRCNet and EEGNet remain useful, with SPaRCNet reaching 62.27\% balanced accuracy and 63.99\% weighted F1. INCEPT improves all three linear-probing metrics relative to the strongest supervised encoder, reaching 65.20\% balanced accuracy, 67.52\% weighted F1 and 51.46\% kappa. Standard deviations are reduced by more than 80\% relative to the average supervised encoder across all three metrics. Full fine-tuning preserves the lead, with INCEPT remaining best at 65.48\%, 67.75\% and 52.29\%. Thus, ADFTD shows both a frozen-readout advantage and a small additional fine-tuning gain.

\paragraph{(iii) Seizure-related neurological assessment under extended clinical montages}
Siena \cite{PhysioNet-siena-scalp-eeg-1.0.0,detti2020eeg,goldberger2000physiobank} provides a montage-shifted brain-health task with an extended 29-channel clinical montage and sparse seizure events. Fig.~\ref{fig:result_bar_health}d and Table~\ref{table:result_health} show that task-specific supervised encoders are weaker here than on montage-matched brain-health datasets, whereas foundation models provide stronger transfer, especially for ranking-based metrics. Under linear probing, INCEPT reaches the highest AUROC (90.42\%) and the second-highest AUC-PR (62.08\%), but its balanced accuracy is lower than the best thresholded result. Compared with the strongest task-specific encoder, INCEPT improves AUROC by 18.2\% and AUC-PR by 48.1\%. After full fine-tuning, INCEPT again has the strongest AUROC (91.26\%) and the second-highest AUC-PR (61.30\%). These results emphasize the positive role of INCEPT as a score-level seizure-ranking model, while avoiding the overclaim that it uniformly dominates every seizure metric.

Siena is metric-specific: AUROC and AUC-PR measure score-level separability under class imbalance, whereas balanced accuracy depends on the thresholded class decision. INCEPT therefore does not uniformly dominate Siena, but it provides the strongest AUROC under an extended clinical montage and remains highly competitive for AUC-PR. After full fine-tuning, it also reduces AUROC and AUC-PR standard deviation relative to CBraMod by 88.4\% and 46.7\%, respectively.

\vspace{0.5em}
\noindent\textbf{\textit{Together,}} the brain-health results show that montage-matched psychiatric, cognitive and neurodegenerative tasks often benefit strongly from frozen INCEPT representations, whereas Siena depends more strongly on the chosen metric. Linear probing is particularly effective for Mumtaz2016 and ADFTD, full fine-tuning preserves INCEPT's disease-related advantage on ADFTD, and Siena highlights the distinction between score-level ranking and thresholded class decisions. More detailed quantitative comparisons for all brain-health datasets are reported in Supplementary Section~\ref{supplement:detailed_downstream}.

\subsection{Neural representation analysis across subject- and state-level EEG organization}
\label{sec:results-representation}

This section analyzes neural representation structure to interpret how INCEPT organizes EEG beyond downstream scores:
\begin{itemize}
\item Section~\ref{sec:results-subject-representation} examines whether one-label-per-subject representations preserve subject-level stability while separating condition-related structure.
\item Section~\ref{sec:results-state-representation} examines whether multi-label-per-subject representations preserve subject-specific structure while encoding task-related state differences.
\end{itemize}

\begin{figure*}[!htbp]
    \centering
    \setlength{\tabcolsep}{0pt}
    \renewcommand{\arraystretch}{1}

    \begin{tabular}{@{}p{0.39\textwidth}@{\hspace{0.02\textwidth}}p{0.568\textwidth}@{}}

    \vspace{0pt}
    \analysispanellabel{width=\linewidth}{a}
        {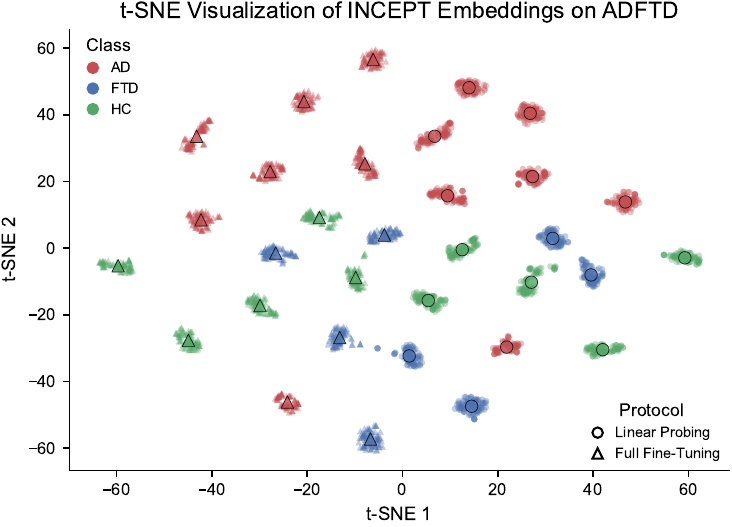}
    &
    \vspace{0pt}
    \analysispanellabel{width=\linewidth}{b}
        {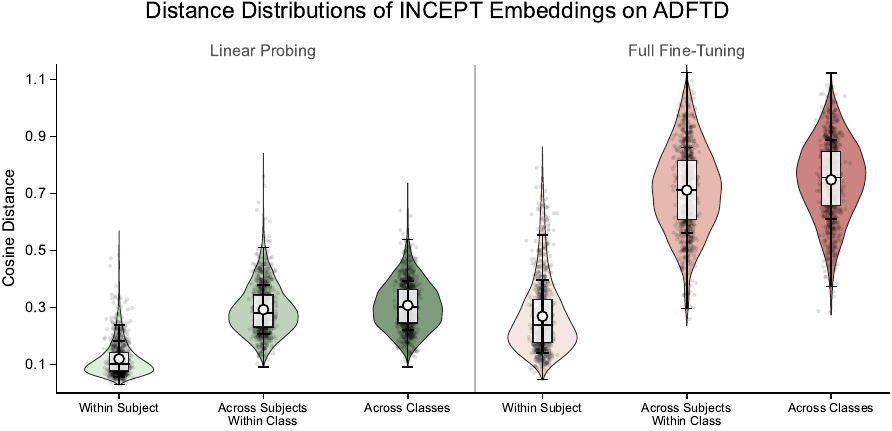}
    \\[-0.35em]

    \vspace{0pt}
    \analysispanellabel{width=\linewidth}{c}
        {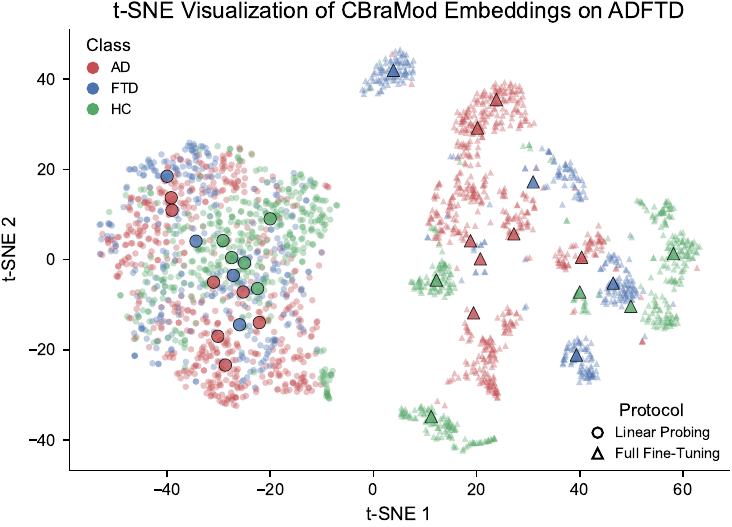}
    &
    \vspace{0pt}
    \analysispanellabel{width=\linewidth}{d}
        {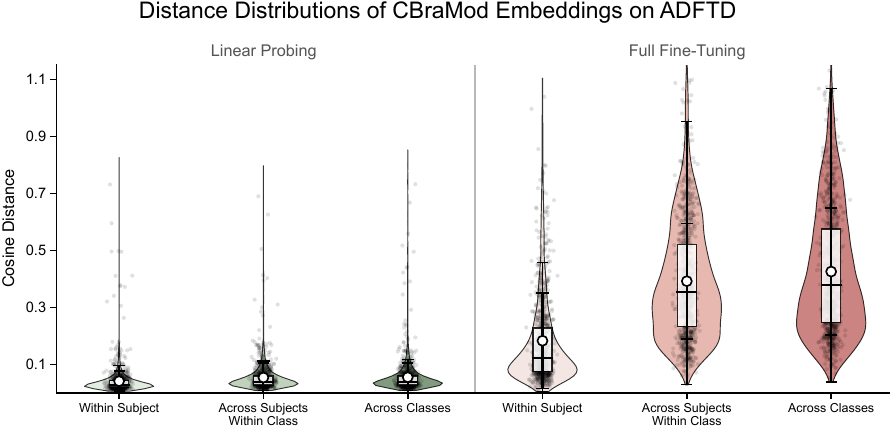}
    \\[-0.35em]

    \vspace{0pt}
    \analysispanellabel{width=\linewidth}{e}
        {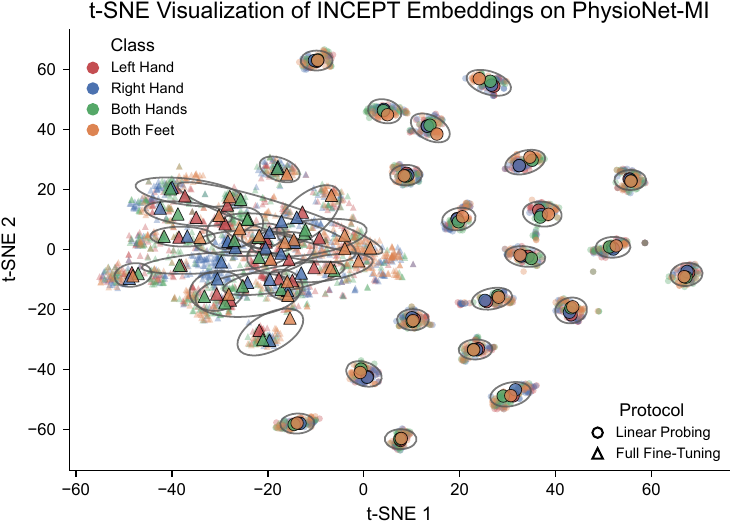}
    &
    \vspace{0pt}
    \analysispanellabel{width=\linewidth}{f}
        {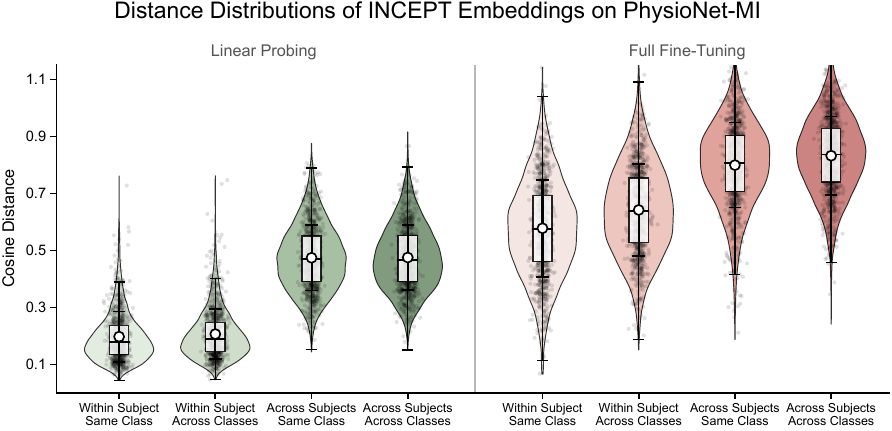}
    \\[-0.35em]

    \vspace{0pt}
    \analysispanellabel{width=\linewidth}{g}
        {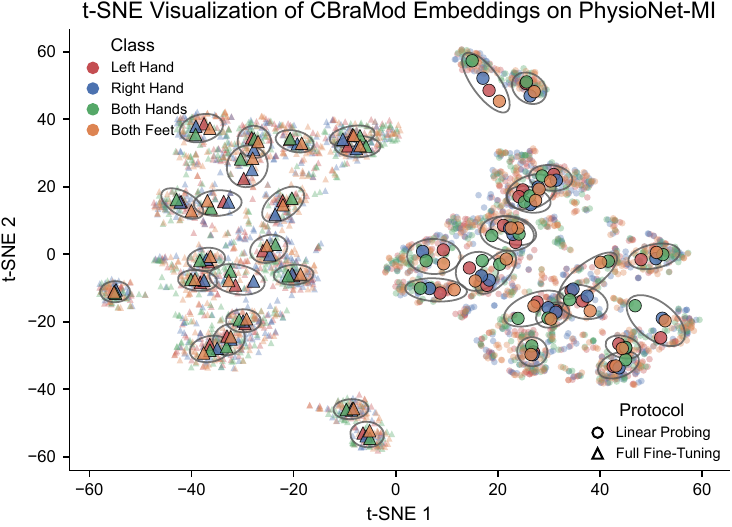}
    &
    \vspace{0pt}
    \analysispanellabel{width=\linewidth}{h}
        {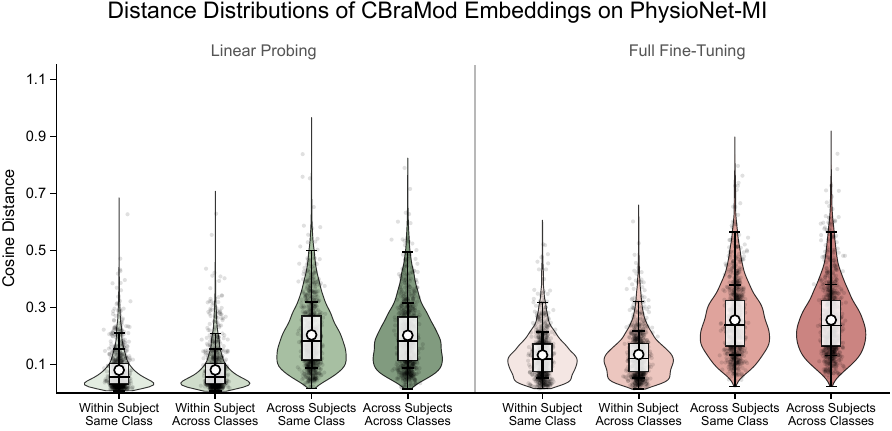}
    \\

    \end{tabular}

    \caption{
        \textbf{Representation analysis of learned EEG features across neurodegenerative disease and motor imagery tasks.}
        \textbf{a}-\textbf{d}, Representation analysis on the ADFTD neurodegenerative disease classification dataset comparing INCEPT (\textbf{a}, \textbf{b}) and CBraMod (\textbf{c}, \textbf{d}). Left panels (\textbf{a}, \textbf{c}) show joint t-SNE visualizations of EEG segments, where colors denote diagnostic classes and marker shapes denote adaptation protocols (circles for linear probing and triangles for full fine-tuning; large markers indicate subject-level centroids). Right panels (\textbf{b}, \textbf{d}) show cosine distance distributions for segment pairs from the same subject, from different subjects within the same class and from different classes.
        \textbf{e}-\textbf{h}, Corresponding analysis on the PhysioNet-MI motor imagery dataset for INCEPT (\textbf{e}, \textbf{f}) and CBraMod (\textbf{g}, \textbf{h}). In \textbf{e} and \textbf{g}, ellipses enclose the four class centroids from the same subject under each adaptation protocol, illustrating subject-level organization across motor imagery conditions. Distance distributions in \textbf{f} and \textbf{h} are computed for four types of trial pairs: same subject and same class, same subject and different classes, different subjects and same class, and different subjects and different classes.
        For all violin plots, box plots summarize the median and interquartile range, and white circles with error bars indicate mean $\pm$ s.d.
    }
    \label{fig:representation_analysis}
\end{figure*}

The preceding quantitative analyses show how well INCEPT transfers across heterogeneous downstream EEG applications, but they do not fully explain how the learned representation space is organized. A model can perform well in downstream decoding without necessarily organizing EEG signals according to stable neural structure. We therefore examine the distribution of learned representations in two complementary settings with different subject-label relationships. This distinction also reflects, in a simplified form, the organization of the preceding application-level results. Brain-health evaluation tasks often emphasize relatively stable subject-level conditions, and therefore frequently resemble one-label-per-subject settings. Brain-state decoding tasks, by contrast, often involve transient cognitive, affective, sleep or sensorimotor states, and therefore more commonly contain multiple labels or states within the same subject. ADFTD \cite{ds004504:1.0.8} serves as a representative one-label-per-subject brain-health setting, where each subject belongs to a single disease-related category. PhysioNet-MI \cite{goldberger2000physiobank,PhysioNet-eegmmidb-1.0.0,schalk2004bci2000} serves as a representative multi-label-per-subject brain-state setting, where each subject contributes trials from multiple motor imagery states. The former tests whether representations organize different subjects according to disease-related neural patterns while preserving subject-specific stability, whereas the latter tests whether representations preserve stable individual physiological attributes while still encoding task-related state differences. Together, these two scenarios evaluate whether INCEPT captures subject-specific attributes, disease-related patterns and task-related state information beyond what can be inferred from quantitative downstream performance alone. All analyses use segment-level whole-brain embeddings before the final classifier. For INCEPT, whole-brain embeddings are obtained by averaging channel-level perception embeddings. For CBraMod \cite{wang2025cbramod}, a representative reconstruction-centered baseline, whole-brain embeddings are obtained by averaging patch-level embeddings produced by its backbone. We use t-SNE \cite{van2008visualizing} as a qualitative visualization of the embedding space, and use pairwise cosine-distance distributions computed from the learned embeddings as quantitative support for the subject-, class- and state-level relationships suggested by the visualization. Both analyses in Fig. \ref{fig:representation_analysis} are performed under linear probing and full fine-tuning protocols using a fixed seed.

\subsubsection{Subject-level organization in one-label-per-subject EEG analysis}
\label{sec:results-subject-representation}

ADFTD \cite{ds004504:1.0.8} provides a montage-matched brain-health evaluation setting, because its 19-channel 10-20 montage is close to the pre-training configuration. It also represents the subject-label structure common to many brain-health evaluation tasks in our benchmark, where EEG segments from the same subject are associated with one relatively stable disease- or condition-related category rather than multiple transient states. This setting complements the preceding downstream results, where montage-matched datasets often show strong performance even under linear probing. As a qualitative visualization of the embedding space, the t-SNE map suggests that INCEPT preserves subject-level structure under both linear probing and full fine-tuning: EEG segments from the same individual tend to occupy compact local regions, and subject centroids show qualitatively visible disease-related organization among Alzheimer's disease (AD), frontotemporal dementia (FTD) and healthy control subjects (HC) (Fig. \ref{fig:representation_analysis}a). This visual pattern is consistent with a frozen representation space that preserves individual stability while retaining disease-related variation. By contrast, CBraMod \cite{wang2025cbramod} also shows some local subject aggregation, but its clusters are more dispersed and its disease-related boundaries are less clearly organized, indicating weaker subject-level and disease-level structure in the feature space (Fig. \ref{fig:representation_analysis}c).

The pairwise cosine-distance analysis provides the quantitative test of this qualitative pattern. For INCEPT, within-subject distances are the smallest, distances between different subjects within the same disease-related category are intermediate, and distances between different classes are the largest (Fig. \ref{fig:representation_analysis}b). This hierarchy provides quantitative support for subject-level compactness and class-level separation, and it becomes more pronounced after full fine-tuning, which increases between-class separation without disrupting within-subject compactness. In CBraMod, the distance distributions are less separated under linear probing and rely more on full fine-tuning to recover disease-related structure (Fig. \ref{fig:representation_analysis}d). Thus, the cosine-distance hierarchy, rather than the visualization alone, supports the interpretation that INCEPT learns a brain-health representation space that jointly encodes subject-specific physiological consistency and disease-related differences.

\subsubsection{State-level organization in multi-label-per-subject EEG analysis}
\label{sec:results-state-representation}

PhysioNet-MI \cite{goldberger2000physiobank,PhysioNet-eegmmidb-1.0.0,schalk2004bci2000} provides a montage-shifted and multi-label-per-subject brain-state decoding setting, where each subject contributes trials from four motor imagery states. This structure reflects many state-decoding datasets in the preceding quantitative analysis, where EEG segments from the same individual may correspond to different affective, sleep, cognitive or sensorimotor states. Under linear probing, the qualitative t-SNE visualization suggests that INCEPT representations are organized primarily by subject (Fig. \ref{fig:representation_analysis}e). Trials from the same subject tend to occupy local regions, and the four motor imagery states from that subject are often embedded within the same subject-specific neighbourhood. In contrast to the one-label-per-subject ADFTD setting, task-related state offsets are less visually pronounced in the frozen qualitative embedding map, which is consistent with the more modest four-class motor imagery performance under linear probing. This pattern suggests that subject-specific physiological attributes remain a dominant source of structure in motor imagery EEG, whereas task-related neural activity may appear as a smaller bias relative to the individual baseline. CBraMod \cite{wang2025cbramod} shows a broadly similar subject-dominant organization, but its subject clusters are generally less compact and less consistently structured (Fig. \ref{fig:representation_analysis}g), in line with the weaker subject-level organization observed in the one-label-per-subject analysis. After full fine-tuning, INCEPT representations appear more interwoven across subjects in the qualitative map, and class-related trends become more visually apparent, with the both-feet class tending to shift towards one side of the embedding space (Fig. \ref{fig:representation_analysis}e). By contrast, CBraMod appears to further pull the four classes from the same subject closer together after full fine-tuning rather than reorganizing the space in an obviously task-directed manner, making the relationship between this qualitative geometry and its improved downstream performance less clear (Fig. \ref{fig:representation_analysis}g).

The cosine-distance distributions provide the quantitative counterpart to this qualitative visualization. For INCEPT, within-subject across-class pairs have slightly larger distances than within-subject same-class pairs under both linear probing and full fine-tuning, and a similar relative pattern is observed for across-subject pairs (Fig. \ref{fig:representation_analysis}f). This distance separation shows that motor imagery states introduce a measurable but modest shift on top of the subject-specific baseline, rather than replacing subject-level organization as the dominant source of structure. In CBraMod, this relative separation between same-class and different-class pairs is much less evident, with distance distributions for different task relationships remaining closer to each other (Fig. \ref{fig:representation_analysis}h). These quantitative distance relationships support the interpretation that INCEPT preserves subject-level stability while retaining more explicit task-related structure in the representation space.

\vspace{0.5em}
\noindent\textbf{\textit{Together,}} the one-label-per-subject and multi-label-per-subject analyses provide complementary evidence that INCEPT does not only improve downstream metrics, but also organizes EEG representations according to the subject-level and state-level structure of the underlying tasks.

\subsection{Cross-view invariance learning improves transfer and organizes subject-sensitive EEG representations}
\label{sec:results-invariance}
\label{sec:objective_ablation}

This section examines invariance learning from three perspectives:
\begin{itemize}
\item Section~\ref{sec:results-invariance-motivation} explains why the intrinsic subject-sensitivity of EEG motivates cross-view invariance learning.
\item Section~\ref{sec:results-objective-ablation} uses objective ablations to quantify whether invariance learning improves downstream transfer beyond masked reconstruction.
\item Section~\ref{sec:results-objective-representation} uses objective ablations to analyze how different objectives organize subject-aware and discriminative neural representations.
\end{itemize}

\subsubsection{Subject-sensitive nature of EEG motivates cross-view invariance learning}
\label{sec:results-invariance-motivation}
The representation analyses above suggest that subject-related structure is a major organizing axis of EEG representations. This observation is consistent with the signal nature of EEG. Scalp EEG is not a collection of interchangeable segments sampled from a subject-independent distribution; it contains subject-distinctive information shaped by stable individual physiology, including anatomical conduction pathways, oscillatory traits, functional coupling patterns and recording-specific baseline structure. Prior studies have shown that resting-state EEG contains subject-identifying information in spectral coherence and functional connectivity, and that EEG connectivity patterns are associated with individual phenotypic variation \cite{la2014human,nentwich2020functional}. Subject-specific structure is also present in more elementary signal properties, including the aperiodic component of the EEG power spectrum and peak individual alpha frequency, which provide stable subject-specific signatures across recording conditions \cite{demuru2020eeg,grandy2013peak}. Evidence of reliable individual differences and task-independent EEG signatures further supports the view that inter-individual variability is not merely measurement noise, but a reproducible component of EEG organization \cite{lopez2023stability,kumar2021evidence}. This subject-sensitive property is also reflected in practice: EEG-based brain-computer interfaces often require user-specific calibration, affective EEG models benefit from personalization, and seizure detection has long used patient-specific modelling strategies \cite{lotte2018review,wu2020transfer,zheng2016personalizing,shoeb2004patient}. These findings imply that a transferable EEG representation should be subject-sensitive rather than subject-agnostic, preserving stable within-subject structure while keeping disease-related, state-related or task-related differences discriminative.

This subject-sensitive nature motivates cross-view invariance learning as a subject-aware pre-training principle for EEG foundation modelling. Here, invariance does not mean removing subject identity or forcing different EEG views to match at the waveform level. Instead, it refers to representation-level stability across correlated observations of the same recording. Macro-level, micro-level and masked macro-level views may differ in local temporal content, electrode coverage, masking pattern and view-specific noise, but because they are derived from the same subject and recording context, they should still share essential subject-sensitive physiological structure and state-relevant neural information. This leads to a testable hypothesis: compared with masked reconstruction alone, a pre-training objective that aligns correlated EEG views should better preserve essential within-recording structure while maintaining discrimination across recordings, subjects, states and conditions. This hypothesis should be reflected not only in downstream transfer, but also in the geometry of the learned representation space.

\subsubsection{Quantitative objective ablations show transfer gains beyond reconstruction}
\label{sec:results-objective-ablation}

\begin{figure*}[!b]
    \centering
    \includegraphics[width=0.9\textwidth]{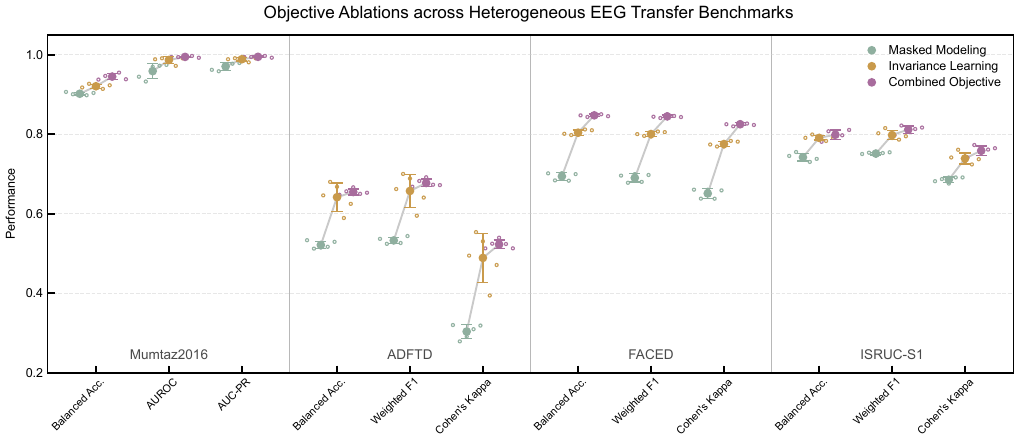}
    
    \caption{
        \textbf{Quantitative ablation of self-supervised pre-training objectives.} Masked modeling, invariance learning and the combined objective are compared under the same backbone and downstream full fine-tuning protocol. Points denote the mean across five random seeds, error bars indicate mean~$\pm$~s.d., and hollow circles show individual seed results.
    }
    \label{fig:ablation_quantitative}
\end{figure*}

We next test the hypothesis that cross-view invariance learning provides a stronger pre-training principle than reconstruction alone by comparing objective variants under matched architecture, pre-training data and downstream evaluation settings. The ablation includes three variants. The first variant, Masked Modeling, uses masked contextual modeling alone and therefore represents a reconstruction-centered objective. The second variant, Invariance Learning, removes the masked contextual objective and retains cross-view alignment between correlated EEG views. The third variant, Combined Objective, corresponds to the full INCEPT objective, in which masked contextual modeling and cross-view invariance learning are jointly optimized. This design isolates the role of the learning objective while keeping the backbone, pre-training corpus and downstream adaptation protocol fixed.

Fig. \ref{fig:ablation_quantitative} compares these variants on representative downstream transfer benchmarks. Mumtaz2016 and ADFTD provide montage-matched brain-health evaluations based on 19-channel 10-20 EEG, whereas FACED and ISRUC-S1 provide montage-shifted brain-state evaluations with different electrode densities, temporal contexts and label structures. This combination tests whether the pre-training objective only benefits tasks close to the pre-training electrode layout, or whether it also supports transfer to downstream settings with different montages and brain-state semantics.

Masked Modeling already yields useful downstream representations, confirming that local temporal, spectral and cross-channel dependencies provide an important contextual foundation for EEG pre-training. However, reconstruction alone does not consistently provide the strongest transfer. Invariance Learning improves over Masked Modeling across all evaluated datasets and metrics, with an average relative gain of 15.5\% across the 12 metric-dataset pairs. The improvement is modest but consistent on the near-ceiling Mumtaz2016 dataset, where Invariance Learning improves balanced accuracy, AUROC and AUC-PR by 2.1\%, 2.9\% and 1.9\%, respectively. The gains become more pronounced in settings where subject-level structure, montage shift or multi-class decision boundaries make transfer more difficult. On ADFTD, Invariance Learning improves balanced accuracy by 23.0\%, weighted F1 by 23.3\% and kappa by 60.9\% over Masked Modeling. On FACED, the corresponding gains are 15.8\%, 16.0\% and 19.1\%, while on ISRUC-S1 they are 6.6\%, 6.2\% and 7.8\%. These results indicate that aligning correlated EEG views provides a transferable global constraint that reconstruction alone does not fully capture.

The Combined Objective achieves the strongest overall performance across the ablation benchmarks. Relative to Masked Modeling, it improves balanced accuracy, AUROC and AUC-PR on Mumtaz2016 by 4.8\%, 3.7\% and 2.5\%, respectively. On ADFTD, it improves balanced accuracy by 25.5\%, weighted F1 by 27.1\% and kappa by 72.1\%. The gains are also substantial under montage shift, with improvements of 22.1\%, 22.4\% and 26.7\% on FACED, and 7.6\%, 8.0\% and 10.7\% on ISRUC-S1. Averaged across all 12 metric-dataset pairs, the Combined Objective improves over Masked Modeling by 19.4\%, and still improves over Invariance Learning alone by 3.2\%. This pattern suggests that masked contextual modeling and cross-view invariance learning are complementary rather than mutually exclusive.

The standard deviations further support this complementarity. Invariance Learning improves mean transfer over Masked Modeling, but its seed-to-seed variability remains dataset-dependent, especially in the subject-sensitive ADFTD setting. This sensitivity is plausible because ADFTD is a small one-label-per-subject brain-health dataset, where disease-related differences are distributed across resting-state EEG segments rather than tied to a single transient event. The Combined Objective mitigates this instability by anchoring cross-view alignment with local contextual recovery. Across the 12 metric-dataset pairs, it reduces the standard deviation relative to Invariance Learning in 10 metrics, with an average relative reduction of 33.1\%. This stabilizing effect is most evident on ADFTD and FACED, suggesting that the combined objective makes the gains from invariance learning more reproducible in subject-sensitive and affective EEG settings.


\subsubsection{Neural representation analysis reveals subject-aware and discriminative structure}
\label{sec:results-objective-representation}

\begin{figure*}[!b]
    \centering
    \setlength{\tabcolsep}{0pt}
    \renewcommand{\arraystretch}{1}

    \begin{tabular}{@{}c@{}}
    \scalingpanellabel{width=0.82\textwidth}{a}
        {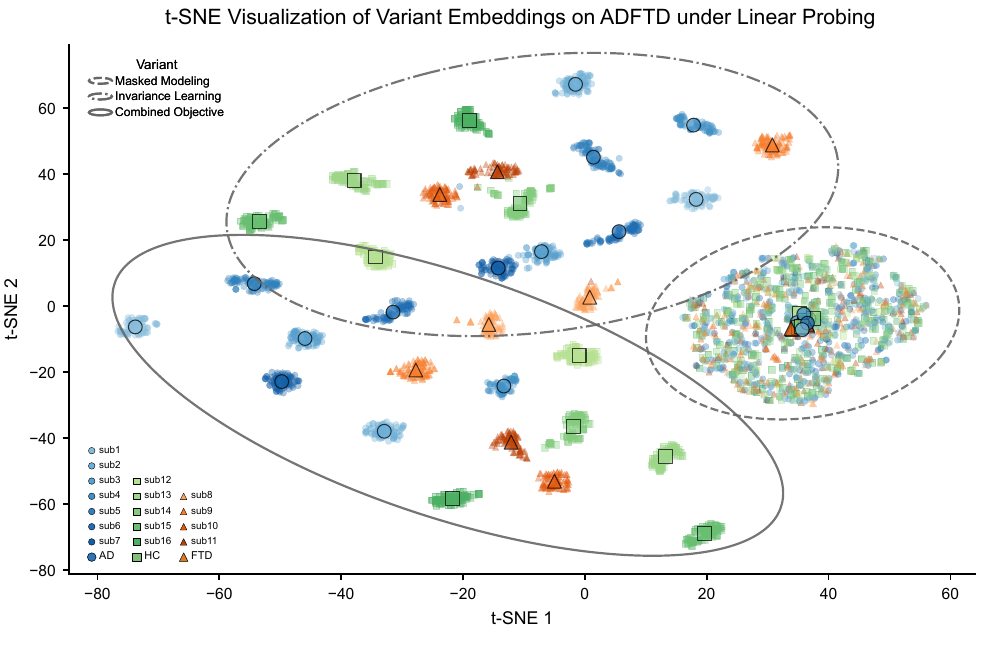}
    \\[-0.2em]
    \scalingpanellabel{width=0.82\textwidth}{b}
        {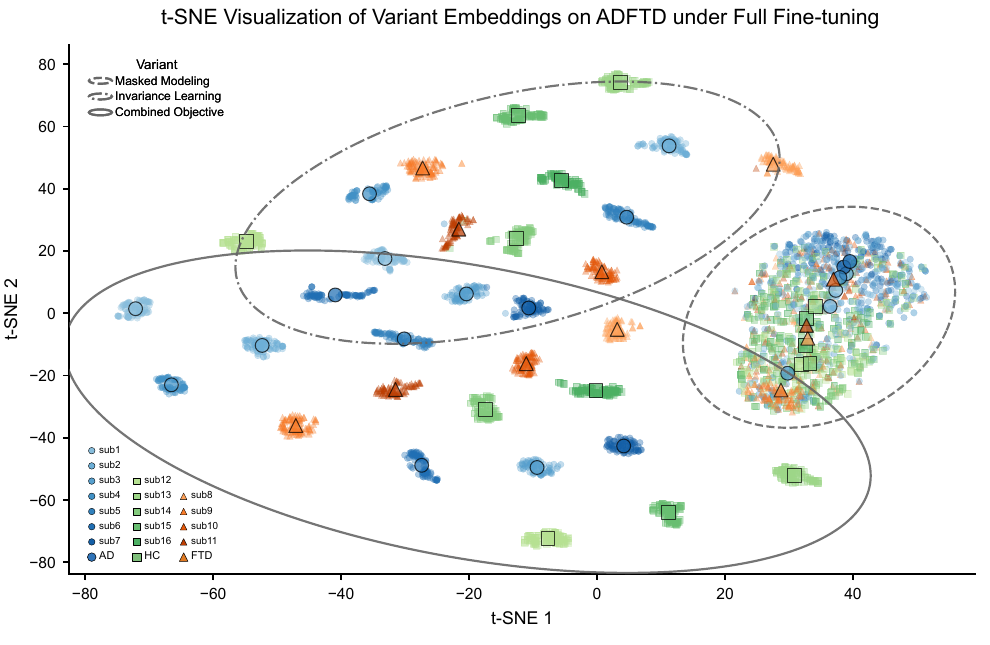}
    \\
    \end{tabular}
    
    \caption{
        \textbf{t-SNE visualization of pre-training objective ablation on ADFTD.}
        \textbf{a,} ADFTD representations under linear probing, where the backbone is frozen and the embedding geometry is determined by the pre-training objective.
        \textbf{b,} ADFTD representations after full fine-tuning, where the backbone is further adapted by downstream diagnostic labels.
        Each point denotes an EEG segment, colors denote diagnostic classes, including Alzheimer's disease (AD), healthy control (HC) and frontotemporal dementia (FTD), and marker shapes denote objective variants. Large markers indicate subject-level centroids. Dashed, dash-dotted and solid ellipses summarize the overall embedding regions of masked modeling, invariance learning and the combined objective, respectively.
    }
    \label{fig:ablation_tsne_analysis}
\end{figure*}

\begin{figure*}[!t]
    \centering
    \setlength{\tabcolsep}{0pt}
    \renewcommand{\arraystretch}{1}

    \begin{tabular}{@{}c@{}}
    \scalingpanellabel{width=0.82\textwidth}{a}
        {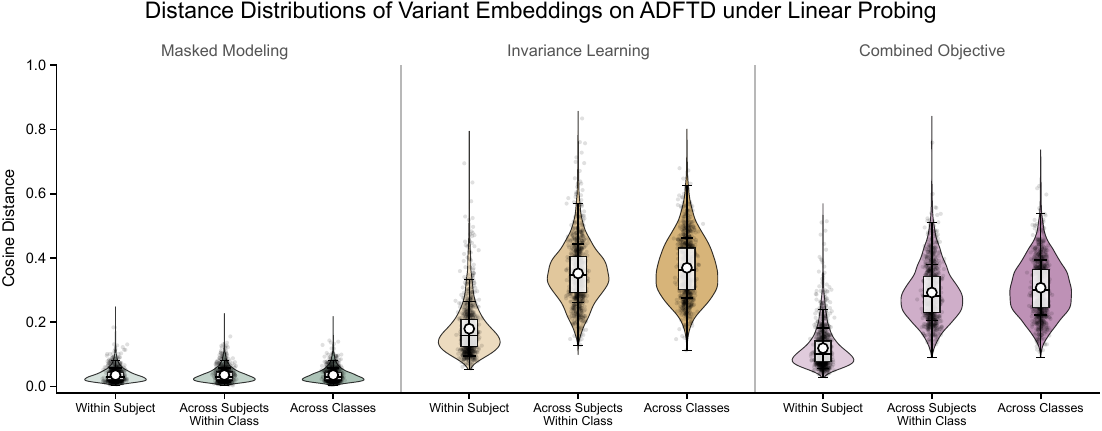}
    \\[-0.2em]
    \scalingpanellabel{width=0.82\textwidth}{b}
        {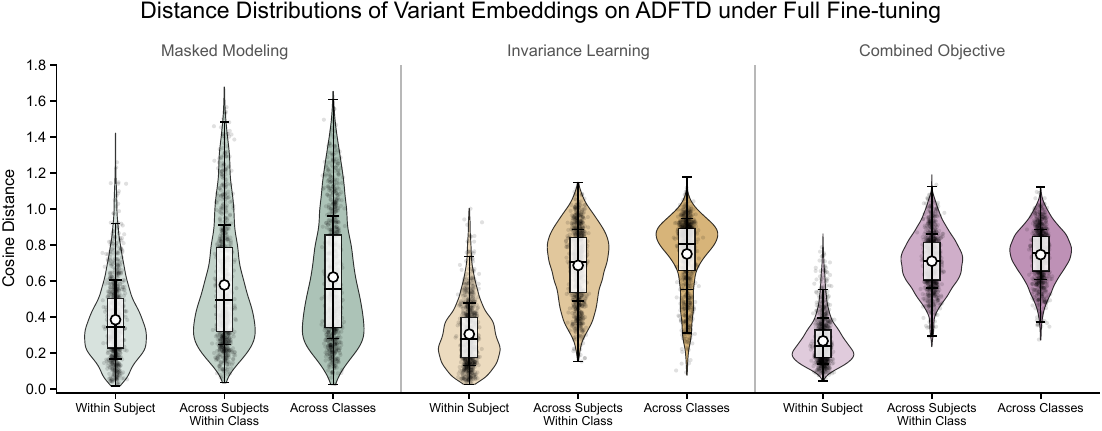}
    \\
    \end{tabular}
    
    \caption{
        \textbf{Pairwise-distance analysis of pre-training objective ablation on ADFTD.}
        \textbf{a,} Cosine-distance distributions of ADFTD representations under linear probing.
        \textbf{b,} Cosine-distance distributions after full fine-tuning.
        For each objective variant, distances are computed for segment pairs from the same subject, from different subjects within the same diagnostic class and from different diagnostic classes. Violin plots show the distribution of pairwise cosine distances, box plots summarize the median and interquartile range, and white circles with error bars indicate mean $\pm$ s.d..
    }
    \label{fig:ablation_distance_analysis}
\end{figure*}

We further examine whether the objective variants differ in the geometry of their learned neural representations. The embeddings are extracted in the same way as in the preceding representation analysis, using segment-level whole-brain embeddings before the final classifier. Here, however, the comparison is performed across the three objective variants under the same backbone and ADFTD evaluation setting. ADFTD is particularly suitable for this analysis because it provides a one-label-per-subject brain-health setting: EEG segments from the same subject share a stable diagnostic label, but also carry subject-specific physiological structure. A desirable representation should therefore preserve within-subject compactness, maintain partial similarity among subjects from the same diagnostic class and separate subjects across disease categories.

We use t-SNE only as a qualitative visualization of these relationships, and use pairwise cosine-distance distributions as the quantitative analysis of subject- and class-level organization. Under linear probing, the backbone is frozen and the representation geometry is determined entirely by the pre-training objective. This makes the linear-probing analysis a direct test of what each pre-training paradigm has organized before downstream labels can reshape the encoder. Masked Modeling treats each EEG segment primarily as an independent reconstruction target, without explicitly constraining different segments from the same subject to share a stable global representation. In the qualitative t-SNE visualization, this variant appears as a mixed embedding region in which subject- and class-level organization is less visually apparent (Fig.~\ref{fig:ablation_tsne_analysis}a). The quantitative distance analysis supports this observation: within-subject, across-subjects within-class and across-class distances are compressed into a narrow low-distance range, making these relationships difficult to distinguish (Fig.~\ref{fig:ablation_distance_analysis}a). By contrast, Invariance Learning and the Combined Objective produce more visible subject-level clusters under linear probing (Fig.~\ref{fig:ablation_tsne_analysis}a). Their distance distributions also follow a more interpretable hierarchy, with the smallest distances within subjects, intermediate distances across subjects within the same diagnostic class and larger distances across classes (Fig.~\ref{fig:ablation_distance_analysis}a). This hierarchy, rather than the visualization alone, supports the interpretation that cross-view invariance learning better preserves subject-sensitive structure while maintaining disease-related separability. Segments from the same subject should be closest because they share individual physiological attributes, subjects from the same diagnostic group should retain some disease-related similarity, and subjects from different classes should become more separable.

Under full fine-tuning, the backbone is further adapted by downstream diagnostic labels, and the representation space becomes more class-oriented. This effect is qualitatively visible across all three objective variants in the t-SNE maps, but it is especially apparent for Masked Modeling. Its linear-probing representation is weakly organized, whereas after full fine-tuning the embedding becomes more class-structured, with AD, HC and FTD showing clearer tendencies in different regions of the t-SNE space (Fig.~\ref{fig:ablation_tsne_analysis}b). The cosine-distance distributions provide the quantitative counterpart to this visual trend. After full fine-tuning, even Masked Modeling begins to show the expected ordering among within-subject, across-subjects within-class and across-class distances, indicating that downstream supervision can impose disease-related separability on a representation that was less organized when frozen (Fig.~\ref{fig:ablation_distance_analysis}b). For Invariance Learning and the Combined Objective, however, subject-aware structure is already present under linear probing, and full fine-tuning mainly sharpens the disease-related separation rather than creating the organization from scratch. Across these variants, the increase in across-class distances relative to within-subject distances indicates that downstream adaptation strengthens discriminative structure while retaining subject-level compactness.

\vspace{0.5em}
\noindent\textbf{\textit{Together,}} the quantitative ablation, qualitative embedding visualization and pairwise distance analysis support the same interpretation. Masked reconstruction provides useful local contextual supervision by encouraging the model to capture temporal continuity, spectral structure and cross-channel dependencies in EEG. However, not all locally recoverable signal structure is equally meaningful for downstream neural decoding. Artifacts, reference-related patterns, montage-dependent regularities and other recording-specific noise can also be predictable from local context, and reconstruction alone does not explicitly require segments from the same subject or recording context to share a subject-aware global representation. Cross-view invariance learning addresses this limitation by aligning correlated views at the representation level, thereby preserving stable subject-sensitive information while maintaining discrimination across subjects, states and diagnostic categories. The combined objective is therefore the most effective choice: masked contextual modeling retains fine-grained local EEG information, whereas invariance learning organizes the global representation around within-subject stability and between-condition separability. This complementarity explains why the combined objective improves transfer performance and makes the gains from invariance learning more reproducible in subject-sensitive EEG settings.

\section{Discussion and conclusion}
\subsection{INCEPT as a backbone for broad-spectrum post-acquisition EEG analysis}
This study positions INCEPT not as a model optimized for a single EEG benchmark, but as a reusable foundation backbone for broad-spectrum post-acquisition EEG analysis. Conventional EEG deep-learning pipelines are often organized around separate labelled datasets, with one model trained for one montage, cohort, label space and analysis goal. INCEPT instead uses large-scale unlabelled clinical EEG to learn representations that can be reused across signal-level assessment, brain-state decoding, and brain-health evaluation. This scope is important because it matches the way EEG is used after acquisition: recordings may need to be assessed for signal quality or abnormality, decoded for transient physiological states, or analysed for clinically relevant brain-health variation.

Across these application levels, INCEPT provides a common representational starting point for both linear probing and full fine-tuning. Linear probing tests whether downstream labels can read task-relevant information from a frozen backbone, whereas full fine-tuning tests whether the same backbone can be reshaped for new montages, temporal contexts and label semantics. The results support the view that EEG foundation models should be evaluated not only by peak performance on one task, but also by whether the same pre-trained representation remains readable, adaptable and stable across heterogeneous downstream settings.

A central implication is that EEG transferability should not be equated with removing subject-specific information. Scalp EEG is subject-sensitive: stable individual physiology, electrode geometry, oscillatory traits, functional coupling and recording-specific baseline structure all shape the observed signal. These factors complicate cross-subject and cross-dataset generalization, but they are also part of the neural organization that a transferable EEG representation should preserve. The representation analyses are consistent with this view: INCEPT retains stable within-subject structure while keeping disease-related, state-related and task-related variation accessible to downstream supervision. A useful EEG foundation model should therefore be subject-sensitive rather than subject-agnostic.

This subject-sensitive view also explains why reconstruction alone is an incomplete pre-training principle. Masked contextual modelling is valuable because it learns local temporal, spectral and cross-channel dependencies in continuous EEG. However, locally predictable EEG structure is not necessarily essential neural information: artifacts, reference effects, montage-specific regularities, device characteristics and acquisition-site patterns may also be predictable from surrounding context. INCEPT addresses this limitation by combining masked contextual modelling with cross-view invariance learning. By aligning the shared component across macro-level, micro-level and masked macro-level views, invariance-oriented contextual pre-training emphasizes essential subject-sensitive information that persists across observations of the same recording, encouraging compact and transferable whole-brain representations rather than representations organized only around signal recovery.

\subsection{Limitations and future directions}
Several limitations remain. First, EEG is intrinsically noisy and heterogeneous. Scalp recordings reflect neural activity together with volume conduction, scalp anatomy, electrode placement, reference scheme, behavioural state and environmental artifacts \cite{nunez1997eeg,jiang2019removal}. The present results suggest that subject-sensitive organization is an important component of transferable EEG representations, but the physiological interpretation of the learned structures remains incomplete. Future work should connect larger corpora and richer acquisition metadata with clinical neurophysiology and neuroscience interpretation, so that learned representations can be related more directly to stable neural mechanisms, longitudinal variation and expert-defined electrophysiological phenomena.

Second, benchmark performance does not by itself establish clinical utility. Public EEG datasets and current machine-learning benchmarks are often organized around retrospective classification labels, which enable controlled model comparison but rarely capture structured expert interpretation, longitudinal follow-up, treatment response, disease progression, diagnostic uncertainty or workflow-level outcomes. This limitation does not weaken the value of benchmark transfer evaluation, but it constrains what can be concluded clinically from such benchmarks alone. Future studies should complement retrospective evaluation with prospective, multi-centre and workflow-aware validation, especially for brain-health applications where model outputs must ultimately be interpreted in relation to expert review, clinical endpoints and added value in real diagnostic or monitoring workflows.

Third, the data and evaluation ecosystem for EEG foundation modelling remains insufficiently standardized. EEG datasets differ in montage, sampling rate, reference scheme, preprocessing, annotation protocol, hardware, cohort composition and task design. Larger unlabelled corpora will be valuable, but scale alone will not remove heterogeneity if metadata, quality-control procedures and evaluation protocols are inconsistent. Progress will require shared preprocessing practices, reporting standards, metadata harmonization, benchmark designs that test cross-centre and cross-montage transfer, and realistic evaluation settings that include sparse montages, limited labelled data and changing recording conditions.

\subsection{Conclusion}
Taken together, this study positions INCEPT as a reusable backbone for broad-spectrum post-acquisition EEG analysis. Through invariance-oriented contextual pre-training, INCEPT converts large-scale unlabelled clinical EEG into representations that are readable when frozen and adaptable when fine-tuned. It provides a common representational starting point for tasks that have traditionally been handled by separate dataset-specific models, spanning signal-level assessment, brain-state decoding, and brain-health evaluation.

INCEPT is not claimed as a final task-specific clinical system. Rather, it is a foundation backbone for building more transferable, subject-sensitive and reusable EEG analysis pipelines. The name reflects this role: INCEPT abbreviates INvariance-oriented Contextual EEG Pre-Training, while also invoking ``to incept'' as a beginning for the next stage of EEG foundation modelling. Reconstruction remains useful for learning local temporal, spectral and cross-channel structure, but the present results show that reconstruction alone is not sufficient as the sole organizing principle for reusable EEG representations. INCEPT therefore does not replace contextual recovery; by aligning correlated views, it emphasizes the essential subject-sensitive information shared across views rather than all locally reconstructable details, organizing contextual information into compact, subject-sensitive and transferable whole-brain representations. As larger EEG corpora, richer metadata and more clinically grounded validation resources become available, invariance-oriented contextual pre-training may provide a scalable route toward general-purpose EEG representation learning.

\section{Methods}
\subsection{The INCEPT model}

\subsubsection{Algorithm overview}
INCEPT, short for \underline{IN}variant \underline{C}ontextual \underline{E}EG \underline{P}re-\underline{T}raining, is a self-supervised pre-training framework designed to learn general-purpose EEG representations from unlabeled recordings. Given an EEG segment, INCEPT first uses a dynamic spatiotemporal sampler to construct multiple correlated views of the same neural activity at different spatial and temporal scales, including macro-level views for preserving global brain-state context, micro-level views for capturing localized neural dynamics, and masked macro-level views for encouraging contextual inference from partially observed signals \cite{caron2021emerging,oquab2024dinov2}. Each view is then passed to a neural token embedder, which converts raw EEG signals into structured tokens suitable for Transformer-based representation learning by encoding both signal content and electrode-specific context. On top of these tokens, INCEPT uses a shared Transformer \cite{vaswani2017attention} backbone to model structured spatiotemporal dependencies across different electrodes and temporal windows. During pre-training, a self-supervised invariance learner plays two coupled roles by enforcing global-local neural semantic invariance between macro- and micro-level views and missing-context invariance between macro-level and masked macro-level views. Together, these objectives enable INCEPT to learn from unlabeled EEG recordings without any task-specific labels, producing representations designed to transfer across EEG datasets, recording configurations and downstream prediction tasks.

\begin{figure*}[t]
    \centering
    \includegraphics[width=\textwidth]{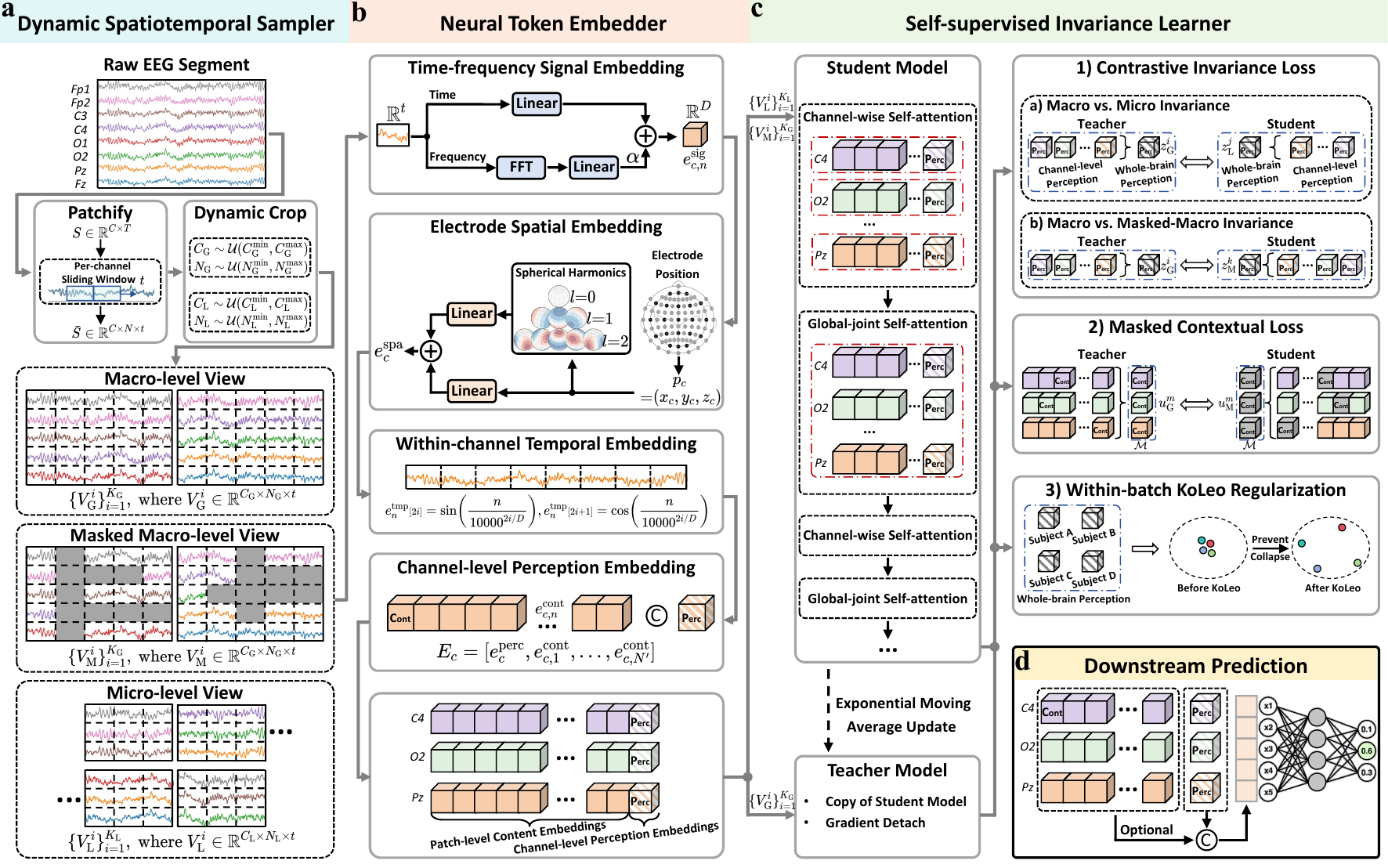}
    \caption{
    \textbf{Overview of the INCEPT pre-training framework and downstream prediction.}
    \textbf{a,} Dynamic spatiotemporal sampler. Given a raw EEG segment, INCEPT first patchifies each channel into non-overlapping temporal patches and then constructs correlated views with dynamic spatial and temporal extents. Macro-level views preserve broad brain-state context, micro-level views capture localized neural dynamics, and masked macro-level views introduce block-wise missing regions for contextual inference.
    \textbf{b,} Neural token embedder. Each sampled view is converted into structured EEG tokens by combining time-frequency signal embeddings, electrode spatial embeddings and within-channel temporal embeddings. Patch-level content embeddings retain local contextual information, whereas channel-level perception embeddings provide an explicit interface for summarizing neural activity at the electrode level.
    \textbf{c,} Self-supervised invariance learner. Student and teacher encoders share an alternating Transformer backbone that interleaves channel-wise self-attention for within-electrode temporal modeling with global-joint self-attention for cross-electrode contextual integration. During pre-training, contrastive invariance learning aligns perception embeddings across macro-micro and macro-masked-macro views, masked contextual learning aligns content embeddings between unmasked and masked macro-level views, and within-batch KoLeo regularization encourages whole-brain perception embeddings from different subjects to remain well distributed in latent space.
    \textbf{d,} Downstream prediction. For downstream tasks, INCEPT primarily uses channel-level perception representations, which first summarize electrode-level neural activity and can then be aggregated into compact whole-brain representations for classification. EEG patch content representations are retained as optional outputs for settings that require finer-grained contextual information.
    }
    \label{fig:pipeline}
\end{figure*}

\subsubsection{Dynamic spatiotemporal sampler}
We denote an EEG segment as $S \in \mathbb{R}^{C \times T}$, where $C$ represents the number of electrode channels and $T$ denotes the number of timestamps. Each channel is first divided into non-overlapping temporal patches of length $t$, yielding a patchified EEG representation $\bar{S} \in \mathbb{R}^{C \times N \times t}$, where $N=T/t$ is the number of temporal patches per channel. Each element in $\bar{S}$ therefore corresponds to a short waveform segment from a specific electrode and serves as the basic spatiotemporal unit for view sampling.

To accommodate the heterogeneity of EEG recordings across datasets, the sampler constructs views with dynamic spatial and temporal extents rather than using fixed-size crops \cite{caron2021emerging,oquab2024dinov2}. This design exposes the model during pre-training to variable electrode coverage and recording duration, thereby reducing its dependence on a specific montage or segment length. From the patchified representation $\bar{S}$, the sampler generates three types of correlated views: macro-level views, micro-level views and masked macro-level views. 1) Macro-level (global) views are intended to preserve broad neural context. For each macro-level view, the number of selected electrode channels and temporal patches are sampled as $C_{\mathrm{G}} \sim \mathcal{U}(C_{\mathrm{G}}^{\min}, C_{\mathrm{G}}^{\max})$ and $N_{\mathrm{G}} \sim \mathcal{U}(N_{\mathrm{G}}^{\min}, N_{\mathrm{G}}^{\max})$, respectively, after which a contiguous spatiotemporal region is cropped from $\bar{S}$ according to the sampled sizes. Repeating this procedure produces $\{V_{\mathrm{G}}^i\}_{i=1}^{K_{\mathrm{G}}}$, where $V_{\mathrm{G}}^i \in \mathbb{R}^{C_{\mathrm{G}} \times N_{\mathrm{G}} \times t}$ and $K_{\mathrm{G}}$ denotes the number of macro-level views. 2) Micro-level (local) views are generated in the same way but with smaller spatial and temporal ranges, with $C_{\mathrm{L}} \sim \mathcal{U}(C_{\mathrm{L}}^{\min}, C_{\mathrm{L}}^{\max})$ and $N_{\mathrm{L}} \sim \mathcal{U}(N_{\mathrm{L}}^{\min}, N_{\mathrm{L}}^{\max})$. The resulting set is denoted as $\{V_{\mathrm{L}}^i\}_{i=1}^{K_{\mathrm{L}}}$, where $V_{\mathrm{L}}^i \in \mathbb{R}^{C_{\mathrm{L}} \times N_{\mathrm{L}} \times t}$ and $K_{\mathrm{L}}$ is the number of micro-level views. These local views provide fine-grained observations that can be aligned with the broader neural context represented by the macro-level views. 3) Moreover, random block-wise masking is further applied to each macro-level view $V_{\mathrm{G}}^i$ to generate a corresponding masked macro-level view. Specifically, the mask removes contiguous regions of patches across electrodes and time while keeping the overall view size unchanged. This requires the model to infer the missing neural context from the remaining visible patches. The resulting masked macro-level views are denoted as $\{V_{\mathrm{M}}^i\}_{i=1}^{K_{\mathrm{G}}}$, where $V_{\mathrm{M}}^i \in \mathbb{R}^{C_{\mathrm{G}} \times N_{\mathrm{G}} \times t}$. In addition, we apply lightweight multi-band principal component analysis (PCA)-based augmentation to the sampled views, which perturbs non-dominant cross-channel components while preserving the dominant signal components within canonical frequency bands $\mathcal{B}=\{\mathrm{delta},\mathrm{theta},\mathrm{alpha},\mathrm{beta},\mathrm{gamma}\}$.

Together, these dynamically sampled macro-level views $\{V_{\mathrm{G}}^i\}_{i=1}^{K_{\mathrm{G}}}$, micro-level views $\{V_{\mathrm{L}}^i\}_{i=1}^{K_{\mathrm{L}}}$ and masked macro-level views $\{V_{\mathrm{M}}^i\}_{i=1}^{K_{\mathrm{G}}}$ provide the multi-scale inputs for INCEPT's self-supervised invariance learning objectives.

\subsubsection{Neural token embedder}
For each sampled view, the neural token embedder converts raw EEG patches into structured embeddings that jointly encode signal characteristics, electrode channel identity and within-channel timestamp. Let a view be denoted as $V \in \mathbb{R}^{C' \times N' \times t}$, where $C'$ and $N'$ are the numbers of electrodes and temporal patches in the view, respectively. Each raw patch token is denoted as $x_{c,n} \in \mathbb{R}^{t}$, with $c \in \{1,\ldots,C'\}$ indicating the electrode and $n \in \{1,\ldots,N'\}$ indicating the within-channel timestamp of the patch.

First, each raw patch token is encoded through two complementary branches to capture both temporal and spectral characteristics of the EEG signal. Given a raw patch token $x_{c,n} \in \mathbb{R}^{t}$, the time-domain branch directly projects the waveform into a $D$-dimensional latent space through a fully-connected layer, producing a temporal signal embedding $e_{c,n}^{\mathrm{time}} \in \mathbb{R}^{D}$. In parallel, the frequency-domain branch computes the amplitude spectrum of the same patch using the Fast Fourier Transform (FFT) \cite{cochran1967fast} and projects it into a spectral signal embedding $e_{c,n}^{\mathrm{freq}} \in \mathbb{R}^{D}$. The two embeddings are then integrated through an asymmetric fusion mechanism to form the signal embedding:
\begin{equation}
e_{c,n}^{\mathrm{sig}} = e_{c,n}^{\mathrm{time}} + \alpha \odot e_{c,n}^{\mathrm{freq}},
\end{equation}
where $\alpha \in \mathbb{R}^{D}$ is a learnable coefficient vector that adaptively controls the contribution of the auxiliary spectral features, and $e_{c,n}^{\mathrm{sig}}$ denotes the resulting signal embedding.

Electrode positional information is introduced to provide each patch with its channel-specific spatial context and to preserve the geometric relationships among scalp electrodes. Instead of assigning independent learnable positional embeddings to electrode identities \cite{yang2023biot,jiang2024large,chen2025uni} or using convolutional kernels to extract positional information dynamically from neighbouring tokens \cite{wang2025cbramod,NEURIPS2025_7e199ad8,ma2026codebrain}, we use a spherical-harmonics-based encoding because spherical harmonics provide continuous basis functions on the sphere and naturally contain parity structures \cite{seeley1966spherical,NIST:DLMF}, which are consistent with the continuous scalp topology and the symmetric-asymmetric organization of bilateral EEG electrodes. For the $c$-th electrode, we obtain its three-dimensional Cartesian coordinate $p_c=(x_c,y_c,z_c)$ from the EEG montage, normalize it onto a unit sphere, and convert it into spherical coordinates $(\theta_c,\phi_c)$. The spherical harmonic of degree $l$ and order $m$ is defined as
\begin{equation}
y_l^m(\theta,\phi)=\sqrt{\frac{2l+1}{4\pi}\frac{(l-m)!}{(l+m)!}}\,\mathcal{P}_l^m(\cos\theta)e^{im\phi},
\end{equation}
where $\mathcal{P}_l^m$ denotes the associated Legendre polynomial \cite{NIST:DLMF}. For each electrode, we evaluate all basis functions up to a maximum degree $L$ and concatenate them into a topology descriptor:
\begin{equation}
q_c=\left[y_l^m(\theta_c,\phi_c)\right]_{0\leq l\leq L,\,-l\leq m\leq l}\in\mathbb{R}^{(L+1)^2}.
\end{equation}
The descriptor $q_c$ is then projected into a $D$-dimensional embedding space to obtain the spherical-harmonics-based spatial embedding $e_c^{\mathrm{sh}}\in\mathbb{R}^{D}$. This branch captures electrode geometry after mapping the montage to an idealized unit sphere. However, the human scalp is not a perfect sphere, and absolute Cartesian coordinates can still provide useful localization cues. We therefore introduce a residual Cartesian pathway that directly projects the original, non-normalized coordinate $p_c$ into $e_c^{\mathrm{xyz}}\in\mathbb{R}^{D}$. The final electrode spatial embedding for the $c$-th electrode is given by
\begin{equation}
e_c^{\mathrm{spa}}=e_c^{\mathrm{sh}}+e_c^{\mathrm{xyz}},
\end{equation}
which provides each patch token with montage-aware spatial information while supporting variable electrode subsets across datasets.

To inform the model of the temporal position of each patch within an electrode channel, we further add a sinusoidal within-channel temporal embedding \cite{vaswani2017attention}. For the $n$-th patch in a channel, the within-channel temporal embedding is denoted as $e_n^{\mathrm{tmp}}\in\mathbb{R}^{D}$ and is defined by
\begin{equation}
e_n^{\mathrm{tmp}}{\scriptstyle[2i]}=\sin(\frac{n}{10000^{2i/D}}),\quad
e_n^{\mathrm{tmp}}{\scriptstyle[2i+1]}=\cos(\frac{n}{10000^{2i/D}}),
\end{equation}
where $i$ indexes the embedding dimension. This encoding provides each patch with its relative temporal position within the corresponding electrode channel in a parameter-free manner.
Thus, the final content embedding of each patch token combines the signal embedding, electrode spatial embedding and within-channel temporal embedding:
\begin{equation}
e_{c,n}^{\mathrm{cont}}=e_{c,n}^{\mathrm{sig}}+e_c^{\mathrm{spa}}+e_n^{\mathrm{tmp}}.
\end{equation}

In addition to patch-level content embeddings, we prepend a channel-level perception embedding to the token sequence of each electrode channel. This token provides a dedicated interface through which the model can form a channel-specific representation during subsequent modeling scheme. Specifically, for the $c$-th electrode channel, we introduce an additional learnable perception embedding $e_c^{\mathrm{perc}}\in\mathbb{R}^{D}$ and concatenate it with the patch-level content embeddings from the same channel:
\begin{equation}
E_c=[e_c^{\mathrm{perc}},e_{c,1}^{\mathrm{cont}},\ldots,e_{c,N'}^{\mathrm{cont}}]\in\mathbb{R}^{(N'+1)\times D},\quad c=1,\ldots,C'.
\end{equation}
Here, $e_c^{\mathrm{perc}}$ is not tied to any local temporal patch, but is associated with the entire signal segment of the $c$-th electrode channel. The embedded representation of the sampled view is then obtained by concatenating all channel-wise token sequences:
\begin{equation}
E=[E_1;E_2;\ldots;E_{C'}]\in\mathbb{R}^{C'(N'+1)\times D}.
\end{equation}
Thus, each electrode contributes one channel-level perception embedding together with its patch-level content embeddings, allowing the model to retain fine-grained signal, spatial and temporal information while providing an explicit embedding for channel-level context aggregation.

\subsubsection{Self-supervised invariance learner}
The embedded view sequence $E=[E_1;E_2;\ldots;E_{C'}]$ is encoded by an alternating Transformer \cite{vaswani2017attention} backbone that couples within-channel temporal modeling with inter-channel contextual integration. Channel-wise self-attention is applied independently to each electrode sequence $E_c=[e_c^{\mathrm{perc}},e_{c,1}^{\mathrm{cont}},\ldots,e_{c,N'}^{\mathrm{cont}}]$, whereas global-joint self-attention is applied to the full sequence $E$. This design first organizes local temporal dynamics within each electrode and then integrates distributed neural context across electrodes. By denoting the input to the Transformer as $H^0=E$, the alternating backbone is written as
\begin{equation}
H^{2r+1}=\mathcal{T}_{\mathrm{chan}}^{2r}\!\left(H^{2r};C'\right),\quad
H^{2r+2}=\mathcal{T}_{\mathrm{glob}}^{2r+1}\!\left(H^{2r+1}\right),
\end{equation}
where $r$ indexes Transformer pairs, $\mathcal{T}_{\mathrm{chan}}^{2r}(\cdot;C')$ denotes a Transformer layer with self-attention restricted within each electrode channel, and $\mathcal{T}_{\mathrm{glob}}^{2r+1}(\cdot)$ denotes a Transformer layer with self-attention over the full view sequence. Both layer types use the same residual Transformer block consisting of multi-head self-attention and a feed-forward network.

INCEPT uses this backbone in a teacher-student self-supervised scheme \cite{tarvainen2017mean}. Macro-level views are encoded by the teacher model, whereas paired micro-level views and masked macro-level views are encoded by the student model. The teacher is updated by an exponential moving average (EMA) \cite{tarvainen2017mean} of the student parameters, providing stable targets without direct gradient optimization. For each encoded view, the output representations of the channel-level perception tokens $\{e_c^{\mathrm{perc}}\}_{c=1}^{C'}$ are pooled to obtain a view-level whole-brain representation $z$, which is mapped by a prediction head into a probability distribution. We denote the teacher and student distributions as $p_t(z)$ and $p_s(z)$, respectively, with teacher targets normalized using Sinkhorn-Knopp centering \cite{sinkhorn1967concerning,caron2020unsupervised}. Let $z_{\mathrm{G}}^{i}$ be the teacher representation of the $i$-th macro-level view, and let $z_{\mathrm{L}}^{j}$ and $z_{\mathrm{M}}^{i}$ be the student representations of the $j$-th micro-level view and the masked counterpart of the $i$-th macro-level view. The contrastive invariance loss is
\begin{equation}
\mathcal{L}_{\mathrm{contrast}}
=
\frac{1}{K_{\mathrm{G}}K_{\mathrm{L}}}
\sum_{i=1}^{K_{\mathrm{G}}}\sum_{j=1}^{K_{\mathrm{L}}}
\mathrm{CE}\Bigl(p_t(z_{\mathrm{G}}^{i}),p_s(z_{\mathrm{L}}^{j})\Bigr)
+
\frac{1}{K_{\mathrm{G}}(K_{\mathrm{G}}-1)}
\sum_{i=1}^{K_{\mathrm{G}}}\sum_{\substack{k=1\\k\neq i}}^{K_{\mathrm{G}}}
\mathrm{CE}\Bigl(p_t(z_{\mathrm{G}}^{i}),p_s(z_{\mathrm{M}}^{k})\Bigr),
\end{equation}
where $\mathrm{CE}(\cdot,\cdot)$ denotes cross-entropy. The first term aligns micro-level views with macro-level contextual targets, and the second term aligns unmasked macro-level contextual targets with alternative masked macro-level views from the same EEG segment.

To further constrain local representations, INCEPT applies a masked contextual loss to the masked macro-level views. Each patch-level content embedding $e_{c,n}^{\mathrm{cont}}$ is contextualized by the Transformer, yielding an output token representation $u_{c,n}$. Let $\mathcal{M}$ be the set of masked token locations. For each masked location, the student token prediction from the masked macro-level view is matched to the corresponding teacher token target from the unmasked macro-level view. The masked contextual loss is
\begin{equation}
\mathcal{L}_{\mathrm{masked}}
=
\frac{1}{|\mathcal{M}|}
\sum_{m\in\mathcal{M}}
\mathrm{CE}\Bigl(p_t(u_{\mathrm{G}}^{m}),p_s(u_{\mathrm{M}}^{m})\Bigr),
\end{equation}
where $u_{\mathrm{G}}^{m}$ and $u_{\mathrm{M}}^{m}$ are the teacher and student token representations at the $m$-th masked location. A KoLeo regularization term \cite{kozachenko1987sample,oquab2024dinov2} is also applied to encourage a more uniform distribution of student view-level representations. The final pre-training objective is
\begin{equation}
\mathcal{L}
=
\lambda_{1}\mathcal{L}_{\mathrm{contrast}}
+
\lambda_{2}\mathcal{L}_{\mathrm{masked}}
+
\lambda_{3}\mathcal{L}_{\mathrm{KoLeo}}.
\end{equation}
This objective trains INCEPT to align EEG representations across spatiotemporal scales, recover missing contextual information and maintain a well-distributed latent space.

\subsection{Pre-training and evaluation datasets}

\subsubsection{Pre-training dataset}
Pre-training is performed on the Temple University Hospital EEG Corpus (TUEG), a large-scale public archive of clinical EEG recordings \cite{obeid2016temple}. The corpus contains 69,672 EDF files from 14,987 patients across 26,846 recording sessions, with a total duration of approximately 27,077 hours. The recordings span more than 40 channel configurations, with sampling frequencies ranging from 250 to 1,024 Hz, and most recordings include electrodes from the international 10-20 system. The large scale, subject-level heterogeneity and broad clinical coverage of TUEG make it a valuable resource for self-supervised EEG pre-training. Nevertheless, since TUEG is a routine clinical archive with substantial unmarked noise, artifacts and faulty channels, we apply an automated pre-processing pipeline for quality control.

For quality control, we first standardize channel labels and retain recordings containing 19 common electrodes from the international 10-20 system: Fp1, Fp2, F3, F4, C3, C4, P3, P4, O1, O2, F7, F8, T3, T4, T5, T6, Fz, Cz and Pz. Recordings shorter than 5 min are excluded, and the first and last minute of each retained recording are discarded to reduce boundary-related noise. The remaining signals are assigned a standard 10-20 montage and re-referenced to linked ears when electrodes A1 and A2 are available, or to the common average otherwise.
A notch filter at 60 Hz is used to suppress power-line noise, followed by a 0.1-50 Hz band-pass filter to remove slow drifts and high-frequency noise. All signals are resampled to 250 Hz. To further reduce stereotyped non-neural components, we apply independent component analysis (ICA) \cite{delorme2004eeglab} with automated ICLabel annotation \cite{pion2019iclabel} and remove components labeled as non-brain sources. The preprocessed recordings are divided into non-overlapping 30-s windows. Windows are excluded if any sample exceeds an absolute amplitude of 120 $\mu$V, and each retained segment is normalized by channel-wise Z-score normalization.

\subsubsection{Downstream evaluation datasets}
We evaluate INCEPT on ten downstream EEG datasets spanning clinical diagnosis, cognitive-state decoding, seizure detection, motor imagery, sleep staging and emotion recognition. Detailed dataset statistics, demographic information, preprocessing procedures and train-valid-test splits are provided in Supplementary Dataset information (Supplementary Section~\ref{supplement:dataset}). The benchmark includes both montage-matched datasets based on 19-channel 10-20 EEG layouts and montage-shifted datasets with different channel densities and electrode configurations, allowing us to assess transferability across task domains as well as acquisition geometries.

The downstream evaluation is organized according to three levels of post-acquisition EEG analysis, from signal quality assessment to brain-state decoding, and brain-health evaluation. At the signal quality level, \textbf{TUAB}~\cite{obeid2016temple}, a subset of TUEG, is used for binary abnormal EEG detection, with recordings labeled as normal or abnormal, and \textbf{TUAR}~\cite{obeid2016temple} is used for four-class artifact recognition, distinguishing eye movement, muscle artifact, electrode-related artifact and background EEG segments. Both datasets use routine clinical EEG layouts close to the 19-channel 10-20 montage used during pre-training, providing signal-level evaluations under a relatively montage-matched setting.

At the brain-state level, the benchmark includes affective, sensorimotor and sleep-state decoding tasks. \textbf{FACED}~\cite{chen2023large} is used for nine-class emotion recognition from 32-channel EEG, and \textbf{SEED-V}~\cite{liu2021comparing} is used for five-class emotion recognition from 62-channel EEG. \textbf{PhysioNet-MI}~\cite{PhysioNet-eegmmidb-1.0.0,schalk2004bci2000,goldberger2000physiobank} is used for four-class motor imagery classification, including left fist, right fist, both fists and both feet. \textbf{ISRUC-S1}~\cite{khalighi2016isruc} is used for five-class sleep staging from sparse sleep EEG derivations. These datasets differ substantially from the 19-channel pre-training montage, covering low-density, high-density and sparse electrode configurations, and therefore evaluate whether the pre-trained representation can transfer to montage-shifted brain-state decoding settings.

At the brain-health level, the benchmark evaluates psychiatric, cognitive and neurological conditions. \textbf{Mumtaz2016}~\cite{mumtaz2016mdd,mumtaz2018machine} is used for binary depression-related classification, distinguishing major depressive disorder from healthy control. \textbf{MentalArithmetic}~\cite{zyma2019electroencephalograms} is used for binary mental stress detection, distinguishing resting-state EEG from EEG recorded during mental arithmetic. \textbf{ADFTD}~\cite{ds004504:1.0.8} is used for three-class neurodegenerative disease classification among Alzheimer's disease, frontotemporal dementia and healthy control subjects.  These three datasets use 19-channel 10-20 EEG layouts and therefore provide montage-matched brain-health evaluations. \textbf{Siena}~\cite{PhysioNet-siena-scalp-eeg-1.0.0,detti2020eeg,goldberger2000physiobank} is used for binary seizure-related neurological assessment from an extended 29-channel clinical EEG montage, providing a montage-shifted brain-health setting. Together, these datasets test whether a single pre-trained EEG encoder can support signal-level quality control, transient brain-state decoding and clinically relevant brain-health assessment across both montage-matched and montage-shifted conditions.

\subsection{Experimental settings}
For self-supervised pre-training, the temporal patch length is set to $t=250$, corresponding to a 1-s EEG window sampled at 250 Hz, and each raw patch is projected by the neural embedder into a 768-dimensional token embedding. The encoder uses 12 Transformer layers. For each 30-s EEG segment, the sampler generates $K_{\mathrm{G}}=2$ macro-level views and $K_{\mathrm{L}}=8$ micro-level views. The macro-level sampling ranges are set to $C_{\mathrm{G}}^{\min}=17$, $C_{\mathrm{G}}^{\max}=19$, $N_{\mathrm{G}}^{\min}=24$ and $N_{\mathrm{G}}^{\max}=30$, whereas the micro-level ranges are set to $C_{\mathrm{L}}^{\min}=12$, $C_{\mathrm{L}}^{\max}=15$, $N_{\mathrm{L}}^{\min}=6$ and $N_{\mathrm{L}}^{\max}=12$. Masked macro-level views are generated by block-wise masking with a sampling probability of 0.5 and a masking ratio sampled from 0.1 to 0.3. The teacher is updated by exponential moving average (EMA), with the momentum increased from 0.994 to 1.0 during training. The prediction head maps both teacher and student outputs, $p_t$ and $p_s$, to 8,192-dimensional probability distributions. The contrastive invariance loss and masked reconstruction loss are assigned equal weights, with $\lambda_{1}=1.0$ and $\lambda_{2}=1.0$, and the KoLeo regularization weight is set to $\lambda_{3}=0.1$. Teacher targets are normalized using Sinkhorn-Knopp centering. Optimization uses AdamW with $\beta_1=0.9$ and $\beta_2=0.999$. The learning rate is set to $8\times10^{-4}$ with square-root batch-size scaling, is warmed up for 37,500 iterations and then decayed to $1\times10^{-6}$ over a total of 250,000 iterations. Pre-training uses a batch size of 96 per GPU and mixed-precision distributed training on two 80-GB NVIDIA A100 GPUs.

For downstream evaluation, we transfer the pre-trained teacher encoder to each target dataset and train a task-specific classifier under two protocols: full fine-tuning and linear probing. In the full fine-tuning setting, the backbone encoder and classifier are optimized jointly for 50 epochs using AdamW, with learning rates of $1\times10^{-4}$ for the backbone and $1\times10^{-3}$ for the classifier. The classifier consists of two fully connected layers with exponential linear unit (ELU) activations and dropout, followed by the output layer. In the linear-probing setting, the encoder is frozen and only a single linear classifier is optimized for 50 epochs using AdamW with a learning rate of 0.01. Both protocols use a batch size of 64, a weight decay of 0.1 and label smoothing of 0.1. For the ISRUC-S1 sleep-staging task, we follow prior work and add a Transformer layer above the classification head to model dependencies across consecutive sleep epochs. Each downstream experiment is repeated with five random seeds, 0, 42, 1024, 1234 and 3407, with all experiments conducted on a single 80-GB NVIDIA A100 GPU.

\section*{Data availability}
The pre-training dataset, the Temple University Hospital EEG Corpus (TUEG), is available through the Neural Engineering Data Consortium (NEDC) subject to the corresponding data-use and access procedures (\url{https://isip.piconepress.com/projects/tuh_eeg}). Regarding the downstream validation datasets, the TUH Abnormal EEG Corpus (TUAB), used as a downstream abnormal EEG detection dataset in this study, is a subset of TUEG and is available through the same TUH EEG resource (\url{https://isip.piconepress.com/projects/tuh_eeg}). The EEG During Mental Arithmetic Tasks dataset (MentalArithmetic) is available online from PhysioNet (\url{https://physionet.org/content/eegmat/1.0.0}); the MDD Patients and Healthy Controls EEG Data dataset used for depression-related classification (Mumtaz2016) is available online from Figshare (\url{https://figshare.com/articles/dataset/EEG_Data_New/4244171}); the dataset of EEG recordings from Alzheimer's disease, frontotemporal dementia and healthy subjects dataset (ADFTD) is available online from OpenNeuro (\url{https://openneuro.org/datasets/ds004504/versions/1.0.8}); the Finer-grained Affective Computing EEG Dataset (FACED) is available through Synapse (\url{https://www.synapse.org/Synapse:syn50614194}); the SEED-V dataset (SEED-V) is available from the BCMI laboratory at Shanghai Jiao Tong University (\url{https://bcmi.sjtu.edu.cn/home/seed/seed-v.html}); the EEG Motor Movement/Imagery Dataset used for motor imagery classification (PhysioNet-MI) is available online from PhysioNet (\url{https://physionet.org/content/eegmmidb/1.0.0}); the ISRUC-Sleep Cohort I dataset (ISRUC-S1) is available through the ISRUC-Sleep database (\url{https://sleeptight.isr.uc.pt}); and the Siena Scalp EEG database (Siena) is available online from PhysioNet (\url{https://physionet.org/content/siena-scalp-eeg/1.0.0}). The derived train, validation and testing splits used in this study, together with preprocessing metadata where permitted by the original dataset licences, will be released with the code repository.

\section*{Code availability}
Code will be released upon acceptance. The public repository will include preprocessing scripts for all datasets used in this study, the implementation of the INCEPT model architecture, self-supervised pre-training code and the downstream evaluation code. Code will also be made available during the peer-review process if required by the editors or reviewers.


\begin{appendices}





\section{Dataset information}\label{supplement:dataset}
\subsection{Dataset Overview}
We use one large-scale unlabeled clinical EEG corpus for self-supervised pre-training and ten downstream EEG datasets for evaluation. The downstream datasets cover abnormal EEG detection, depression-related classification, neurodegenerative disease classification, mental stress detection, seizure detection, motor imagery classification, sleep staging and emotion recognition. Together, these datasets span clinical, cognitive, sleep, motor-imagery and affective EEG scenarios, with substantial variation in cohort composition, acquisition setting, electrode montage, channel density, recording duration and label structure. This heterogeneous benchmark is designed to assess whether INCEPT learns transferable EEG representations rather than representations optimized for a single dataset or montage.

Table~\ref{tab:dataset_summary} provides the task-level signal settings and available demographic metadata for the pre-training corpus and all downstream evaluation datasets. Because public EEG datasets differ in metadata completeness, unavailable demographic information is marked with ``/''. For datasets with available subject-level metadata, age is reported either by sex or diagnostic group, depending on how the original metadata are organized.

\subsection{Dataset Details}
\textbf{TUEG pre-training corpus} \cite{obeid2016temple} is used as the unlabeled corpus for self-supervised pre-training. The Temple University Hospital EEG Corpus is a large public clinical EEG archive collected in routine hospital settings and contains recordings from 14,987 subjects. In the available subject-level metadata, male participants have a mean age of 49.3 $\pm$ 20.1 years and female participants have a mean age of 50.1 $\pm$ 20.5 years. The dataset contains 48.8\% male participants. Detailed diagnostic status, cognitive measurements and education information are not available. For pre-training, we retain recordings containing the 19 common international 10-20 electrodes and standardize channel labels across recordings. Recordings shorter than 5 min are excluded, and the first and last minute of each retained recording are discarded to reduce boundary-related noise. The remaining signals are assigned a standard 10-20 montage, re-referenced to linked ears when A1 and A2 are available or to the common average otherwise, notch-filtered at 60 Hz, band-pass filtered from 0.1 to 50 Hz and resampled to 250 Hz. Independent component analysis with automated ICLabel annotation is used to reduce stereotyped non-brain components. The preprocessed recordings are divided into non-overlapping 30-s windows, and each retained segment is normalized by channel-wise z-score normalization.

\textbf{TUAB} \cite{obeid2016temple} is used for binary abnormal EEG detection. It is a clinical EEG subset derived from the Temple University Hospital EEG corpus, with recordings labeled as normal or abnormal. Separate age and sex statistics are not available in the metadata. EEG signals are originally recorded at 256 Hz using channels from the international 10-20 system. Signals are band-pass filtered between 0.3 and 75 Hz and notch-filtered at 60 Hz. Each recording is divided into non-overlapping 10-s windows, yielding 409,455 EEG segments from 2,383 subjects. We follow the official split, using the official evaluation split for testing and further partitioning the official training split into training and validation sets at a subject-level 80\%:20\% ratio within each class.

\textbf{TUAR} \cite{obeid2016temple} is used for segment-level artifact recognition in EEG signal quality control. The dataset contains routine clinical EEG recordings from the Temple University Hospital EEG resource with time-resolved artifact annotations. The original annotations include five artifact types, namely eye movement, chewing, shivering, muscle artifact and electrode-related artifact, together with background EEG. Because chewing and shivering contain very few usable samples, we exclude these two categories and formulate TUAR as a four-class task, including eye movement artifact, muscle artifact, electrode-related artifact and background EEG. EEG signals are processed using the 19-channel 10-20 layout, notch-filtered at 60 Hz, band-pass filtered between 0.3 and 75 Hz.Recordings are divided into non-overlapping 5-s windows, and segments containing multiple retained artifact labels are excluded to keep the class definition unambiguous. This procedure yields 57,392 four-class EEG segments, including 34,829 background segments, 9,066 eye-movement segments, 8,159 muscle-artifact segments and 5,338 electrode-related artifact segments. We use a subject-level split with an 80\%:10\%:10\% ratio for training, validation and testing datasets.

\textbf{Mumtaz2016} \cite{mumtaz2016mdd,mumtaz2018machine} is used for binary depression-related EEG classification. The dataset contains EEG recordings from subjects with major depressive disorder and healthy controls. In the available metadata, the major depressive disorder group has a mean age of 40.3 $\pm$ 12.9 years and the healthy control group has a mean age of 38.3 $\pm$ 15.6 years; 59.4\% of participants are male. BDI-II and HADS are used in the original study to assess depression severity, although the specific scores are not available in the compiled metadata used here. EEG signals are originally recorded at 256 Hz using 19 channels from the international 10-20 system. Signals are band-pass filtered between 0.3 and 75 Hz and notch-filtered at 50 Hz. Recordings are segmented into non-overlapping 5-s windows, yielding 7,143 samples. We use a subject-level split, assigning 22 patients with major depressive disorder and 21 healthy controls to the training set, 5 patients and 4 controls to the validation set, and 6 patients and 5 controls to the testing set.

\textbf{ADFTD} \cite{ds004504:1.0.8} is used for three-class neurodegenerative disease classification among Alzheimer's disease, frontotemporal dementia and healthy control subjects. The dataset contains 88 subjects, including 36 subjects with Alzheimer's disease, 23 subjects with frontotemporal dementia and 29 healthy controls. The mean age is 66.4 $\pm$ 7.9 years for Alzheimer's disease, 63.6 $\pm$ 8.2 years for frontotemporal dementia and 67.9 $\pm$ 5.4 years for healthy controls. The cohort is balanced by sex at the full-dataset level, with 50.0\% male participants. MMSE information is available and reflects the expected cognitive-severity differences among the diagnostic groups. EEG signals are originally recorded at 500 Hz using 19 channels from the international 10-20 system. Recordings are segmented into non-overlapping 10-s windows, yielding 6,948 samples. We use a subject-level split within each class, assigning 60\% of subjects to the training set, 20\% to the validation set and 20\% to the testing set.

\textbf{MentalArithmetic} \cite{zyma2019electroencephalograms} is used for binary mental stress detection. EEG recordings are labeled according to whether subjects are at rest or performing mental arithmetic. The dataset contains 36 participants, with 25.0\% male participants. Male participants have a mean age of 19.7 $\pm$ 3.5 years and female participants have a mean age of 17.8 $\pm$ 1.3 years. All participants are students of the Educational and Scientific Centre ``Institute of Biology and Medicine'', National Taras Shevchenko University of Kyiv, Ukraine. The original dataset also includes a performance grouping, with 24 subjects classified as good mental-arithmetic performers and 12 as poor performers. EEG signals are originally recorded at 500 Hz using 19 channels from the international 10-20 system. Recordings are segmented into non-overlapping 5-s windows, yielding 1,707 samples. We use a subject-level split, assigning 28 subjects to the training set, 4 subjects to the validation set and 4 subjects to the testing set.

\textbf{Siena} \cite{PhysioNet-siena-scalp-eeg-1.0.0,detti2020eeg,goldberger2000physiobank} is used for binary seizure detection. EEG segments are labeled as normal or seizure according to expert seizure annotations. The dataset contains scalp EEG recordings from 14 epilepsy patients monitored at the Unit of Neurology and Neurophysiology of the University of Siena. Male participants have a mean age of 47.3 $\pm$ 13.8 years and female participants have a mean age of 38.5 $\pm$ 14.4 years; 57.1\% of participants are male. Cognitive-measurement and education information are not reported. EEG signals are originally recorded at 512 Hz using 29 channels from an extended 10-20 montage, including Fc1/Fc2, Fc5/Fc6, Cp1/Cp2, Cp5/Cp6 and F9/F10 in addition to the common 19-channel 10-20 montage. Signals are band-pass filtered between 0.1 and 75 Hz and notch-filtered at 50 Hz. Recordings are first divided into non-overlapping 10-s windows, and seizure segments are additionally augmented using a 10-s sliding window with a 5-s step around annotated seizure intervals. This procedure yields 51,697 samples. We use a subject-level split, assigning 10 subjects to the training set, 2 subjects to the validation set and 2 subjects to the testing set.

\textbf{PhysioNet-MI} \cite{goldberger2000physiobank,PhysioNet-eegmmidb-1.0.0,schalk2004bci2000} is used for four-class motor imagery classification, including left fist, right fist, both fists and both feet. The dataset contains 109 healthy volunteers. Age and sex information are not reported in the available metadata. EEG signals are originally recorded at 160 Hz using 64 electrodes from a high-density 10-10 system. Signals are high-pass filtered at 0.1 Hz and notch-filtered at 60 Hz. Motor imagery trials are extracted from the task annotations, with each trial represented as a 4-s segment. The final dataset contains 9,837 samples. We use a subject-level split, assigning 70 subjects to the training set, 19 subjects to the validation set and 20 subjects to the testing set.

\textbf{ISRUC-S1} \cite{khalighi2016isruc} is used for five-class sleep staging. Each epoch is labeled as Wake, N1, N2, N3 or rapid eye movement sleep according to expert annotations. The dataset contains 100 subjects from a sleep clinical cohort, with 56.0\% male participants. Male participants have a mean age of 52.3 $\pm$ 17.4 years and female participants have a mean age of 49.6 $\pm$ 14.1 years. Diagnosis categories include sleep apnea obstructive syndrome, roncopatia, affective disorder, sleep deprivation, REM sleep behaviour disorder, periodic limb movements during sleep, epilepsy, restless legs, parasomnia and sleep-related ventilation abnormality syndrome. EEG signals are originally recorded at 200 Hz using six channels from standard sleep EEG derivations. Signals are band-pass filtered between 0.3 and 35 Hz and notch-filtered at 50 Hz. Recordings are divided into 30-s epochs, yielding 88,340 samples. For sequence-to-sequence sleep-stage classification, epochs are further grouped into non-overlapping sequences of 20 epochs. We use a subject-level split, assigning 80 subjects to the training set, 10 subjects to the validation set and 10 subjects to the testing set.

\textbf{FACED} \cite{chen2023large} is used for nine-class emotion recognition, covering anger, disgust, fear, sadness, neutral emotion, amusement, inspiration, joy and tenderness. The dataset contains 123 subjects, with 39.0\% male participants. Male participants have a mean age of 23.5 $\pm$ 5.1 years and female participants have a mean age of 23.0 $\pm$ 4.4 years. Health status, cognitive measurements and education information are not reported in the compiled metadata. EEG signals are originally recorded at 250 Hz using 32 electrodes from a low-density 10-10 system. Recordings are segmented according to video trials and divided into non-overlapping 30-s windows, yielding 3,444 samples. We use a subject-level split over 123 subjects, assigning 80 subjects to the training set, 20 subjects to the validation set and 23 subjects to the testing set.

\textbf{SEED-V} \cite{liu2021comparing} is used for five-class emotion recognition, with EEG recordings labeled as happy, sad, neutral, disgust or fear. The dataset contains 16 subjects, with 37.5\% male participants. Male participants have a mean age of 21.2 $\pm$ 1.7 years and female participants have a mean age of 21.9 $\pm$ 1.6 years. All participants are students at Shanghai Jiao Tong University. Health status and cognitive-measurement information are not reported in the compiled metadata. EEG signals are originally recorded at 1,000 Hz using a 62-channel ESI NeuroScan system. Signals are band-pass filtered between 0.1 and 50 Hz and notch-filtered at 50 Hz. Task periods are segmented into non-overlapping 4-s windows, yielding 29,168 samples. For each session, we use a trial-level split, assigning 5 trials to the training set, 5 trials to the validation set and 5 trials to the testing set.

For all downstream datasets, EEG signals are resampled to 250 Hz before model input construction, consistent with the temporal resolution used during self-supervised pre-training. Dataset-specific filtering, segmentation and split strategies are retained when required by the original dataset protocol or downstream task definition. After segmentation, each input segment is normalized by channel-wise z-score normalization before being fed into the network. These shared preprocessing steps reduce implementation differences across datasets while preserving the clinically or experimentally defined structure of each benchmark.

\section{Experimental details}
\subsection{Baseline models}
We compare INCEPT with two groups of baseline models: task-specific supervised EEG encoders and EEG foundation models. The task-specific supervised baselines represent the conventional one-dataset-one-model paradigm, in which a model is trained from scratch for each labelled downstream dataset. The EEG foundation model baselines represent the large-scale pre-training paradigm, in which a model first learns representations from unlabeled EEG recordings and is then adapted to downstream tasks. All baselines are evaluated under the same dataset splits and downstream evaluation protocols.

\textbf{EEGNet} \cite{lawhern2018eegnet} is a compact convolutional neural network designed for EEG-based brain-computer interface decoding. The model first applies a temporal convolution to learn frequency-sensitive filters from raw EEG waveforms, and then uses a depthwise convolution across electrodes to learn frequency-specific spatial filters for each temporal feature map. A subsequent separable convolution further summarizes temporal information within each feature map and uses pointwise convolution to mix feature maps efficiently. This factorized design substantially reduces the number of trainable parameters while retaining the ability to model temporal, spectral and spatial EEG structure. In this study, EEGNet represents a lightweight task-specific supervised encoder trained from scratch on each downstream dataset. The implementation is available at \url{https://github.com/vlawhern/arl-eegmodels}.

\textbf{ST-Transformer} \cite{song2021transformer} is an attention-based supervised EEG decoder that learns spatial and temporal dependencies from labelled EEG trials. It first applies common spatial pattern-inspired spatial filtering and feature-channel attention to enhance discriminative spatial components, and then slices the signal along the temporal dimension for multi-head self-attention. A convolution-based positional encoding layer preserves local temporal order before temporal attention, and the final representation is classified after global average pooling. In this study, ST-Transformer serves as a task-specific supervised encoder trained from scratch on each downstream dataset. The implementation is available at \url{https://github.com/anranknight/EEG-Transformer}.

\textbf{EEGConformer} \cite{song2022eeg} is a hybrid convolutional Transformer model for supervised EEG decoding. It first uses temporal and spatial convolutional layers to extract low-level local EEG features from raw trials, and then rearranges the pooled convolutional feature maps into temporal tokens. A multi-head self-attention module further models global temporal dependencies among these tokens, complementing the limited receptive field of convolutional operations. The resulting representation is passed to a compact fully connected classifier for downstream prediction. In this study, EEGConformer serves as a task-specific supervised encoder trained from scratch on each downstream dataset. The implementation is available at \url{https://github.com/eeyhsong/EEG-Conformer}.

\textbf{SPaRCNet} \cite{jing2023development} is a temporal convolutional neural network based on the DenseNet architecture. It stacks dense convolutional blocks to extract hierarchical temporal features from multi-channel EEG segments, with dense connections allowing later layers to reuse features learned by earlier layers. This design improves gradient flow and supports efficient feature propagation across network depth. In this study, SPaRCNet serves as a convolutional task-specific supervised encoder trained from scratch on each downstream dataset. The implementation is available at \url{https://github.com/bdsp-core/IIIC-SPaRCNet}.

\textbf{CBraMod} \cite{wang2025cbramod} is a reconstruction-centered EEG foundation model that learns reusable representations by recovering masked EEG patches from their surrounding context. It segments multi-channel EEG signals into temporal patches and encodes each patch with a time-frequency patch encoder that combines convolutional time-domain features with fast Fourier transform-based frequency-domain features. CBraMod uses asymmetric conditional positional encoding, implemented as a depthwise two-dimensional convolution, to dynamically encode spatial and temporal positional information from neighboring patches. Its criss-cross Transformer backbone separately models spatial dependencies across channels and temporal dependencies within each channel through parallel attention heads. During pre-training, the model optimizes a patch-based masked EEG reconstruction objective, and downstream adaptation is performed by replacing the reconstruction head with a task-specific prediction head. The implementation is available at \url{https://github.com/wjq-learning/CBraMod}.

\textbf{CSBrain} \cite{NEURIPS2025_7e199ad8} is a reconstruction-centered EEG foundation model that uses masked autoencoding to learn generalizable representations from unlabeled EEG signals. It is designed around the cross-scale spatiotemporal structure of EEG, where neural patterns may appear at different temporal windows and anatomical brain regions. The model introduces cross-scale spatiotemporal tokenization to aggregate multi-resolution features within local time windows and brain regions into compact scale-aware tokens. These tokens are then processed by structured sparse attention, which models long-range dependencies across temporal windows and brain regions while reducing redundant dense interactions. During pre-training, CSBrain reconstructs masked EEG segments, and downstream adaptation is performed by attaching a lightweight task-specific head. The implementation is available at \url{https://github.com/yuchen2199/CSBrain}.

\textbf{CodeBrain} \cite{ma2026codebrain} is a reconstruction-centered EEG foundation model built on a two-stage masked modeling framework. In the first stage, it trains a tokenizer that decouples temporal and frequency-domain EEG components into separate discrete codebooks, enabling domain-specific token representations for raw waveform patterns and spectral rhythms. In the second stage, CodeBrain uses an EEG structured state space model to predict the discrete token indices of masked EEG patches. This backbone combines structured global convolution for sparse long-range dependencies with sliding-window attention for local intra-patch dependencies, and uses dynamic positional embeddings to adapt to different channel configurations. This design emphasizes interpretable time-frequency tokenization and multi-scale dependency modeling within a reconstruction-centered pre-training paradigm. The implementation is available at \url{https://github.com/jingyingma01/CodeBrain}.

\subsection{Adaptation protocols}
We evaluate EEG foundation models under two downstream adaptation protocols: linear probing and full fine-tuning. These protocols assess complementary properties of a pre-trained EEG encoder. Linear probing tests whether downstream-relevant information is already accessible from a frozen representation space, whereas full fine-tuning tests whether the pre-trained backbone provides an effective initialization that can be further adapted by labelled downstream data.

For linear probing, the pre-trained encoder is kept frozen throughout downstream training. The backbone is used only for feature extraction and no gradients are propagated through its parameters. For most downstream datasets, only a lightweight linear classifier is optimized on top of the frozen representations. Thus, task-specific learning is restricted to the readout layer, providing a direct assessment of the separability of the learned EEG representations. For the ISRUC-S1 \cite{khalighi2016isruc} sleep-staging dataset, which is evaluated as a sequence prediction task, consecutive EEG epochs are first encoded by the frozen backbone and then passed to a lightweight sequence head for epoch-wise sleep-stage prediction. This design follows the temporal structure of sleep staging while keeping the pre-trained EEG encoder fixed.

For full fine-tuning, the pre-trained encoder and the downstream classification head are optimized jointly. For most datasets, the classification head maps the backbone representations to task labels using fully connected layers with nonlinear activation and dropout, followed by the final output layer. The backbone and the classification head are trained with separate learning rates, with a smaller learning rate for the backbone and a larger learning rate for the newly initialized task head. For ISRUC-S1, the same sequence-based formulation is used as in linear probing, but the backbone is also updated together with the sequence head and classifier.

All downstream models are trained using AdamW optimization, cross-entropy loss with label smoothing and gradient clipping. A cosine learning-rate schedule is used during training. Model selection is performed on the validation set, and the checkpoint with the best validation performance is used for final evaluation on the held-out test set. All experiments are repeated with five random seeds, and results are reported as mean $\pm$ standard deviation.

\subsection{Evaluation metrics}
We use evaluation metrics according to the label structure of each downstream task. For binary classification datasets, we report balanced accuracy, area under the receiver operating characteristic curve (AUROC) and area under the precision-recall curve (AUC-PR). For multi-class classification datasets, we report balanced accuracy, weighted F1 score and Cohen's kappa coefficient. These metrics provide complementary assessments of discriminative performance, class-imbalance robustness and agreement beyond chance.

\textbf{Balanced accuracy.}
Balanced accuracy measures the average class-wise recall and is therefore less dominated by majority classes than standard accuracy. For a classification problem with $C$ classes, it is defined as
\begin{equation}
\mathrm{BalAcc}
=
\frac{1}{C}
\sum_{i=1}^{C}
\frac{\mathrm{TP}_{i}}
{\mathrm{TP}_{i}+\mathrm{FN}_{i}},
\end{equation}
where $\mathrm{TP}_{i}$ and $\mathrm{FN}_{i}$ denote the numbers of true positives and false negatives for class $i$, respectively. In binary classification, the same definition corresponds to the mean of sensitivity and specificity.

\textbf{Weighted F1 score.}
The weighted F1 score summarizes class-wise F1 scores after weighting each class by its empirical support. Let $n_i$ be the number of test samples belonging to class $i$, and let $N=\sum_{i=1}^{C} n_i$ be the total number of test samples. The class weight is $w_i=n_i/N$, and the weighted F1 score is
\begin{equation}
\mathrm{F1}_{\mathrm{weighted}}
=
\sum_{i=1}^{C}
w_i
\frac{
2\,\mathrm{Precision}_{i}\,\mathrm{Recall}_{i}
}{
\mathrm{Precision}_{i}+\mathrm{Recall}_{i}
},
\end{equation}
where
\begin{equation}
\mathrm{Precision}_{i}
=
\frac{\mathrm{TP}_{i}}
{\mathrm{TP}_{i}+\mathrm{FP}_{i}},
\qquad
\mathrm{Recall}_{i}
=
\frac{\mathrm{TP}_{i}}
{\mathrm{TP}_{i}+\mathrm{FN}_{i}}.
\end{equation}
Here, $\mathrm{FP}_{i}$ denotes the number of false positives for class $i$. This metric accounts for both precision and recall while preserving the contribution of each class according to its frequency in the test set.

\textbf{Cohen's kappa.}
Cohen's kappa coefficient measures the agreement between predicted and ground-truth labels after correcting for the agreement expected by chance. It is defined as
\begin{equation}
\kappa
=
\frac{p_o-p_e}
{1-p_e},
\end{equation}
where $p_o$ denotes the observed agreement and $p_e$ denotes the expected agreement under independent marginal label distributions. Given the confusion matrix $\mathbf{M}\in\mathbb{R}^{C\times C}$, with $M_{ij}$ denoting the number of samples whose true class is $i$ and predicted class is $j$, these two quantities are computed as
\begin{equation}
p_o
=
\frac{1}{N}
\sum_{i=1}^{C} M_{ii},
\qquad
p_e
=
\sum_{i=1}^{C}
\left(
\frac{\sum_{j=1}^{C} M_{ij}}{N}
\right)
\left(
\frac{\sum_{j=1}^{C} M_{ji}}{N}
\right).
\end{equation}
A kappa value of 1 indicates perfect agreement, whereas a value close to 0 indicates agreement close to chance.

\textbf{AUROC.}
AUROC evaluates the ranking quality of the predicted scores for binary classification. Let $s(x)$ be the predicted score for the positive class. For a threshold $\tau$, the true positive rate and false positive rate are
\begin{equation}
\mathrm{TPR}(\tau)
=
\frac{\mathrm{TP}(\tau)}
{\mathrm{TP}(\tau)+\mathrm{FN}(\tau)},
\qquad
\mathrm{FPR}(\tau)
=
\frac{\mathrm{FP}(\tau)}
{\mathrm{FP}(\tau)+\mathrm{TN}(\tau)}.
\end{equation}
AUROC is the area under the receiver operating characteristic curve obtained by varying $\tau$:
\begin{equation}
\mathrm{AUROC}
=
\int_{0}^{1}
\mathrm{TPR}(u)\,du,
\qquad
u=\mathrm{FPR}.
\end{equation}
Equivalently, AUROC can be interpreted as the probability that a randomly selected positive sample receives a higher score than a randomly selected negative sample, with ties counted with half weight:
\begin{equation}
\mathrm{AUROC}
=
\Pr\!\left(s^{+}>s^{-}\right)
+
\frac{1}{2}\Pr\!\left(s^{+}=s^{-}\right),
\end{equation}
where $s^{+}$ and $s^{-}$ denote scores from positive and negative samples, respectively.

\textbf{AUC-PR.}
AUC-PR evaluates the trade-off between precision and recall across decision thresholds and is particularly informative when the positive class is rare. For a threshold $\tau$, precision and recall are defined as
\begin{equation}
\mathrm{Precision}(\tau)
=
\frac{\mathrm{TP}(\tau)}
{\mathrm{TP}(\tau)+\mathrm{FP}(\tau)},
\qquad
\mathrm{Recall}(\tau)
=
\frac{\mathrm{TP}(\tau)}
{\mathrm{TP}(\tau)+\mathrm{FN}(\tau)}.
\end{equation}
The AUC-PR is the area under the precision-recall curve:
\begin{equation}
\mathrm{AUC\mbox{-}PR}
=
\int_{0}^{1}
\mathrm{Precision}(r)\,dr,
\qquad
r=\mathrm{Recall}.
\end{equation}
Compared with AUROC, AUC-PR places greater emphasis on the model's ability to identify positive samples with a low false-positive burden.

\section{Supplementary results}
\label{sec:supp_results}

The main text organizes downstream results by application level, including signal quality assessment, brain-state decoding, and brain-health evaluation. Here, we provide a complementary view organized by the relationship between each downstream montage and the 19-channel 10-20 configuration used during pre-training. This supplementary analysis reports the complete metric set for each dataset and separates montage-matched datasets from montage-shifted datasets, providing another perspective on how INCEPT transfers across electrode layouts, channel densities and recording contexts. Each dataset is summarized by three complementary metrics. For binary classification tasks, we report balanced accuracy, AUROC and AUC-PR. For multi-class classification tasks, we report balanced accuracy, Cohen's kappa and weighted F1 score.

\begin{figure*}[!bp]
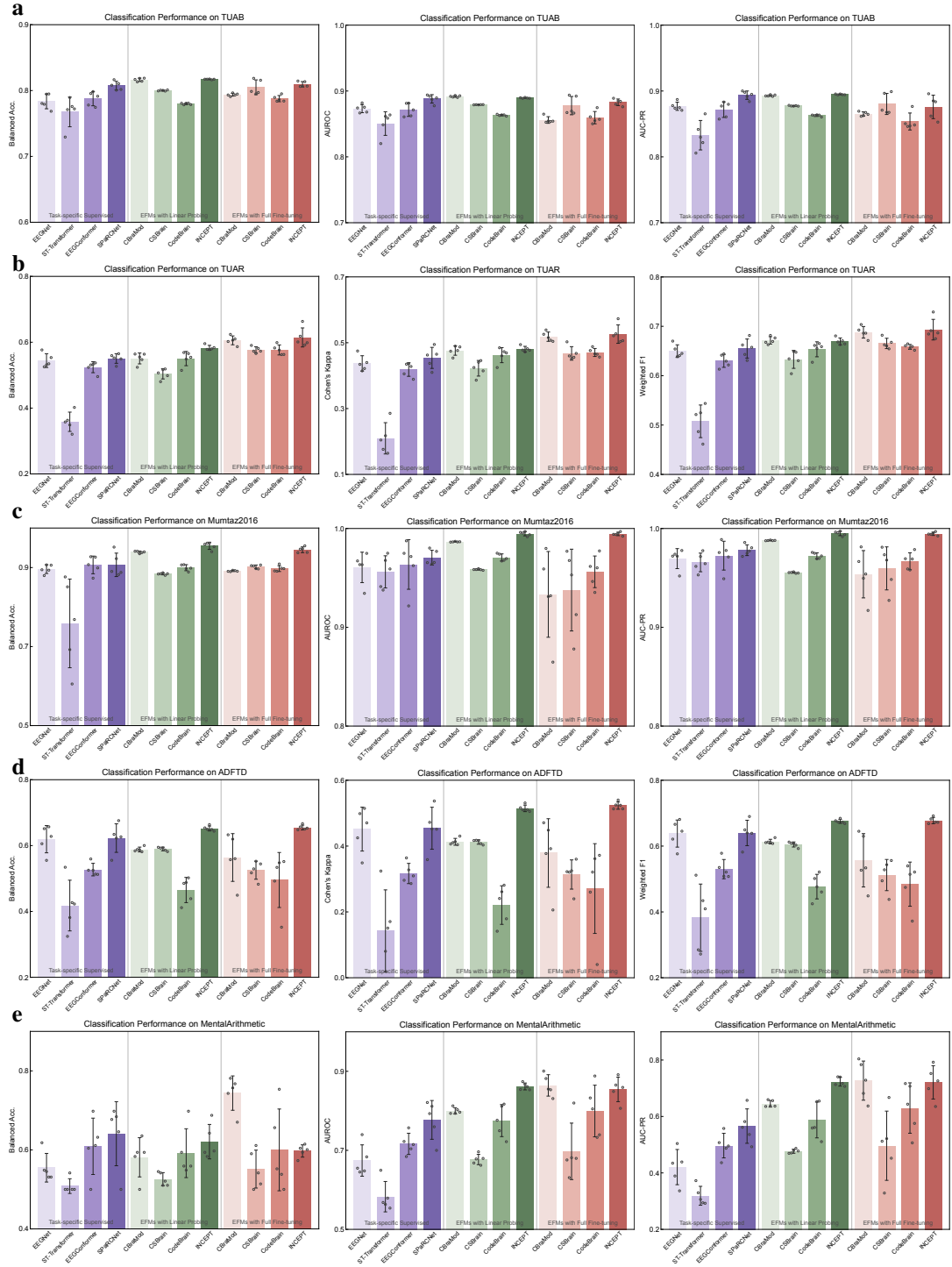

    \centering
    \setlength{\tabcolsep}{-1pt}
    \renewcommand{\arraystretch}{1}

    \begin{tabular}{@{}ccc@{}}

    \imgpanellabel{0.335\textwidth}{a}{figures/TUAB/TUAB_Bal_Acc.pdf}
    &
    \includegraphics[width=0.335\textwidth]{figures/TUAB/TUAB_AUROC.pdf}
    &
    \includegraphics[width=0.335\textwidth]{figures/TUAB/TUAB_AUC_PR.pdf}
    \\[-0.5em]

    \imgpanellabel{0.335\textwidth}{b}{figures/TUAR/TUAR_Bal_Acc.pdf}
    &
    \includegraphics[width=0.335\textwidth]{figures/TUAR/TUAR_Kappa.pdf}
    &
    \includegraphics[width=0.335\textwidth]{figures/TUAR/TUAR_F1_W.pdf}
    \\[-0.5em]

    \imgpanellabel{0.335\textwidth}{c}{figures/Mumtaz2016/Mumtaz2016_Bal_Acc.pdf}
    &
    \includegraphics[width=0.335\textwidth]{figures/Mumtaz2016/Mumtaz2016_AUROC.pdf}
    &
    \includegraphics[width=0.335\textwidth]{figures/Mumtaz2016/Mumtaz2016_AUC_PR.pdf}
    \\[-0.5em]

    \imgpanellabel{0.335\textwidth}{d}{figures/ADFTD/ADFTD_Bal_Acc.pdf}
    &
    \includegraphics[width=0.335\textwidth]{figures/ADFTD/ADFTD_Kappa.pdf}
    &
    \includegraphics[width=0.335\textwidth]{figures/ADFTD/ADFTD_F1_W.pdf}
    \\[-0.5em]

    \imgpanellabel{0.335\textwidth}{e}{figures/MentalArithmetic/MentalArithmetic_Bal_Acc.pdf}
    &
    \includegraphics[width=0.335\textwidth]{figures/MentalArithmetic/MentalArithmetic_AUROC.pdf}
    &
    \includegraphics[width=0.335\textwidth]{figures/MentalArithmetic/MentalArithmetic_AUC_PR.pdf}

    \end{tabular}

    \caption{
        \textbf{Complete downstream performance on montage-matched datasets.}
        \textbf{a}, TUAB.
        \textbf{b}, TUAR.
        \textbf{c}, Mumtaz2016.
        \textbf{d}, ADFTD.
        \textbf{e}, MentalArithmetic.
        Each row corresponds to one dataset and each column corresponds to one evaluation metric. For TUAB, Mumtaz2016 and MentalArithmetic, the metrics are balanced accuracy, AUROC and AUC-PR. For TUAR and ADFTD, the metrics are balanced accuracy, Cohen's kappa and weighted F1 score. Bars indicate mean~$\pm$~s.d. over five random seeds, with hollow circles denoting individual seeds. Methods are grouped as task-specific supervised encoders, EEG foundation models with linear probing and EEG foundation models with full fine-tuning.
    }
    \label{fig:supp_multimetric_matched}
\end{figure*}

\begin{figure*}[!htbp]
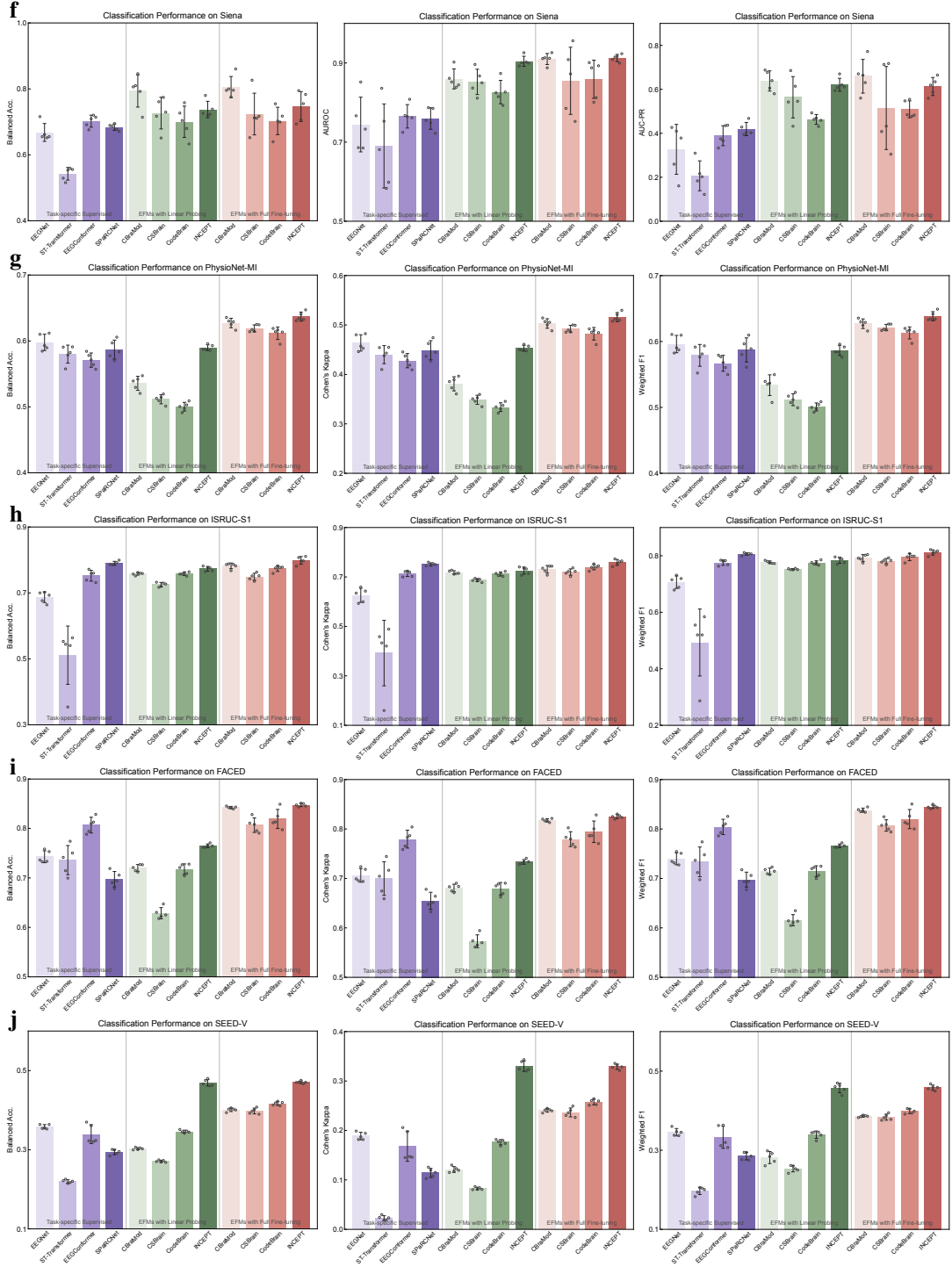

    \centering
    \setlength{\tabcolsep}{-1pt}
    \renewcommand{\arraystretch}{1}

    \begin{tabular}{@{}ccc@{}}

    \imgpanellabel{0.335\textwidth}{f}{figures/Siena/Siena_Bal_Acc.pdf}
    &
    \includegraphics[width=0.335\textwidth]{figures/Siena/Siena_AUROC.pdf}
    &
    \includegraphics[width=0.335\textwidth]{figures/Siena/Siena_AUC_PR.pdf}
    \\[-0.5em]

    \imgpanellabel{0.335\textwidth}{g}{figures/PhysioNet-MI/PhysioNet-MI_Bal_Acc.pdf}
    &
    \includegraphics[width=0.335\textwidth]{figures/PhysioNet-MI/PhysioNet-MI_Kappa.pdf}
    &
    \includegraphics[width=0.335\textwidth]{figures/PhysioNet-MI/PhysioNet-MI_F1_W.pdf}
    \\[-0.5em]

    \imgpanellabel{0.335\textwidth}{h}{figures/ISRUC-S1/ISRUC-S1_Bal_Acc.pdf}
    &
    \includegraphics[width=0.335\textwidth]{figures/ISRUC-S1/ISRUC-S1_Kappa.pdf}
    &
    \includegraphics[width=0.335\textwidth]{figures/ISRUC-S1/ISRUC-S1_F1_W.pdf}
    \\[-0.5em]

    \imgpanellabel{0.335\textwidth}{i}{figures/FACED/FACED_Bal_Acc.pdf}
    &
    \includegraphics[width=0.335\textwidth]{figures/FACED/FACED_Kappa.pdf}
    &
    \includegraphics[width=0.335\textwidth]{figures/FACED/FACED_F1_W.pdf}
    \\[-0.5em]

    \imgpanellabel{0.335\textwidth}{j}{figures/SEED-V/SEEDV_Bal_Acc.pdf}
    &
    \includegraphics[width=0.335\textwidth]{figures/SEED-V/SEEDV_Kappa.pdf}
    &
    \includegraphics[width=0.335\textwidth]{figures/SEED-V/SEEDV_F1_W.pdf}

    \end{tabular}

    \caption{
        \textbf{Complete downstream performance on montage-shifted datasets.}
        \textbf{f}, Siena.
        \textbf{g}, PhysioNet-MI.
        \textbf{h}, ISRUC-S1.
        \textbf{i}, FACED.
        \textbf{j}, SEED-V.
        Each row corresponds to one dataset and each column corresponds to one evaluation metric. For Siena, the metrics are balanced accuracy, AUROC and AUC-PR. For PhysioNet-MI, ISRUC-S1, FACED and SEED-V, the metrics are balanced accuracy, Cohen's kappa and weighted F1 score. Bars indicate mean~$\pm$~s.d. over five random seeds, with hollow circles denoting individual seeds. Methods are grouped as task-specific supervised encoders, EEG foundation models with linear probing and EEG foundation models with full fine-tuning.
    }
    \label{fig:supp_multimetric_shifted}
\end{figure*}

\subsection{Results on montage-matched downstream datasets}
\label{subsec:supp_multimetric_montage_matched}
Fig.~\ref{fig:supp_multimetric_matched} shows the complete metric profiles for TUAB \cite{obeid2016temple}, TUAR \cite{obeid2016temple}, Mumtaz2016 \cite{mumtaz2016mdd,mumtaz2018machine}, ADFTD \cite{ds004504:1.0.8} and MentalArithmetic \cite{zyma2019electroencephalograms}, whose electrode layouts are close to the 19-channel 10-20 configuration used during pre-training. Across these montage-matched datasets, task-specific supervised encoders remain competitive in several settings, particularly on TUAB and Mumtaz2016, indicating that dedicated supervised models can still perform well when the downstream dataset is sufficiently aligned with their architectural inductive biases. However, their relative ranking varies substantially across datasets and metrics.

EEG foundation models show more stable behaviour under linear probing, suggesting that pre-training provides representations that can be read out without updating the backbone. This trend is most evident on TUAB, TUAR, Mumtaz2016 and ADFTD, where INCEPT achieves strong performance with a frozen encoder and remains competitive across all metrics. On Mumtaz2016, INCEPT reaches near-ceiling performance across the three binary-classification metrics under both linear probing and full fine-tuning. On ADFTD, INCEPT shows a clearer advantage across all three multi-class metrics, and full fine-tuning further improves the separation of diagnostic classes. MentalArithmetic shows a slightly different pattern: although INCEPT is marginally lower than CBraMod under full fine-tuning, it still outperforms all other models and shows a marked advantage under linear probing. These results indicate that, on montage-matched datasets, INCEPT learns representations that are already strongly readable before backbone adaptation, while still allowing further task-specific refinement when needed.

\subsection{Results on montage-shifted downstream datasets}
\label{subsec:supp_multimetric_montage_shifted}
Fig.~\ref{fig:supp_multimetric_shifted} shows the complete metric profiles for Siena, PhysioNet-MI, ISRUC-S1, FACED and SEED-V, which differ from the pre-training setting in electrode layout, channel density, segment duration or task context. Compared with the montage-matched datasets, these settings show a clearer distinction between frozen readout and supervised backbone adaptation. Linear probing remains informative, but full fine-tuning generally provides stronger performance, especially on PhysioNet-MI, ISRUC-S1, FACED and SEED-V. This pattern indicates that montage-shifted downstream tasks benefit from allowing the pre-trained backbone to adapt to new electrode configurations and task-specific neural patterns.

Across the multi-class montage-shifted datasets, INCEPT shows the most consistent benefits after full fine-tuning. On PhysioNet-MI, full fine-tuning improves INCEPT across balanced accuracy, Cohen's kappa and weighted F1 score, indicating that the pre-trained representation can be effectively adapted to the 64-channel motor imagery setting. On ISRUC-S1, INCEPT achieves the strongest overall performance across the three sleep-staging metrics despite the sparse 6-channel montage. On FACED and SEED-V, INCEPT also performs strongly across all three metrics, with particularly clear gains on SEED-V, suggesting that its representations remain adaptable to affective EEG decoding under both low-density and high-density electrode configurations. Siena shows a slightly different pattern: INCEPT remains competitive, especially on AUROC, whereas balanced accuracy and AUC-PR vary more across methods. Nevertheless, when considered together with PhysioNet-MI, ISRUC-S1, FACED and SEED-V, the montage-shifted results indicate that INCEPT is not only a frozen feature extractor, but also a transferable initialization that can be reshaped by downstream supervision when the target montage and task distribution depart from the pre-training corpus.

\section{Detailed Analysis of Downstream Results}
\label{supplement:detailed_downstream}

\subsection{Signal-level EEG assessment}

\subsubsection{(i) Clinical abnormality assessment from routine EEG}
On TUAB, the detailed comparisons show that foundation pre-training improves the reproducibility of abnormality assessment. Taking balanced accuracy as an example, EEG foundation models under linear probing reduce the average seed-to-seed standard deviation by 88.0\% relative to task-specific supervised encoders. Within this more stable foundation-model regime, INCEPT improves balanced accuracy by 3.9\% relative to the average task-specific supervised encoder and by 2.3\% relative to the average of the other linear-probing EEG foundation models. It also reduces the standard deviation of balanced accuracy by 96.5\% relative to the average task-specific supervised encoder and by 70.7\% relative to the average of the other linear-probing EEG foundation models. Because TUAB is derived from TUEG, the clinical EEG resource used for pre-training by all evaluated EEG foundation models, these comparisons suggest that INCEPT makes more effective use of the same unlabeled clinical EEG corpus. Under full fine-tuning, INCEPT remains stronger than the other fine-tuned EEG foundation models, improving AUROC by 2.1\% on average, although linear probing remains the more favourable protocol for both INCEPT and CBraMod in this montage-matched abnormality task.

\subsubsection{(ii) Artifact-type recognition for EEG quality control}
On TUAR, INCEPT also provides strong frozen readability for artifact recognition. Under linear probing, it achieves the strongest balanced accuracy and Cohen's kappa among EEG foundation models, with weighted F1 remaining comparable to the best foundation-model result. Relative to the average task-specific supervised encoder, INCEPT improves balanced accuracy by 17.9\%, weighted F1 by 9.6\% and kappa by 26.5\%. Compared with the average of the other linear-probing EEG foundation models, it improves balanced accuracy by 9.0\%, weighted F1 by 2.5\% and kappa by 5.9\%. It also reduces the standard deviation of balanced accuracy, weighted F1 and kappa by 63.5\%, 64.0\% and 72.7\%, respectively, relative to the average task-specific supervised encoder, and by 58.5\%, 48.3\% and 58.3\% relative to the average of the other linear-probing EEG foundation models.

Full fine-tuning further improves INCEPT on TUAR. Relative to its own linear-probing result, fine-tuned INCEPT improves balanced accuracy by 5.4\%, weighted F1 by 3.6\% and kappa by 9.5\%. It achieves the strongest fine-tuned performance among all evaluated models, improving balanced accuracy, weighted F1 and kappa by 24.3\%, 13.5\% and 38.5\% relative to the average task-specific supervised encoder, and by 4.8\%, 3.4\% and 8.4\% relative to the average of the other fine-tuned EEG foundation models. The gain in mean performance is accompanied by a larger seed-to-seed standard deviation than under linear probing, indicating that backbone adaptation can improve artifact discrimination while introducing additional optimization variability.

\subsection{Brain-state decoding}

\subsubsection{(i) Affective-state decoding across low- and high-density emotion EEG}
FACED and SEED-V highlight complementary affective-transfer regimes. On FACED, task-specific supervised encoders remain strong, with EEGConformer performing competitively. However, foundation pre-training provides a more reproducible readout: under linear probing, EEG foundation models reduce the average standard deviation of weighted F1 by 55.8\% relative to task-specific supervised encoders. Among foundation models, INCEPT improves weighted F1 by 16.8\% relative to the average of the other EEG foundation models and reduces the standard deviation of weighted F1 by 61.2\%. After full fine-tuning, INCEPT improves weighted F1 by 13.6\% relative to the average task-specific supervised encoder and by 2.8\% relative to the average of the other EEG foundation models, indicating that the same representation remains adaptable to the FACED label space.

SEED-V is more difficult, with shorter segments, higher channel density and lower overall agreement across models. In this dataset, INCEPT's advantage is more direct. Under linear probing, INCEPT improves weighted F1 by 57.6\% relative to the average task-specific supervised encoder and by 56.9\% relative to the average of the other EEG foundation models. Full fine-tuning preserves this advantage, with INCEPT improving weighted F1 by 57.9\% relative to the average task-specific supervised encoder. Compared with the average of the other fine-tuned EEG foundation models, INCEPT further improves weighted F1 by 17.6\% and kappa by 34.4\%.

\subsubsection{(ii) Sensorimotor-state decoding from high-density motor-imagery EEG}
On PhysioNet-MI, task-specific supervised encoders remain competitive because they learn spatial and temporal filters directly from labelled BCI trials. Existing EEG foundation models show a substantial drop under linear probing, with only INCEPT reaching performance comparable to task-specific supervised encoders. Under linear probing, INCEPT improves weighted F1 by 13.8\% and kappa by 27.8\% relative to the average of the other EEG foundation models, while reducing the standard deviation of weighted F1 by 57.6\%. After full fine-tuning, INCEPT improves weighted F1 by 7.1\% relative to the strongest task-specific supervised encoder and by 9.6\% relative to the average task-specific supervised encoder. Compared with the average of the other fine-tuned EEG foundation models, it further improves balanced accuracy by 2.9\%, weighted F1 by 2.9\% and kappa by 4.8\%.

\subsubsection{(iii) Sleep-state staging from sparse clinical sleep montages}
On ISRUC-S1, sparse-montage sleep staging benefits from pre-training. Task-specific supervised encoders show strong performance when their architectural bias matches sleep-stage structure, with SPaRCNet providing a competitive supervised baseline. Under linear probing, EEG foundation models improve weighted F1 by 11.1\% on average relative to task-specific supervised encoders and reduce the average seed-to-seed standard deviation of weighted F1 by 83.6\%. Within the foundation-model comparison, INCEPT improves balanced accuracy by 3.5\%, weighted F1 by 2.1\% and kappa by 2.8\% relative to the average of the other EEG foundation models under linear probing. Full fine-tuning further strengthens this advantage, with INCEPT improving balanced accuracy by 3.9\%, weighted F1 by 2.8\% and kappa by 3.9\% relative to the average of the other fine-tuned EEG foundation models. Compared with the average task-specific supervised encoder, full fine-tuned INCEPT improves weighted F1 by 16.8\% and kappa by 22.3\%.

\subsection{Brain-health evaluation}

\subsubsection{(i) Depression- and stress-related EEG assessment}
Mumtaz2016 shows a more stable montage-matched brain-health pattern. Task-specific supervised encoders remain competitive, especially SPaRCNet and EEGConformer, but EEG foundation models as a group provide a stronger and more reproducible representation space. Under linear probing, the average EEG foundation model improves balanced accuracy by 6.0\%, AUROC by 1.6\% and AUC-PR by 0.6\% relative to the average task-specific supervised encoder, while reducing the average seed-to-seed standard deviation by 87.6\%, 89.8\% and 85.1\% for the three metrics. INCEPT further improves this foundation-model baseline: under linear probing, it improves balanced accuracy by 5.3\% relative to the strongest task-specific supervised encoder and by 10.1\% relative to the supervised-model average. Compared with the average of the other EEG foundation models, it improves balanced accuracy by 5.2\%, AUROC by 2.3\% and AUC-PR by 2.4\%. Full fine-tuning again provides limited additional benefit. CBraMod, CodeBrain and INCEPT all decrease relative to their own linear-probing results across the three metrics, whereas CSBrain shows only a modest increase in balanced accuracy and AUC-PR. Even after this decrease, INCEPT still remains the strongest fine-tuned model on Mumtaz2016, improving balanced accuracy by 4.2\% relative to the strongest supervised encoder and by 9.0\% relative to the supervised-model average. It also improves balanced accuracy by 5.4\%, AUROC by 5.5\% and AUC-PR by 3.6\% relative to the average of the other fine-tuned EEG foundation models.

MentalArithmetic exposes a distinction between score-level separability and final class-decision performance. Task-specific supervised encoders vary substantially, with SPaRCNet providing the strongest balanced accuracy among task-specific baselines. Under linear probing, CBraMod achieves higher AUROC and AUC-PR than EEGConformer and SPaRCNet, but its balanced accuracy remains lower than both supervised models. INCEPT shows the clearest frozen transfer among foundation models: under linear probing, it improves balanced accuracy by 9.6\%, AUROC by 14.8\% and AUC-PR by 27.0\% relative to the average of the other EEG foundation models. It also improves balanced accuracy by 7.4\%, AUROC by 25.3\% and AUC-PR by 60.5\% relative to the average task-specific supervised encoder. Full fine-tuning does not further improve INCEPT on this montage-matched stress task, with balanced accuracy, AUROC and AUC-PR decreasing by 3.8\%, 0.9\% and 0.3\% relative to its own linear-probing result. Even so, the fine-tuned model retains strong ranking performance, improving AUROC by 9.9\% and AUC-PR by 27.4\% relative to the strongest task-specific supervised encoder, and improving AUROC by 8.5\% and AUC-PR by 16.7\% relative to the average of the other fine-tuned EEG foundation models.

\subsubsection{(ii) Neurodegenerative disease evaluation from resting-state EEG}
On ADFTD, task-specific supervised encoders remain useful, especially SPaRCNet and EEGNet, but their performance varies substantially across architectures. INCEPT provides stronger and more stable transfer than this task-specific strategy. Under linear probing, it improves balanced accuracy by 19.3\%, weighted F1 by 23.3\% and kappa by 50.8\% relative to the average task-specific supervised encoder. Compared with the strongest task-specific supervised encoder, INCEPT still improves balanced accuracy by 4.7\%, weighted F1 by 5.5\% and kappa by 13.3\%. It also reduces seed-to-seed standard deviation by 83.4\%, 88.6\% and 86.1\% for balanced accuracy, weighted F1 and kappa relative to the average task-specific supervised encoder. Within the foundation-model comparison, INCEPT improves balanced accuracy by 19.1\%, weighted F1 by 19.5\% and kappa by 47.6\% relative to the average of the other EEG foundation models under linear probing, and by 10.6\%, 10.0\% and 24.6\% relative to the strongest existing EEG foundation model under the same protocol.

Full fine-tuning on ADFTD reveals the stability difference among foundation models. CBraMod decreases after full fine-tuning, CSBrain decreases more substantially, CodeBrain improves only modestly, and INCEPT shows a small but consistent increase. The three reconstruction-centered EEG foundation models show markedly larger standard deviations after full fine-tuning, whereas INCEPT remains stable, with standard deviations below 1.0 across all three metrics. Relative to the average of the other fine-tuned EEG foundation models, INCEPT reduces standard deviation by 87.9\%, 85.2\% and 88.5\% for balanced accuracy, weighted F1 and kappa. After full fine-tuning, INCEPT remains the strongest model on ADFTD, improving balanced accuracy by 24.0\%, weighted F1 by 30.9\% and kappa by 62.8\% relative to the average of the other fine-tuned EEG foundation models, and by 5.2\%, 5.9\% and 15.1\% relative to the strongest task-specific supervised encoder.

\subsubsection{(iii) Seizure-related neurological assessment under extended clinical montages}
Siena shows that seizure-related assessment under an extended clinical montage is metric-specific. Under linear probing, the average of the four EEG foundation models improves balanced accuracy by 13.9\%, AUROC by 16.3\% and AUC-PR by 70.3\% relative to the average task-specific supervised encoder. Full fine-tuning preserves this advantage, with the average EEG foundation model improving balanced accuracy by 14.7\%, AUROC by 19.5\% and AUC-PR by 71.1\%. The advantage is especially clear for AUROC and AUC-PR, suggesting that pre-training provides stronger score-level separability for seizure and non-seizure segments under label imbalance. CodeBrain has the lowest balanced accuracy among EEG foundation models under both adaptation protocols, but it still improves AUROC and AUC-PR relative to the strongest task-specific supervised encoder by 8.0\% and 10.5\% under linear probing, and by 12.2\% and 21.4\% after full fine-tuning.

Within the foundation-model comparison on Siena, INCEPT and CBraMod show complementary strengths. CBraMod achieves higher balanced accuracy and AUC-PR, whereas INCEPT achieves the strongest AUROC under both linear probing and full fine-tuning. Under linear probing, INCEPT improves AUROC by 6.9\% relative to the average of the other EEG foundation models and by 5.2\% relative to the strongest existing EEG foundation model. After full fine-tuning, it further improves AUROC by 4.3\% relative to the average of the other fine-tuned EEG foundation models. INCEPT also provides a more stable seizure-related ranking signal: compared with CBraMod, it reduces the standard deviation of balanced accuracy by 49.0\%, AUROC by 50.0\% and AUC-PR by 36.4\% under linear probing. After full fine-tuning, INCEPT still reduces the standard deviation of AUROC by 88.4\% and AUC-PR by 46.7\% relative to CBraMod, although its balanced-accuracy standard deviation is higher.

\section{Scaling trends}
\label{supplement:scaling_trends}

A defining property of foundation models is that representation quality should improve as the amount of pre-training data and model capacity increase. We therefore examine whether INCEPT shows scaling behaviour in EEG foundation pre-training by varying two factors: the amount of unlabelled EEG used for pre-training and the number of trainable model parameters. The resulting models are evaluated on two representative downstream datasets under the same full fine-tuning protocol: Mumtaz2016 \cite{mumtaz2016mdd,mumtaz2018machine}, which uses a standard 10-20 montage and is relatively close to the pre-training distribution, and FACED \cite{chen2023large}, which uses a low-density 10-10 montage and provides a more distribution-shifted evaluation setting.

\begin{figure*}[!t]
    \centering
    \setlength{\tabcolsep}{1.5pt}
    \renewcommand{\arraystretch}{1}

    \begin{minipage}[t]{0.68\textwidth}
        \vspace{0pt}
        \begin{tabular}{@{}cc@{}}
        \scalingpanellabel{width=0.49\linewidth}{a}
            {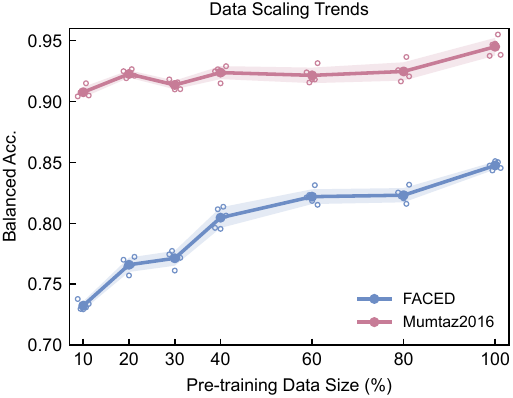}
        &
        \includegraphics[width=0.49\linewidth]
            {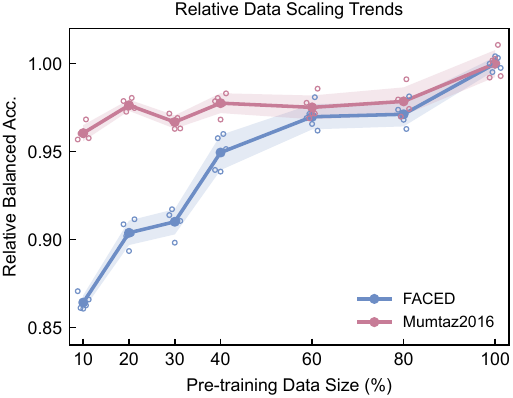}
        \\[-0.35em]
        \scalingpanellabel{width=0.49\linewidth}{b}
            {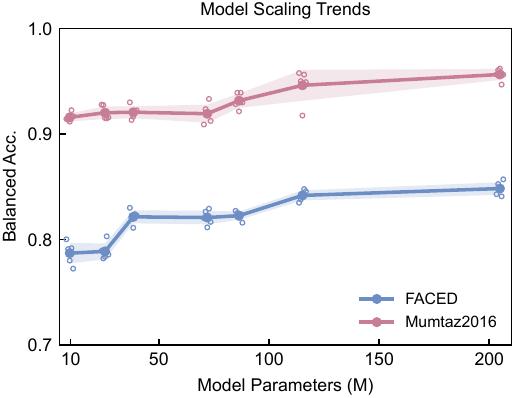}
        &
        \includegraphics[width=0.49\linewidth]
            {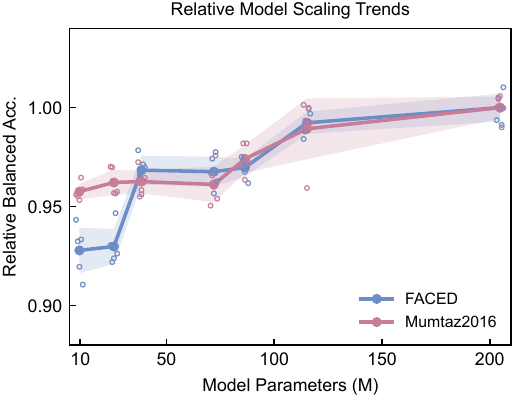}
        \\
        \end{tabular}
    \end{minipage}
    \hfill
    \begin{minipage}[t]{0.31\textwidth}
        \vspace{0pt}
        \scalingpanellabel{width=\linewidth}{c}
            {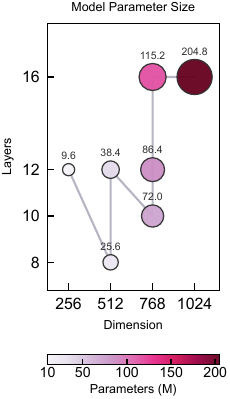}
    \end{minipage}
    
    \caption{
        \textbf{Data and model scaling behaviour of EEG foundation pre-training.}
        \textbf{a}, Effect of increasing pre-training data size on downstream balanced accuracy under the full fine-tuning protocol. Results are shown as absolute scores and normalized scores, with each downstream dataset normalized by the performance obtained using 100\% of the pre-training data. Lines denote the mean across five random seeds, shaded regions indicate mean~$\pm$~s.d., and hollow circles show individual seed results.
        \textbf{b}, Effect of increasing model capacity on downstream balanced accuracy under the full fine-tuning protocol, shown as absolute scores and normalized scores, with each downstream dataset normalized by the performance of the largest model.
        \textbf{c}, Parameter counts for the Transformer depth and embedding-dimension combinations used in the model-scaling analysis.
    }
    \label{fig:supp_scaling_trends}
\end{figure*}

\subsection{Increasing pre-training data size improves downstream transfer}
To assess data scaling, we construct a series of pre-training subsets from the cleaned EEG corpus. After preprocessing and artifact control, the corpus contains approximately 11,000 hours of usable recordings \cite{obeid2016temple}. We randomly sample 10\%, 20\%, 30\%, 40\%, 60\%, 80\% and 100\% of the pre-training data, while keeping the model architecture, optimization settings and downstream full fine-tuning protocol unchanged. The 100\% setting corresponds to the full INCEPT pre-training configuration. This experiment tests whether the proposed invariance-oriented pre-training strategy continues to benefit from additional unlabelled EEG data.

INCEPT shows a clear positive data-scaling trend (Supplementary Fig.~\ref{fig:supp_scaling_trends}a). As the amount of pre-training data increases, downstream performance generally improves, indicating that larger unlabelled EEG corpora provide more useful information for transferable representation learning. The effect is particularly evident on FACED \cite{chen2023large}. Because FACED differs from the pre-training corpus in montage configuration, electrode density and affective brain-state context, this improvement suggests that additional pre-training data help INCEPT learn representations that are more robust under distribution shift. The gains are most pronounced in the lower-data regime, where adding more unlabelled EEG rapidly improves downstream transfer. At larger data fractions, performance continues to increase but the marginal gains become smaller, suggesting partial saturation at the current model scale.

The scaling effect is more modest on Mumtaz2016 \cite{mumtaz2016mdd,mumtaz2018machine}. This difference is consistent with the closer match between Mumtaz2016 and the clinical 10-20 EEG recordings used for pre-training. Smaller pre-training subsets may already capture much of the acquisition geometry and resting-state structure needed for this downstream task. Task-specific factors, including cohort size, label noise and the intrinsic difficulty of depression-related EEG classification, may also limit the magnitude of observable gains. Overall, the data-scaling results indicate that increasing pre-training data is especially beneficial when downstream recordings differ more substantially from the pre-training distribution.

\subsection{Increasing model size improves downstream transfer}
We next assess model scaling by training INCEPT variants with different parameter sizes on the full pre-training corpus. Model capacity is varied by changing the input embedding dimension and the number of Transformer layers \cite{vaswani2017attention}, yielding models with parameter counts ranging from 9.61 million to 202.26 million (Supplementary Fig.~\ref{fig:supp_scaling_trends}c). All variants use the same pre-training objective and are evaluated on Mumtaz2016 \cite{mumtaz2016mdd,mumtaz2018machine} and FACED \cite{chen2023large} using the same full fine-tuning protocol. Because large-scale EEG foundation pre-training is computationally expensive, the model-scaling experiments use half the number of pre-training iterations used for the main INCEPT model. This design tests whether the proposed pre-training objective continues to benefit from increased model capacity.

Downstream performance generally improves as model size increases (Supplementary Fig.~\ref{fig:supp_scaling_trends}b). This trend appears on both Mumtaz2016 and FACED, indicating that larger INCEPT variants can better exploit the same unlabelled EEG corpus across different acquisition settings. Larger models may capture richer long-range temporal dependencies, cross-channel interactions and global segment-level structure, all of which are important for transferable EEG representation learning. The largest model with 202.26 million parameters outperforms the 85.54-million-parameter base model used in the main experiments, suggesting that the performance of the default INCEPT configuration is not the upper limit of the method. However, the model-scaling gains are moderate relative to the additional computational cost. Compared with the base model, the largest model improves performance by approximately 3\% on FACED and by less than 1\% on Mumtaz2016. This pattern suggests that scaling benefits depend on downstream distribution shift: when the downstream dataset is closer to the pre-training corpus, additional capacity yields smaller marginal improvement; when the downstream dataset differs more strongly from the pre-training distribution, larger models provide more useful representational flexibility.

\vspace{0.5em}
\noindent\textbf{\textit{Overall,}} the data and model scaling experiments show that INCEPT benefits from scale. Increasing the amount of unlabelled EEG improves representation quality, particularly under montage and brain-state shift, and increasing model capacity further enhances downstream transfer. These trends support viewing EEG foundation modelling as a scalable representation learning problem. More importantly, they suggest that invariance-oriented pre-training provides an effective organizing principle through which larger EEG corpora and larger models can be converted into more transferable neural representations.

\section{Ablation studies}


\begin{table}[!bp]
    \centering
    \caption{\textbf{Matrix for staged component ablation.}
    Check marks indicate components included in each variant. PCA Aug., multi-band PCA-based signal augmentation; Dynamic Views, construction of macro-level, micro-level and masked macro-level views with dynamic spatiotemporal configurations (i.e., varying channel and temporal extents); Freq. Branch, auxiliary frequency-domain embedding branch in addition to time-domain encoding; SH Enc., spherical-harmonics-based electrode encoding added to Cartesian coordinates; Channel-wise Self-Attn., alternating channel-wise and global-joint self-attention.}
    \label{tab:supp_component_design}
    \setlength{\tabcolsep}{1.0pt}
    \renewcommand{\arraystretch}{1.08}
    \begin{tabular*}{\textwidth}{@{\extracolsep{\fill}}clccccc@{}}
        \toprule
        ID & Variant
        & \shortstack{PCA\\Aug.}
        & \shortstack{Dynamic\\Views}
        & \shortstack{Freq.\\Branch}
        & \shortstack{SH\\Enc.}
        & \shortstack{Channel-wise\\Self-Attn.} \\
        \midrule
        A & Basic Baseline & - & - & - & - & - \\
        B & + Multi-band PCA Augmentation & \checkmark & - & - & - & - \\
        C & + Dynamic Multi-view Sampling & \checkmark & \checkmark & - & - & - \\
        D & + Dual-domain Signal Embedding & \checkmark & \checkmark & \checkmark & - & - \\
        E & + Geometry-aware Electrode Encoding & \checkmark & \checkmark & \checkmark & \checkmark & - \\
        F & + Channel-wise Self-attention & \checkmark & \checkmark & \checkmark & \checkmark & \checkmark \\
        \bottomrule
    \end{tabular*}
\end{table}

\begin{table}[!htbp]
    \centering
    \caption{\textbf{Performance of staged component ablation on Mumtaz2016 and FACED.}
    Mumtaz2016 is a montage-matched 10-20 EEG dataset for major depressive disorder classification, while FACED is a montage-shifted emotion recognition dataset with a low-density 10-10 montage configuration. Variant IDs correspond to Table~\ref{tab:supp_component_design}. Values are reported as mean $\pm$ s.d. over five random seeds.}
    \label{tab:supp_component_performance}
    \setlength{\tabcolsep}{2.0pt}
    \renewcommand{\arraystretch}{1.08}
    \begin{tabularx}{\textwidth}{c *{6}{Y}}
        \toprule
        \multirow{2}{*}{ID}
        & \multicolumn{3}{c}{Mumtaz2016 (2-Class)}
        & \multicolumn{3}{c}{FACED (9-Class)} \\
        \cmidrule(lr){2-4} \cmidrule(lr){5-7}
        & Bal. Acc. & AUROC & AUC-PR & Bal. Acc. & W. F1 & Kappa \\
        \midrule
        A & 90.40$_{\pm 0.28}$ & 96.90$_{\pm 0.48}$ & 97.51$_{\pm 0.34}$ & 71.32$_{\pm 0.62}$ & 71.18$_{\pm 0.47}$ & 67.47$_{\pm 0.60}$ \\
        B & 90.48$_{\pm 0.33}$ & 97.47$_{\pm 0.69}$ & 97.86$_{\pm 0.44}$ & 72.48$_{\pm 0.52}$ & 72.30$_{\pm 0.56}$ & 68.77$_{\pm 0.61}$ \\
        C & 91.00$_{\pm 0.96}$ & 97.88$_{\pm 0.64}$ & 98.15$_{\pm 0.43}$ & 74.23$_{\pm 0.60}$ & 74.05$_{\pm 0.65}$ & 70.74$_{\pm 0.70}$ \\
        D & 91.78$_{\pm 0.63}$ & 98.78$_{\pm 0.24}$ & 98.81$_{\pm 0.23}$ & 74.65$_{\pm 0.52}$ & 74.53$_{\pm 0.57}$ & 71.18$_{\pm 0.58}$ \\
        E & 92.08$_{\pm 0.69}$ & 99.01$_{\pm 0.27}$ & 99.04$_{\pm 0.23}$ & 78.54$_{\pm 0.85}$ & 78.23$_{\pm 0.74}$ & 75.50$_{\pm 0.86}$ \\
        F & \textbf{94.50}$_{\pm 0.65}$ & \textbf{99.45}$_{\pm 0.16}$ & \textbf{99.46}$_{\pm 0.15}$ & \textbf{84.75}$_{\pm 0.29}$ & \textbf{84.48}$_{\pm 0.31}$ & \textbf{82.51}$_{\pm 0.35}$ \\
        \bottomrule
    \end{tabularx}
\end{table}

We examine how each designed component of INCEPT contributes to downstream transfer performance. The ablation is conducted on two representative EEG datasets with different relationships to the pre-training electrode configuration. Mumtaz2016 uses a 19-channel 10-20 montage and is therefore closely aligned with the pre-training setting, whereas FACED uses a 32-channel low-density 10-10 montage and represents a more challenging montage-shifted evaluation scenario. This setup enables systematic assessment of how each component affects representation learning under both montage-matched and montage-shifted conditions.

All variants are pre-trained on the same unlabeled EEG corpus and evaluated under the same downstream full fine-tuning protocol. The \textbf{basic baseline} model follows a standard EEG foundation model design: it takes raw EEG segments without any augmentation, applies only time-domain patch embedding without frequency branch, uses Cartesian electrode coordinates as positional encoding, constructs deterministic multi-scale views (i.e., macro-level and micro-level views) using fixed channel subsets and segment durations, and applies global-joint self-attention Transformer blocks without within-channel modeling. In this setting, all EEG tokens are treated uniformly in a single global attention space, without explicitly modeling the spherical topology of EEG electrode arrangements. We then incrementally introduce four groups of components. First, \textbf{multi-band PCA-based augmentation} perturbs EEG signals in a low-dimensional principal subspace to improve robustness against amplitude and spectral variability; without this module, the model directly learns from raw normalized EEG segments. Second, \textbf{dynamic multi-view sampling} constructs macro-level, micro-level, and masked macro-level views with varying spatiotemporal extents, aligning with downstream variability; without it, the model relies on deterministic multi-scale views with fixed spatiotemporal configurations. Third, \textbf{dual-domain signal embedding} introduces an additional frequency-domain branch in parallel to the time-domain encoder to capture complementary spectral structure; without it, representation learning relies purely on temporal waveforms.  Fourth, \textbf{geometry-aware electrode encoding} introduces spherical-harmonics-based representations to capture non-Euclidean spatial relationships among EEG channels beyond Cartesian coordinates; without it, the model is restricted to Euclidean coordinate-based positional encoding only. Finally, \textbf{channel-wise self-attention} introduces an additional interaction modeling mechanism that alternates with global-joint self-attention to explicitly separate within-channel temporal structure from cross-channel dependencies. The component matrix is summarized in Table~\ref{tab:supp_component_design}, and the corresponding downstream performance is reported in Table~\ref{tab:supp_component_performance}. The final row (variant F) corresponds to the full INCEPT model used in the main experiments.

The ablation results show a consistent and cumulative improvement as components are introduced. Multi-band PCA-based augmentation improves robustness by reducing sensitivity to local amplitude fluctuations and spectral noise, while dynamic multi-view sampling adapts representation learning to variability in downstream configurations, particularly differences in channel number and segment duration across datasets. Dual-domain signal embedding improves representation quality by jointly modeling temporal waveforms and frequency-domain structure. 
Geometry-aware electrode encoding yields substantial improvement, with a more pronounced gain on FACED than on Mumtaz2016, indicating that spherical topology modeling is especially critical under montage shift. Channel-wise self-attention provides the largest improvement overall, achieving strong and consistent gains across both datasets. This suggests that explicitly separating within-channel temporal dynamics from inter-channel dependencies offers a complementary inductive bias to electrode-level geometric modeling. Overall, the full INCEPT configuration achieves the best performance across both datasets, demonstrating that every component is individually beneficial and that their complementary contributions collectively improve learning transferable EEG representations.

\end{appendices}


{\normalsize
\setlength{\bibsep}{3pt}
\bibliography{sn-bibliography}
}

\end{document}